\documentclass[manuscript,screen]{acmart}

\usepackage{hyperref}
\usepackage[utf8]{inputenc}
\usepackage{url}
\usepackage{comment}
\usepackage{microtype}
\usepackage{siunitx}
\usepackage{subcaption}
\usepackage{graphicx} 
\usepackage{wrapfig}
\usepackage{tabularx}
\usepackage{ragged2e}
\usepackage{pifont}
\usepackage[utf8]{inputenc}
\usepackage{algorithm}
\usepackage[noend]{algpseudocode}

\usepackage{longtable} 
\usepackage{pdflscape} 

\begin{document}

\title{Multimodal Methods for Analyzing Learning and Training Environments: A Systematic Literature Review}


\author{Clayton Cohn}
\email{clayton.a.cohn@vanderbilt.edu}
\orcid{0000-0003-0856-9587}
\affiliation{%
  \institution{Vanderbilt University}
  \city{Nashville}
  \state{TN}
  \country{USA}
}

\author{Eduardo Davalos}
\email{eduardo.davalos.anaya@vanderbilt.edu}
\orcid{0000-0001-7190-7273}
\affiliation{%
  \institution{Vanderbilt University}
  \city{Nashville}
  \state{TN}
  \country{USA}
}

\author{Caleb Vatral}
\email{cvatral@tnstate.edu}
\orcid{0000-0001-5710-2911}
\affiliation{%
  \institution{Tennessee State University}
  \city{Nashville}
  \state{TN}
  \country{USA}
}

\author{Joyce Horn Fonteles}
\email{joyce.h.fonteles@vanderbilt.edu}
\orcid{0000-0001-9862-8960}
\affiliation{%
  \institution{Vanderbilt University}
  \city{Nashville}
  \state{TN}
  \country{USA}
}

\author{Hanchen David Wang}
\email{hanchen.wang.1@vanderbilt.edu}
\orcid{0000-0001-5990-5865}
\affiliation{%
  \institution{Vanderbilt University}
  \city{Nashville}
  \state{TN}
  \country{USA}
}

\author{Austin Coursey}
\email{austin.c.coursey@vanderbilt.edu}
\orcid{0000-0003-1774-6442}
\affiliation{%
  \institution{Vanderbilt University}
  \city{Nashville}
  \state{TN}
  \country{USA}
}

\author{Surya Rayala}
\email{surya.chand.rayala@vanderbilt.edu}
\orcid{0009-0005-8192-8138}
\affiliation{%
  \institution{Vanderbilt University}
  \city{Nashville}
  \state{TN}
  \country{USA}
}

\author{Ashwin T S}
\email{ashwindixit9@gmail.com}
\orcid{0000-0002-1690-1626}
\affiliation{%
  \institution{Vanderbilt University}
  \city{Nashville}
  \state{TN}
  \country{USA}
}

\author{Meiyi Ma}
\email{meiyi.ma@vanderbilt.edu}
\orcid{0000-0001-6916-8774}
\affiliation{%
  \institution{Vanderbilt University}
  \city{Nashville}
  \state{TN}
  \country{USA}
}

\author{Gautam Biswas}
\email{gautam.biswas@vanderbilt.edu}
\orcid{0000-0002-2752-3878}
\affiliation{%
  \institution{Vanderbilt University}
  \city{Nashville}
  \state{TN}
  \country{USA}
}

\thanks{This work is supported under National Science Foundation grants IIS-2327708, DRL-2112635, and IIS-2017000; and US Army CCDC Soldier Center Award \#W912CG2220001. Any opinions, findings, conclusions, or recommendations expressed in this material are those of the authors and do not necessarily reflect the views of the National Science Foundation or United States Government, and no official endorsement by either party should be inferred.}

\renewcommand{\shortauthors}{Cohn et al.}


\begin{abstract}
Recent technological advancements in multimodal machine learning---including the rise of large language models (LLMs)---have improved our ability to collect, process, and analyze diverse multimodal data such as speech, video, and eye gaze in learning and training contexts. While prior reviews have addressed individual components of the multimodal pipeline (e.g., conceptual models, data fusion), a comprehensive review of empirical methods in applied multimodal environments remains notably absent. This review addresses that, introducing a taxonomy and framework that capture both established practices and recent innovations driven by LLMs and generative AI. We identify five \textit{modality groups}: Natural Language, Vision, Physiological Signals, Human-Centered Evidence, and Environment Logs. Our analysis reveals that integrating modalities enables richer insights into learner and trainee behaviors, revealing latent patterns often overlooked by unimodal approaches. However, persistent challenges in multimodal data collection and integration continue to hinder the adoption of these systems in real-time classroom settings.

\end{abstract}




\maketitle

\section{Introduction} \label{sec:intro}

Recent advances in education and technology are driving the personalization of educational and training curricula to meet the unique needs of learners and trainees. This shift is underpinned by data-driven approaches integrated into the field of \textit{learning analytics} \cite{WhatIsMMLA}, which focuses on gathering and evaluating data on learners' and trainees' behaviors to understand how they approach learning and training tasks \cite{maseleno2018demystifying, Zilvinskis2017}. For example, intelligent tutoring systems such as the Practical Algebra Tutor \cite{koedinger1997intelligent} focus on diagnosing student errors; open-ended environments like Betty's Brain \cite{leelawong2008designing} adaptively scaffold learning; and teacher-feedback tools (e.g., \citet{rodriguez2018teacher, Hutchins2023}) assist educators in enhancing instruction through insights into students' problem-solving processes.

As learning and training environments have grown more diverse and reliant on complex forms of human activity, learning analytics has expanded beyond its early dependence on environment log data \cite{azevedo2013using,kinnebrew2013contextualized,vanlehn2006behavior} to now incorporate multimodal evidence that captures the cognitive, affective, behavioral, and psychomotor dimensions of performance \cite{sharma2020multimodal,blikstein2016multimodal}. Contemporary learning and training scenarios include manipulating physical objects, both verbal and non-verbal communication, coordination within teams, and high-stakes medical procedural tasks---none of which can be adequately represented solely through digital traces. The growing use of video, audio, physiological signals, eye-tracking measures, and human-centered artifacts like surveys reflects a broader methodological transition in which multimodal data collection has become essential for understanding how learners and trainees engage with instructional tasks under real-world constraints \cite{cukurova2020promise,ochoa2017multimodal}. 

This shift has made applied multimodal learning analytics (MMLA\footnote{While MMLA traditionally refers to multimodal learning analytics, this review uses the term to encompass both learning and training environments.}) a critical area of inquiry, requiring systematic attention to how multimodal data are obtained, transformed, and analyzed in authentic settings.

\subsection{Motivation}
\label{subsec:motivation}

Applied MMLA differs fundamentally from multimodal research in core AI and machine learning. In real-world settings, data must be collected under conditions that introduce noise, partial observability, missing data, and privacy constraints \cite{slade2013learning,kitto2019practical}. Dataset size and structure are often limited by institutional schedules, small participant pools, and infrequent opportunities for repeated measurement \cite{cukurova2020promise,martinez2023lessons}. Annotation frequently depends on human expertise, subjective judgment, and labor-intensive manual coding, introducing methodological constraints not present in large-scale AI corpora \cite{fonteles2026JLI}. These factors shape every stage of the multimodal pipeline---from sensing and feature extraction to data fusion and analysis. Consequently, applied MMLA requires methodological adaptations distinct from those in multimodal machine learning, as practices must align with pedagogical goals, maintain ecological validity, and operate within practical constraints. 

Prior surveys primarily describe fusion taxonomies \cite{chango2022review,zhao2024deep,mu2020multimodal}, conceptual frameworks \cite{di2018signals,ochoa2022multimodal}, methodological overviews \cite{giannakos2023role}, cross-domain multimodal machine learning \cite{fang2022multi}, or domain-specific applications \cite{zhu2024review} like healthcare \cite{begum2024federated}, sensing \cite{ma2024multilevel}. They do not synthesize how methodological decisions are shaped by the constraints of real-world data collection. These constraints are unique to applied settings and are central to research in classrooms, clinical simulations, workplace training, and other settings where MMLA is increasingly deployed; yet they remain underexplored in the current literature. A comprehensive synthesis of applied multimodal methods is, therefore, needed to support the design, implementation, and interpretation of multimodal analyses in learning and training contexts.

This need is amplified by recent advances in machine learning, particularly generative AI (GenAI) and large multimodal models (LLMs) that have transformed natural language processing, computer vision, and cross-modal representation learning \cite{yan2024generative,liu2024llava}. The advent of GenAI has significantly impacted how MMLA is practiced, marking a fundamental shift in the field's capabilities and paradigms. As a result, the literature increasingly bifurcates into GenAI and non-GenAI approaches, with each operating under different assumptions about model capability, data availability, and the role of automation in analytic workflows. Recognizing and unpacking this divide is essential for contextualizing current methods within the rapidly evolving MMLA landscape---yet it remains unaddressed in prior reviews. 

In this review, we address these gaps by providing an applied methodological grounding for multimodal learning and training research, examining how methods are implemented in real-world contexts, and synthesizing empirical practices across diverse environments, modalities, and analytic approaches. We also consider how recent advances in LLMs and GenAI---particularly following the release of ChatGPT---are beginning to reshape data collection, analysis, and system design within this space.

\subsection{Contributions}

This review makes four primary contributions, aligned with the need to survey applied multimodal learning and training methodological research. We align each of the four core contributions with a specific study objective, presented in the following table.

\begin{table}[htbp]
    \centering
    \begin{tabular}{p{0.20\textwidth} p{0.26\textwidth} p{0.38\textwidth}}
    \hline
    \textbf{Focus} & \textbf{Objective} & \textbf{Contribution} \\
    \hline
    
    \textbf{Characterizing the Methodological Landscape} &
    To characterize how multimodal data are collected, fused, and analyzed in applied learning and training contexts. &
    We systematically map the methodological space across data collection media, modality types, fusion strategies, and analysis approaches, offering a comprehensive account of empirical practices in the field. \\
    
    \hline
    
    \textbf{A Unified Framework and Taxonomy} &
    To organize and make sense of the methodological diversity in applied multimodal learning analytics. &
    We introduce a unified, empirically grounded framework and taxonomy that reveal how methodological choices are shaped by environment characteristics, data observability, and analytic goals, synthesizing diverse empirical methods into a coherent analytic structure. \\
    
    \hline
    
    \textbf{Synthesizing Methodological Challenges} &
    To identify and structure the key challenges encountered in real-world multimodal learning and training research. &
    We identify persistent methodological and practical issues and examine their implications for system design, deployment, and evaluation. \\
    
    \hline
    
    \textbf{Methodological Archetypes} &
    To contextualize how multimodal methods are employed to address different classes of research problems.\footnotemark &
    We identify recurring methodological archetypes and illustrate their application through representative case studies, demonstrating what multimodal learning and training methodology looks like in practice. \\
    
    \hline
    \end{tabular}
    \label{tab:objectives_contributions}
\end{table}
\footnotetext{Although archetypes emerged inductively during analysis, we formalized them as an objective once their explanatory value became clear.}

\subsection{Literature Review Structure}

Section~\ref{sec:background} provides background and related work, situating this review within the broader landscape of multimodal learning and training analytics, multimodal data fusion, and prior multimodal literature reviews. Section~\ref{sec:methods} outlines the methodology used to construct the review corpus, including scope definition, literature search, study selection, and feature extraction and analysis procedures. Section~\ref{sec:results} presents our theoretical framework and taxonomy for multimodal methods in learning and training environments, derived empirically through our analysis of the literature. Findings and taxonomies are aligned to each framework component, with illustrative examples of how they are applied in practice. Section~\ref{sec:archetypes} introduces three research archetypes that characterize how MMLA research is conducted in practice, each accompanied by a real-world case study. Section~\ref{sec:discussion} discusses persistent challenges, directions for future research, and the implications of our findings.

\section{Background and Related Work} \label{sec:background}

Across more than a decade of research, MMLA has been shaped by conceptual frameworks, sensing technologies, and advances in machine learning that seek to understand how multimodal evidence informs the study of learning processes. Foundational contributions include work on data observability \cite{sharma2020multimodal}, the distinction between directly measurable signals and inferred constructs \cite{di2018signals}, and the role of feature-level and decision-level fusion in multimodal pipelines \cite{zhao2024deep,li2024multimodal}. These developments establish MMLA as a field focused on how various types of evidence are collected, transformed, integrated, and interpreted to explain learner behavior and design instructional support. As the collection of multimodal data becomes more prevalent in real-world learning and training environments, it is essential to systematically synthesize the methodological foundations of MMLA.

\begin{table*}[t]
    \centering
    \caption{Comparison of Existing Multimodal Review Studies and Contributions of This Review}
    \label{tab:review_comparison}
    \renewcommand{\arraystretch}{1.25}
    \small
    
    \begin{tabular}{lcccccccccccc}
        \toprule
        \textbf{Review} &
        \textbf{V} & 
        \textbf{A} & 
        \textbf{T} & 
        \textbf{S} & 
        \textbf{H} &
        \textbf{Fus.} &
        \textbf{App.} &
        \textbf{X-Mod.} &
        \textbf{Trans.} &
        \textbf{MMLA} &
        \textbf{L/T Env} &
        \textbf{Sim} \\
        
         & 
        \textbf{} & 
        \textbf{} & 
        \textbf{} & 
        \textbf{} & 
        \textbf{} &
        \textbf{} &
        \textbf{Chall.} &
        \textbf{Chall.} &
        \textbf{} &
        \textbf{Focus} &
        \textbf{Focus} &
        \textbf{} \\
        
        \midrule
        \multicolumn{13}{l}{\textit{Education-Focused MMLA Reviews}} \\
        \midrule
        
         Di Mitri et al. (2018) \cite{di2018signals} & \checkmark & \checkmark & \checkmark & \checkmark & \checkmark & \checkmark & \texttimes & \texttimes & \texttimes & \checkmark & \checkmark & \texttimes \\
        Chango et al. (2022) \cite{chango2022review}  & \checkmark & \checkmark & \checkmark & \checkmark & \texttimes & \checkmark & \texttimes & \texttimes & \texttimes & \checkmark & \checkmark & \texttimes \\
        Alwahaby et al. (2022) \cite{alwahaby2022evidence} & \checkmark & \checkmark & \checkmark & \checkmark & \checkmark & \texttimes & \texttimes & \texttimes & \texttimes & \checkmark & \checkmark & \texttimes \\
        Shankar et al. (2018) \cite{8433495} & \checkmark & \checkmark & \checkmark & \checkmark & \checkmark & \checkmark & \texttimes & \texttimes & \texttimes & \checkmark & \checkmark & \texttimes \\
        Crescenzi et al. (2020) \cite{crescenzi2020multimodal} & \checkmark & \checkmark & \checkmark & \checkmark & \checkmark & \texttimes & \checkmark & \texttimes & \texttimes & \checkmark & \checkmark & \texttimes \\
        Mu et al. (2020) \cite{mu2020multimodal} & \checkmark & \checkmark & \checkmark & \checkmark & \checkmark & \checkmark & \texttimes & \texttimes & \texttimes & \checkmark & \checkmark & \texttimes \\
        
        \midrule
        \multicolumn{13}{l}{\textit{General Multimodal Deep Fusion Surveys}} \\
        \midrule
        
        Zhao et al. (2024) \cite{zhao2024deep} & \checkmark & \checkmark & \checkmark & \checkmark & \texttimes & \checkmark & \texttimes & \checkmark & \checkmark & \texttimes & \texttimes & \texttimes \\
        Hussain et al. (2024) \cite{hussain2024comprehensive} & \checkmark & \checkmark & \checkmark & \checkmark & \texttimes & \checkmark & \texttimes & \checkmark & \checkmark & \texttimes & \texttimes & \texttimes \\
        Gaw et al. (2022) \cite{gaw2022multimodal} & \checkmark & \checkmark & \texttimes & \checkmark & \texttimes & \checkmark & \checkmark & \checkmark & \texttimes & \texttimes & \texttimes & \texttimes \\
        Teoh et al. (2024) \cite{teoh2024advancing} & \checkmark & \checkmark & \checkmark & \checkmark & \texttimes & \checkmark & \checkmark & \checkmark & \texttimes & \texttimes & \texttimes & \checkmark \\
        Cengiz et al. (2025) \cite{cengiz2025survey} & \texttimes & \texttimes & \checkmark & \checkmark & \texttimes & \checkmark & \texttimes & \checkmark & \checkmark & \texttimes & \texttimes & \texttimes \\
        Mondal et al. (2025) \cite{mondal2025survey} & \checkmark & \checkmark & \checkmark & \checkmark & \texttimes & \checkmark & \texttimes & \checkmark & \checkmark & \texttimes & \texttimes & \texttimes \\
        
        \midrule
        \multicolumn{13}{l}{\textit{Healthcare / Sensing Reviews}} \\
        \midrule
        
        Shaik et al. (2024) \cite{shaik2024survey} & \checkmark & \checkmark & \checkmark & \checkmark & \texttimes & \checkmark & \checkmark & \checkmark & \texttimes & \texttimes & \texttimes & \checkmark \\
        Khoo et al. (2024) \cite{khoo2024machine} & \texttimes & \checkmark & \checkmark & \checkmark & \checkmark & \checkmark & \checkmark & \checkmark & \texttimes & \texttimes & \texttimes & \texttimes \\
        
        \midrule
        
        \textbf{This Review (2025)} & 
        \textbf{\checkmark} & \textbf{\checkmark} & \textbf{\checkmark} & \textbf{\checkmark} & \textbf{\checkmark} & \textbf{\checkmark} & \textbf{\checkmark} & \textbf{\checkmark} & \textbf{\checkmark} & \textbf{\checkmark} & \textbf{\checkmark} & \textbf{\checkmark} \\
        
        \bottomrule
    \end{tabular}
    \begin{flushleft}
        \footnotesize
        \textbf{Legend:} V = Vision, A = Audio, T = Text, S = Sensors, H = Human-centered;  
        Fus. = discusses any multimodal fusion strategy;  
        App. Chall. = applied, real-world data collection challenges and constraints;
        X-Mod. Chall. = cross-modal interaction challenges;  
        Trans. = transformer-era multimodal methods covered;  
        MMLA Focus = directly aligned with multimodal learning analytics;  
        L/T Env = learning or training environment relevance;  
        Sim = simulation-based or training simulation environments.
    \end{flushleft}                 
\end{table*}

Prior surveys and literature reviews of multimodal learning analytics span several domains and research traditions, including education-focused MMLA \cite{giannakos2023role}, general multimodal machine learning \cite{baltruvsaitis2018multimodal}, healthcare sensing \cite{shaik2024survey,khoo2024machine}, and multimodal systems engineering \cite{liang2024foundations}. Table~\ref{tab:review_comparison} summarizes these reviews across twelve methodological dimensions central to applied multimodal learning and training analytics. Education-focused reviews, such as \citet{di2018signals}, \citet{chango2022review}, \citet{alwahaby2022evidence}, and related work, established influential conceptual models and taxonomies but offered limited analysis of how multimodal methods operate under real-world constraints such as noise, occlusion, privacy restrictions, and small datasets. These reviews also preceded the widespread use of transformer-based multimodal modeling and therefore do not incorporate recent advances in GenAI and LLMs.

General multimodal fusion surveys, including \citet{zhao2024deep}, \citet{hussain2024comprehensive}, \citet{mondal2025survey}, and \citet{gaw2022multimodal}, provide comprehensive treatments of multimodal integration techniques but are oriented toward large-scale machine learning applications rather than learning or training environments. They assume controlled data-collection conditions, high-volume datasets, and domain-general modeling objectives, thereby limiting their applicability to the methodological challenges faced in applied MMLA. Surveys in healthcare and sensing, such as \citet{soenksen2022integrated} and \citet{khoo2024machine}, examine multimodal data collection under conditions of measurement noise and human variability, but they do not articulate the pedagogical, behavioral, or feedback-oriented dimensions characteristic of learning and training research.

Our literature review shows that no existing review synthesizes applied methodological practices across learning and training environments, including the implications of human-subject constraints, ecological validity, and task structure. Prior surveys do not examine the methodological consequences of integrating heterogeneous modalities under real-world conditions---particularly cross-modal interactions and temporal alignment. No evidence-derived taxonomy exists that is grounded in how multimodal data are actually collected, transformed, and analyzed in applied MMLA research. These limitations underscore the need for a review that connects methodology to the characteristics of learning and training environments while incorporating recent advances in multimodal modeling.

\section{Methods} \label{sec:methods}

This section describes the procedures used to construct the literature corpus, filter the corpus using quantitative and qualitative techniques, extract methodological features, and prepare the data for analysis. The goal of this process is to obtain a representative set of studies that examine multimodal data collection and analysis in learning and training environments. Each stage of our pipeline was designed to support the development of the unified framework and taxonomy presented in Section~\ref{sec:framework_taxonomy}.

As outlined in Section~\ref{sec:intro}, there is a pressing need to examine how the emergence of LLMs has reshaped the MMLA landscape. To address this, this literature review is partitioned into two distinct corpora: Corpus A includes publications from 2017-2022 (preceding the release of ChatGPT), and Corpus B comprises publications from 2022-2025 (after the release of ChatGPT). Independent literature searches were conducted for each corpus using comparable search strategies and inclusion criteria, unless otherwise noted. Comparative analysis of the two corpora is presented in Section~\ref{sec:framework_taxonomy}, and the full details of the search and distillation protocol are provided in Appendix~\ref{app:corpus_distillation_procedure}.

\subsection{Scope}

This review focuses on multimodal learning and training analytics in empirical studies conducted in authentic instructional and training settings. We exclude studies centered exclusively on virtual reality environments, as they face scalability constraints for applied educational deployment. These boundaries informed both the inclusion criteria and corpus filtering, ensuring methodological alignment with an applied synthesis.

Given inconsistent terminology across the literature, we define key concepts to clarify our analytical focus and justify the scope of this review. \textit{Learning} environments emphasize knowledge acquisition and conceptual understanding through didactic instruction, exploration, and problem-solving, whereas \textit{training} environments focus on skill development through structured practice and repetition. A \textit{data collection medium} is a raw data stream produced by a sensing or logging device (e.g., video, audio, physiological sensors). A \textit{modality} is a derived attribute conveying a specific type of information (e.g., affect, pose, gaze, transcribed speech), and may be computed from one or more data streams. The mapping between mediums and modalities is many-to-many: a single video stream can yield both affect and pose, while affect can also be derived from fused audio and video. \textit{Modality groups} are inductively identified clusters of related modalities used to structure our taxonomy. We adopt the term \textit{multimodal} to refer to studies involving multiple modalities or synchronized data streams, following established usage in learning analytics~\cite{blikstein2013multimodal}.

\subsection{Search Strategy}
\label{sec:search_strategy}

The literature search was conducted programmatically using Google Scholar via SerpAPI, which was chosen for its ability to reliably return organic search results across different queries. Twenty-one search queries were used to conduct our search (see Appendix~\ref{app:corpus_distillation_procedure}). These queries were informed by the authors' expertise in multimodal learning analytics and aligned with the objectives of an applied methodological review focused on the collection, transformation, and analysis of multimodal data across various learning and training contexts. For each query, the top 100 results were retrieved (five pages from Google Scholar with 20 records per page). Duplicate records were removed using hash-based matching on the UUIDs provided by Google Scholar, and non-English papers were excluded from the dataset. This process resulted in an initial corpus of 2,120 papers for Corpus A and 845 for Corpus B (Corpus B is smaller because LLMs have only recently emerged, compared to the more extended history of MMLA).

\subsection{Inclusion and Exclusion Criteria}
\label{sec:inclusion_exclusion}

Inclusion and exclusion criteria were guided by the need to focus on applied multimodal methods in authentic learning and training environments. A study was included if it (1) collected or analyzed data from at least two modalities or two data collection media; and (2) examined a learning or training setting involving human participants. These settings include fully physical environments (e.g., traditional classroom instruction), mixed-reality contexts (e.g., manikin-based nursing simulations), and technology-enhanced or online instructional environments (e.g., Khan Academy). Studies were excluded if they involved purely synthetic datasets, controlled laboratory tasks lacking instructional relevance, medical imaging applications, or multimodal architectures designed only for model training.

\subsection{Corpus Filtering}
\label{sec:cgp}

Following the initial literature search, we applied a programmatic corpus reduction procedure to identify studies closely aligned with the multimodal learning and training literature, distilling the corpus to a practical size for manual review. We created directed citation graphs using NetworkX, where nodes represent papers and edges represent citation relationships retrieved via SerpAPI (i.e., $\textit{a} \rightarrow \textit{b}$ indicates paper $\textit{a}$ cites paper $\textit{b}$). We then iteratively pruned the graph by removing weakly-connected nodes (treating citation relationships symmetrically) until none remained. We refer to this process as \textit{citation graph pruning}, which serves as a pragmatic filtering heuristic (detailed in Appendix~\ref{app:corpus_distillation_procedure}). This process resulted in 1,063 remaining papers for Corpus A and 559 remaining papers for Corpus B. 

\begin{wrapfigure}{r}{0.5\linewidth}
    \centering
    \includegraphics[width=\linewidth]{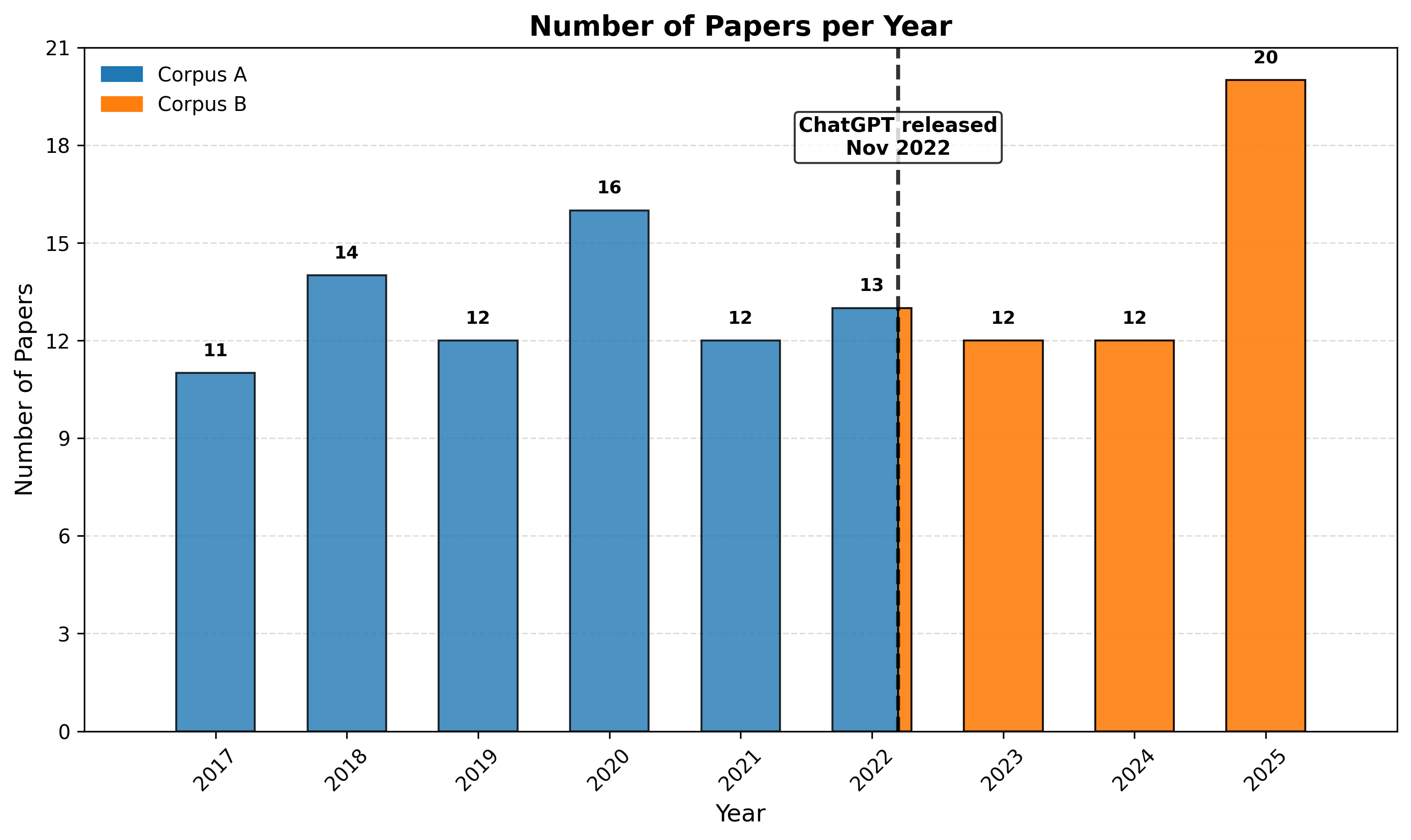}
    \caption{Distribution of full corpus papers by year. Blue bars represent works
    published prior to the release of ChatGPT in November 2022 (Corpus A); orange bars
    represent those published afterward (Corpus B). \textit{Number of Papers} refers to the number of papers selected for this review.}
    \label{fig:corpus_dist_by_year}
\end{wrapfigure}

Each paper in both corpora was then qualitatively screened by at least two of the authors of this paper. We used a multi-stage review process based on Kitchenham's systematic review methodology~\cite{kitchenham2004procedures}---involving sequential filtering by title, abstract, and full text---which we adapted to the goals of an applied multimodal synthesis. Title filtering was employed to exclude clearly irrelevant papers, such as those focused on multimodal neural network training, medical applications, or general signal processing---papers that utilized multimodality but did not apply it to learning or training environments. Abstract filtering removed papers for which both reviewers agreed the content fell outside the scope of this review. Full-text screening applied the exclusion criteria detailed in Section~\ref{sec:inclusion_exclusion}. This process resulted in 73 papers for Corpus A and 49 for Corpus B, totaling 122 papers in the final corpus used for analysis. Figure~\ref{fig:corpus_dist_by_year} shows the distribution of included papers by year, with a dividing line when ChatGPT was released at the end of 2022.\footnote{Five papers included in Corpus B were published in 2022.}

\subsection{Data Extraction}
\label{subsec:data_extraction}

All of the 122 papers were coded to capture methodological characteristics related to multimodal learning and training analytics. The features extracted for analysis included data collection media, derived modalities, fusion strategies, analysis methods, types of environments, participant interaction structures, didactic characteristics, levels of instruction, domains of study, student and system feedback, and publication metadata. These features provide the empirical foundation for the unified framework and taxonomy discussed in Section~\ref{sec:framework_taxonomy}. Notably, several categorical distinctions---such as modality groups and participant interaction structures---were not predetermined. Instead, these categories emerged inductively during the extraction process as recurring patterns observed across studies. This approach aligns with creating a bottom-up taxonomy based on empirical evidence rather than conceptual assumptions. Complete definitions of all extracted features and their possible values are presented in Section~\ref{sec:framework_taxonomy}, with further elaboration in Appendix~\ref{subsec:feature_extraction_appendix}.

\subsection{Analysis Procedure}
\label{subsec:analysis_procedure}

The extracted features were analyzed through qualitative thematic analysis informed by the framework we present in Figure~\ref{fig:framework}. Studies were grouped by modality, environment, and analytic approach to identify methodological trends and patterns. Five modality groups---(1) natural language, (2) vision, (3) physiological signals, (4) human-centered evidence, and (5) logs---were derived through iterative coding and used to structure the analysis presented in the following section. Within each group, we examined common fusion strategies and analytic methods, along with challenges and constraints unique to applied settings. Our analysis also identified three archetypes of multimodal learning and training research, which are presented in Section~\ref{sec:archetypes}. These archetypes highlight recurring methodological configurations and illustrate how multimodal methods are aligned with different research aims. Together, these procedures support the development of a unified, empirically grounded understanding of multimodal methods in learning and training environments.

\section{Results}
\label{sec:results} \label{sec:framework_taxonomy}

\begin{figure}[htbp]
    \centering
    \begin{picture}(400, 200) 
    \put(0,0){\includegraphics[width=0.9\textwidth]{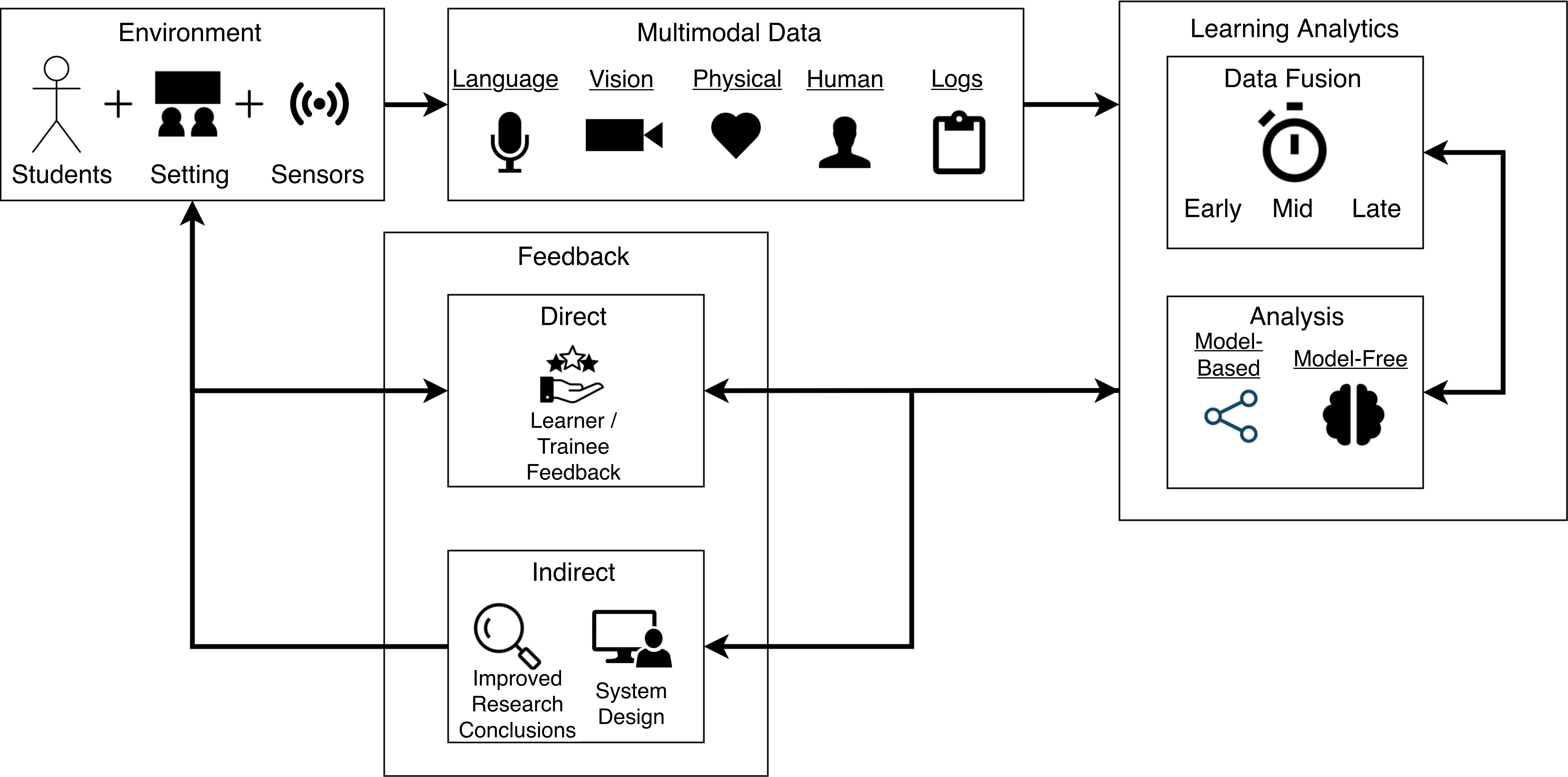}}
     \put(83,180){\makebox(0,0)[b]{\tiny  \colorbox[gray]{.95}{Sec. \ref{subsec:environment_type}}}}
     \put(241,180){\makebox(0,0)[b]{\tiny  \colorbox[gray]{.95}{Sec. \ref{subsec:multimodal_data}}}}
     \put(375,181){\makebox(0,0)[b]{\tiny  \colorbox[gray]{.95}{Sec. \ref{subsec:learning_analytics}}}}
     \put(178,124){\makebox(0,0)[b]{\tiny  \colorbox[gray]{.95}{Sec. \ref{subsec:feedback}}}}
     \end{picture}
    \caption{Multimodal Learning and Training Environments Literature Review Framework}
    \label{fig:framework}
    \Description[Multimodal Learning and Training Environments Framework]{Multimodal Learning and Training Environments Framework}
\end{figure}

Through our inductive analysis of the review corpus, we developed a theoretical framework that captures the core components of multimodal learning and training pipelines along with their interrelations. As illustrated in Figure~\ref{fig:framework}, the framework decomposes the MMLA process into four primary, sequential component processes: (1) the learning or training environment from which student data are collected through sensors, (2) multimodal data and the modalities derived from them, (3) learning analytics for making sense of that data, and (4) feedback for stakeholders like students, teachers, and researchers. We provide an overview of each component below, followed by subsections presenting taxonomies and findings corresponding to each.

In the following subsections, each framework component is presented via: (1) its significance within the context of multimodal learning and training methodology; (2) a taxonomy derived from data extracted from the reviewed studies; (3) relevant findings, including a comparison of methodologies from the pre-LLM and post-LLM eras and their challenges; and (4) examples of how each component is put into practice.

\subsection{Environments}
\label{subsec:environment_type} 

\begin{wrapfigure}{R}{0.5\textwidth}
    \begin{center}
\includegraphics[width=0.4\textwidth]{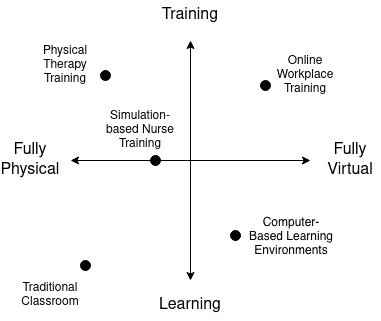}
    \end{center}
    \caption{Learning-Training Continuum}
    \label{fig:ltcontinuum}
\end{wrapfigure}

Our paper explores a spectrum of environments on a learning-training continuum (Figure~\ref{fig:ltcontinuum}). The environments span from traditional classrooms to online courses and are categorized along two dimensions: the learning-training axis \cite{2345021698,2055153191,85990093,3625722965} and the physical-virtual space continuum \cite{4019205162,1326191931,2836996318}. Systems such as nurse training simulations with manikins and embodied learning environments used in K-12 education (where students actively move around the classroom as part of the learning experience~\cite{fonteles2026JLI}) combine physical and virtual elements and are referred to as \textit{mixed-reality} environments \cite{kaplan2021effects,ke2016teaching}.

Multimodal methods in learning environments aim to enhance educational outcomes by analyzing student engagement and learning patterns. In contrast, training environments focus on skill acquisition and task proficiency, serving individuals from personal development to professional enhancement across fields such as healthcare \cite{di2020real}, athletics \cite{3625722965}, the workplace \cite{algerafi2023unlocking,kaplan2021effects}, and the military \cite{853680639}. These settings range from fully virtual simulations to physical training drills, with mixed reality bridging the gap. MMLA objectives differ between learning and training, requiring context-specific strategies. While the distinction between learning and training can be ambiguous, as seen in game-based platforms \cite{1847468084,666050348}, our review spans this spectrum. We employ a fuzzy qualitative categorization to place each study on this continuum, acknowledging the approach's complexity and utility for analyzing MMLA research sub-communities.

In the following subsections, we present findings for the three components specified in our framework for environment: \textbf{learners/trainees} (students), \textbf{setting}, and \textbf{data collection media} (sensors). 

\subsubsection{Learners/Trainees (``Students'' in Figure~\ref{fig:framework})} \label{subsubsec:environment_subject}

Learners and trainees are central to the design, deployment, and evaluation of multimodal learning and training analytics systems. The identity of the participants, the subjects they study, their methods of interaction, and the instructional settings in which they are situated influence the multimodal data that can be collected, the models that can be developed, and how the resulting analytics should be implemented. Therefore, clearly defining these learner characteristics is essential to our framework and provides a consistent perspective for understanding how studies are situated within authentic learning and training environments.

Across both Corpora A and B, we describe the learner context along four dimensions: (1) \textbf{domain of study}, (2) \textbf{participant interaction structure}, (3) \textbf{didactic nature of the environment}, and (4) \textbf{level of instruction or training}. The same taxonomy is applied throughout, and individual studies can receive multiple labels along each dimension.

\paragraph{Domain of Study.}

Our corpus revealed three primary domains of study. \textbf{STEM+C} includes Science, Technology, Engineering, Mathematics, and Computing, as well as healthcare and medicine~\cite{818492192,1118315889,3888330750}. \textbf{Humanities} spans literature, debate, oral presentation, and writing~\cite{3408664396,1844320601,85990093}. \textbf{Psychomotor Skills} refers to domains emphasizing motor coordination, such as CPR training~\cite{2070224207}, woodworking~\cite{1731146538}, and video games like PAC-MAN~\cite{4278392816}.

Both Corpora~A and~B primarily focused on STEM+C learning (A: 55/73, 75\%; B: 37/49, 76\%), covering topics from programming~\cite{2936220551} to nursing~\cite{4035649049} to geometry and chemistry~\cite{1196965665}. Psychomotor skills were less represented (A: 5/73, 7\%; B: 1/49, 2\%). Corpus~B showed increased attention to the humanities (A: 11/73, 15\%; B: 13/49, 27\%), where LLMs enabled support for open-ended tasks such as multimodal composition and essay writing~\cite{1161441004}.

A key distinction between domains lies in their level of structure: \textbf{structured} domains (e.g., STEM+C) are characterized by constrained problems and clear evaluation metrics. In contrast, \textbf{unstructured} domains (e.g., humanities) involve open-ended tasks with subjective or variable outcomes. For instance, algebra problems typically follow well-defined procedures. They can be assessed using rule-based systems such as decision trees~\cite{10628321}, whereas creative writing tasks~\cite{zedelius2019beyond} resist formulaic evaluation and are often poorly served by metrics like word count or sentence length~\cite{gansle2002moving}. However, with the advent of LLMs, analytical rubrics are being developed that break down writing into specific criteria, such as uniqueness of storyline, logical sequencing of ideas and content (e.g., a connected beginning, middle, and end of the narrative), style of writing, and engaging vocabulary \cite{gomez2023confederacy,kim2025evaluating}. 

Structural differences affect how multimodal learning environments are designed and how data are captured and analyzed. Structured domains often afford quantitative, model-based analysis, while unstructured domains typically require model-free, qualitative methods such as thematic or interaction coding~\cite{jiang2022empirical}. However, the line between the two is becoming increasingly blurred with the integration of LLMs, which can interpret and evaluate multimodal data even in unstructured, dynamic contexts such as embodied learning.

\paragraph{Participant Interaction Structure.}

Participant interaction structure describes whether learners engage individually (\textbf{individual}) or with others (\textbf{multi-learner}). Individual settings typically involve a single learner interacting with systems such as educational games~\cite{3856280479}, intelligent tutoring systems~\cite{4277812050}, open-ended learning environments~\cite{leelawong2008designing}, and creative platforms~\cite{3537775194}. Multi-learner environments include pairs, small groups, or full-class activities such as paired programming~\cite{3308658121}, game-based competitions~\cite{3135645357}, and collaborative play~\cite{2668965770}.

Our corpus reveals a growing emphasis on collaborative learning, with multi-learner studies increasing from 42\% (31/73) in Corpus~A~\cite{1296637108,3051560548} to 51\% (25/49) in Corpus~B~\cite{2995141815,3522635517}. These studies leverage multimodal data to examine activities such as collaborative experimentation~\cite{1285699194}, clinical simulations~\cite{209328204}, and group reflections using dashboards~\cite{1040787959,3888330750}. In contrast, individual-focused studies emphasize personalization~\cite{177743022} and self-regulation, often through AI-driven tutors \cite{2166765216}, gaze-adaptive systems \cite{1844320601}, or LLM-integrated dashboards~\cite{1441411748}.

Multi-learner settings introduce social interaction as a central dimension, enabling insights not possible in individual contexts. Learners often externalize their thinking, allowing researchers to analyze dialogue \cite{209328204}, coordination \cite{227355655}, and socially shared regulation of learning (SSRL) via audio and video \cite{jarvela2023predicting}. However, such contexts present analytic and logistical challenges, including smaller effective sample sizes ($n$)~\cite{janssen2013multilevel} and reduced transcription quality due to classroom noise~\cite{blanchard2015study}.

\paragraph{Didactic Nature of the Environment.} 

Didactic nature refers to how learning or training is presented to participants. \textbf{Instruction} involves formal activities with defined objectives (e.g., courses, labs)~\cite{4019205162,1326191931,141378338}. \textbf{Training} emphasizes skill development through practice (e.g., clinical simulations, vocational drills)~\cite{4035649049,3625722965,martinez2023lessons}. \textbf{Informal} settings lack fixed goals and occur in loosely structured contexts (e.g., game-based learning, exploratory play)~\cite{1598166515,123412197,3809293172,666050348}.

Both corpora are dominated by instruction, with an even stronger emphasis in Corpus~B (A: 45/73; 62\%; B: 40/49; 82\%). For example, \citet{3783339081} analyzed student interactions in a chemistry virtual lab, integrating logs with audio and video to uncover learning difficulties not evident from system data alone. Training environments accounted for 20-25\% of both corpora (A: 15/73; 21\%; B: 11/49; 23\%), such as a simulated social skills trainer for youth with autism using audiovisual cues like head pose and smiling ratio. Informal learning settings declined notably from Corpus~A to~B (A: 12/73; 16\%; B: 1/49; 2\%), as LLMs were primarily applied in traditional instructional contexts. For instance, \citet{1844320601} combined real-time gaze-based engagement detection with ChatGPT-generated summaries to support reading comprehension while studying.

Training prioritizes repetition and performance \cite{3625722965}, while instructional and informal settings differ in design, data, and analysis. Instructional tasks are structured, allowing controlled data capture and model-based analysis \cite{zha2025data}. Informal settings are open-ended, generating noisy, varied, and often incomplete data requiring qualitative, model-free, and human-in-the-loop decision-making approaches \cite{amershi2014power,song2022learning}. Goals also vary: instruction targets conceptual understanding \cite{jonassen2006role}, while informal learning fosters creativity, exploration, and inquiry \cite{doering2015fostering}.

\paragraph{Level of Instruction or Training.} 

Our framework defines three levels of instruction: (1) \textbf{K--12} (primary and secondary education)~\cite{1581261659,3051560548,141378338}, (2) \textbf{University} (undergraduate and graduate)~\cite{2070224207,4019205162,190066185}, and (3) \textbf{Professional Development} (e.g., workplace learning, continuing education)~\cite{3135645357,di2020real,3537775194}.

Both corpora show similar trends across instructional level, skewed heavily toward university learners (A: 36/73; 49\%; B: 29/49; 59\%). For example, \citet{3304069824} demonstrates that ChatGPT's classroom integration, paired with physiological signals (e.g., galvanic skin response) supports emotional state detection and structured AI-tutoring interventions with college students, improving both engagement and learning outcomes. K--12 settings followed (A: 30/73; 41\%; B: 19/49; 39\%), while professional development was least represented (A: 5/73; 7\%; B: 2/49; 4\%).

K-12 and adult learning contexts pose distinct challenges. Research in K-12 settings faces significant ethical and logistical constraints due to the involvement of minors. Multimodal data capture, especially video and physiological sensing, raises privacy and health data concerns, often requiring approval from parents, teachers, administrators, and district officials~\cite{cohn2025theory}. The emergence of GenAI raises additional concerns, including student misuse \cite{mintz2023artificial} and unintended LLM behaviors \cite{jiao2025llms}, making school-based research difficult and often requiring on-site presence.

In contrast, adult learning environments offer greater flexibility, with fewer institutional hurdles and support for both in-person and remote studies. Despite the formative importance of K--12 education---where students acquire foundational knowledge, social skills, problem-solving strategies, and good study habits \cite{bransford2000people,english2013supporting}---the dominance of university-focused research is unsurprising. Expanding the reach of multimodal systems in K--12 contexts will require coordinated efforts from educators, researchers, parents, and policymakers to ensure these technologies are deployed ethically and effectively \cite{si2022multimodal}.

\subsubsection{Setting} \label{subsubsec:environment_setting}

Settings describe where and how multimodal learning and training activities unfold. Whether learners are on virtual platforms, in physical classrooms, in clinical simulations, or in play spaces constrains which traces can be captured and how analytics and AI-based tools can be meaningfully embedded. Setting links sensing choices, models, and interpretations to the realities of computer-based, in-person, and blended scenarios, and clarifies how multimodal learning analytics systems are deployed across different contexts. With both corpora, we characterize setting along two dimensions: (1) \textbf{environment function} and (2) \textbf{environment interaction setting}.

\paragraph{Environment Function.}

We distinguish environments by their primary function, in line with this review's dual focus on \textbf{learning}~\cite{4019205162,86191824,2846172025} and \textbf{training}~\cite{1296637108,di2020real,3888330750} contexts. Some research examines both (e.g., language learning and woodworking in the same study~\cite{1731146538}). Learning environments dominate both corpora (A: 57/73; 78\%~\cite{4019205162,86191824,1598166515}; B: 40/49; 82\%~\cite{116733479,2995141815}), with training making up a smaller portion (A: 16/73; 22\%; B: 12/49; 24\%~\cite{martinez2023lessons,3888330750}). 

A key distinction between environments is the level of physical engagement: \textbf{active} versus \textbf{stationary}. Although exceptions exist---such as embodied learning contexts where students move around the classroom and stationary training tasks like oral presentations~\cite{2634033325}---most learning environments involve seated participants interacting with a computer, classroom teacher, or each other. In contrast, training typically entails physical activity, such as movements and interactions with objects, to accomplish a task. This distinction, in turn, shapes both data capture and analysis. In active environments, motion capture, video, and physiological sensors generate complex, high-dimensional data that generally require model-based methods (e.g., deep learning). For example, \citet{vatral2023prediction} employed gradient-boosted regression trees on eye-gaze and speech features to predict nursing trainees' self-efficacy. Extending this multimodal approach, \citet{martinez2023lessons} integrated smartwatches, microphones, and positioning sensors. However, the authors noted the need for standardized, researcher-provided devices to ensure data quality---highlighting current limitations of ``bring-your-own-device'' strategies for scalable deployment.

\paragraph{Environment Interaction Setting.}

\textbf{Virtual} environments occur entirely online or in simulated spaces without physical co-presence~\cite{3339002981,3093310941,1441411748}. \textbf{Physical} environments involve in-person activity in real-world spaces such as classrooms, labs, training facilities, and clinics~\cite{1118315889,3095923626,1196965665}. \textbf{Blended} settings combine both, as in robotics courses where students program physical robots using online interfaces~\cite{261302708}. A shift is evident across corpora: virtual environments dominated Corpus~A (51/73; 70\%)~\cite{818492192,3339002981,3093310941}, while Corpus~B was led by physical settings (34/49; 69\%)~\cite{209328204,1935812764}, reflecting the post-COVID return to in-person learning.

Virtual environments are easier to scale and monitor, primarily because students tend to focus on their computer screens, which limits their movement and enables streamlined data collection through computer-based logs, screen recordings, and webcam feeds \cite{snyder2023analyzing}. For example, researchers have combined heart rate data from Fitbits with system logs to investigate PhD students' self-regulated learning \cite{dindar2020matching}. These conditions facilitate large-scale, quantitative, and model-based analyses.

However, virtual learning lacks the authenticity of physical settings, especially for K-12 education~\cite{gokccearslan2024benefits,shamir2021facilitating}. Studying learners \textit{in situ} (i.e., in their natural environment) offers critical insight into real-world learning processes. It enables access to affordances like social interactions using sensors that are not replicable online. Yet physical environments require on-site researcher presence, heightened IRB oversight, and face logistical barriers to scale \cite{alwahaby2022evidence}. Noise, technical failures, and unstructured dynamics often produce incomplete or messy datasets~\cite{cukurova2020promise}. Consequently, physical settings frequently rely on model-free methods, such as qualitative coding or statistical correlations between observed behaviors and outcomes~\cite{jiang2022empirical}.

\subsubsection{Data Collection Media (``Sensors'' in Figure~\ref{fig:framework})} \label{subsec:data_collection_mediums}

Data collection media determine which aspects of learning and training can be observed, modeled, and ultimately supported. Their selection shapes the granularity of multimodal traces, the feasibility of signal fusion, and the types of constructs that can be inferred (e.g., performance, collaboration, reflection; see the following subsections). Media can also be combined to form multimodal signals. For instance, prosodic and semantic features extracted from audio can be fused with visual cues to predict affect~\cite{poria2016fusing}. Across both corpora, we identified a common set of data collection media, summarized in Table~\ref{tab:data_collection_mediums}.

\begin{table}[htbp]
    \renewcommand{\arraystretch}{1.3}%
    \centering
    \begin{tabular}{p{0.18\linewidth}@{\hskip .09in} | @{\hskip .09in}p{0.76\linewidth}@{\hskip .09in}}
        \toprule
        \textbf{Medium} & \textbf{Definition} \\
        \midrule
        Video & Sequences of image frames captured from a camera source \cite{2836996318,1581261659,3809293172}. \\
        Audio & Audio signals captured by a microphone \cite{3093310941,85990093,957160695}. \\
        Screen Recording (Screen) & Sequences of image frames displaying a device's screen contents \cite{2456887548,3783339081,86191824}. \\
        Eye & Eye movement data and gaze points captured by tracking devices \cite{4277812050,1345598079,123412197}. \\
        Logs & Participant's actions within the system and its state data \cite{3408664396,1118315889,1886134458}. \\
        Physiological Sensors (Physical) & Specialized sensors used to gather participants' physiological data \cite{804659204,853680639,3398902089}. \\
        Interview & Structured or unstructured conversations between researchers and participants \cite{3146393211,3625722965,2609260641}. \\
        Survey & Standardized sets of questions administered to participants \cite{1374035721,4019205162,1576545447}. \\
        Participant-Produced Artifacts (PPA) & Materials produced by study participants using various mediums, including physical objects created for a task or written responses to formative assessment questions \cite{3637456466,2936220551,2634033325}. \\
        Researcher-Produced Artifacts (RPA) & Materials produced by the researchers that contribute to analysis and findings, such as observational notes \cite{853680639,3796643912,martinez-maldonado_data_2020}. \\
        Motion & Raw motion data collected via various different devices/technologies \cite{2055153191,3625722965,di2020real}. \\
        Text & Raw textual input \cite{666050348,141378338,190066185}. \\
        \bottomrule
    \end{tabular}
    \caption{Data collection media.}
    \label{tab:data_collection_mediums}
\end{table}

\begin{figure}[htbp]
    \centering
    \begin{minipage}[t]{0.48\textwidth} 
        \centering
        \includegraphics[width=\textwidth]{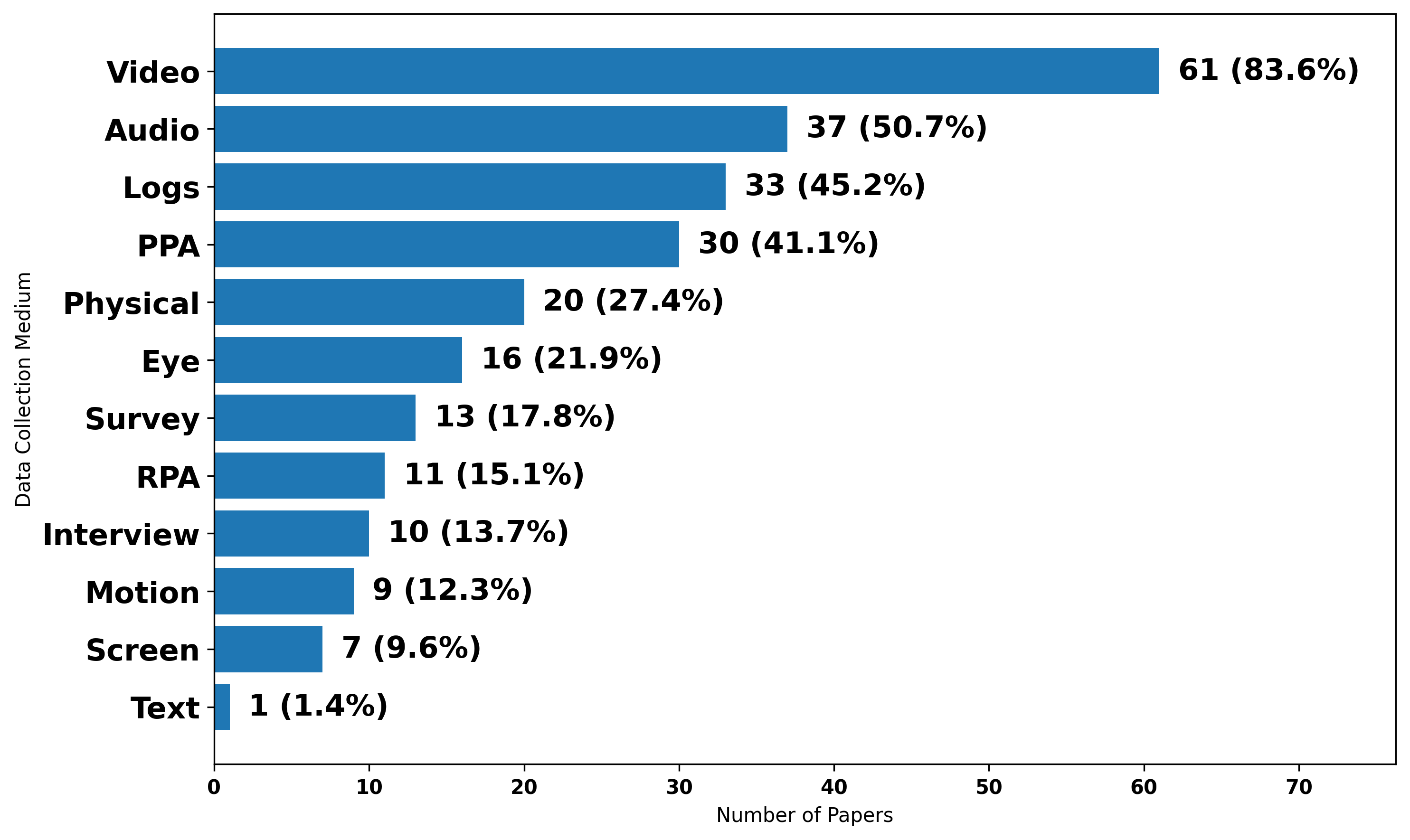}
        \caption{Corpus A data collection media distribution.}
        \label{fig:data_collection_mediums_A}
    \end{minipage}
    \hfill
    \begin{minipage}[t]{0.48\textwidth} 
        \centering
        \includegraphics[width=\textwidth]{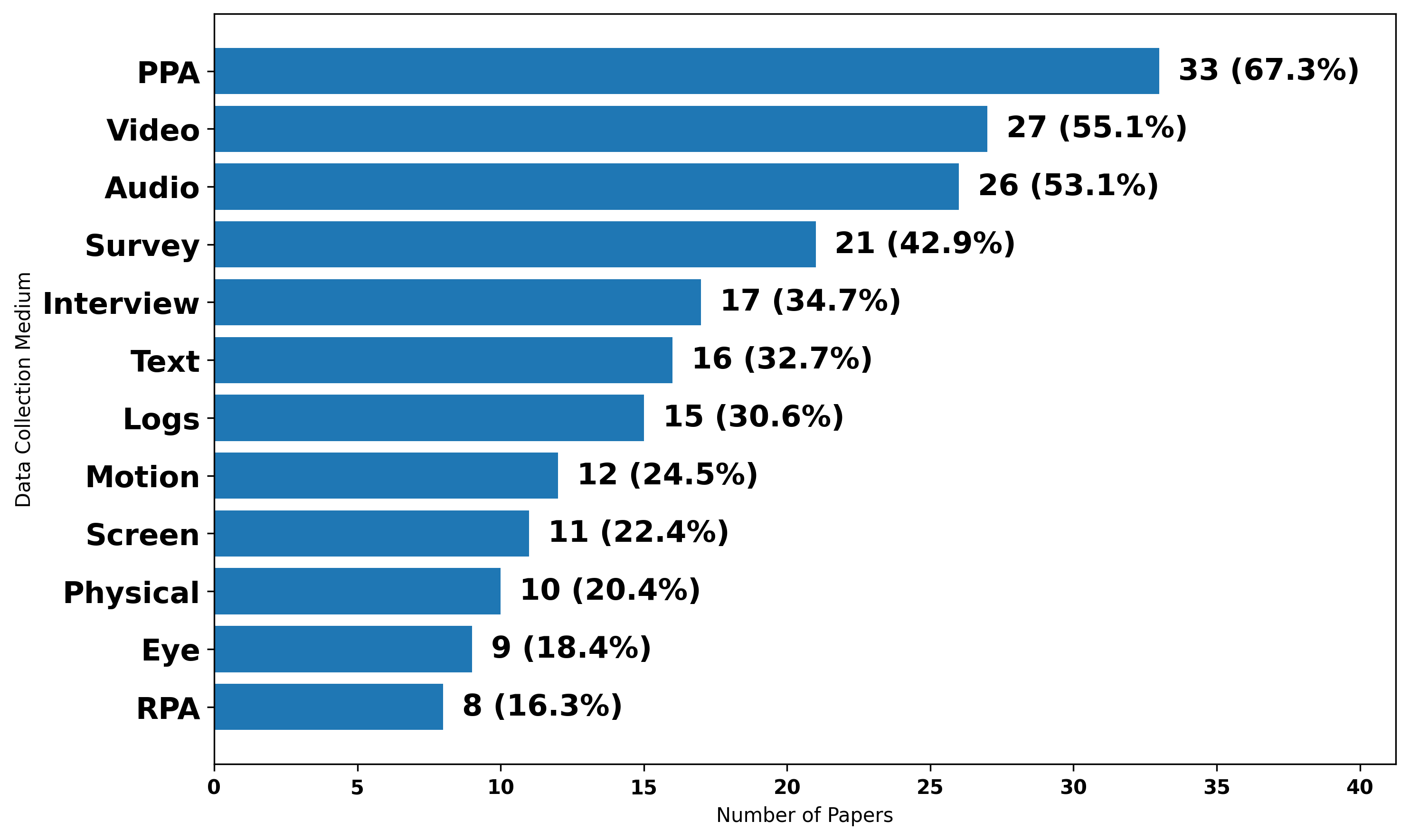}
        \caption{Corpus B data collection media distribution.}
        \label{fig:data_collection_mediums_B}
    \end{minipage}
    \label{fig:data_collection_mediums_distributions}
    \Description[Data collection media distributions.]{Data collection media distributions.}
\end{figure}

Figures~\ref{fig:data_collection_mediums_A} and~\ref{fig:data_collection_mediums_B} compare the distribution of data modalities across Corpora A and B. Video and audio data collection were prevalent across both corpora, indicating the richness and usefulness of integrating these modalities in multimodal settings. Audio data can yield prosodic information, such as tone, pitch, pauses, and volume, alongside the semantic meaning of the spoken words, which can be fused with other data streams to derive modalities downstream \cite{chango2022review,sharma2020multimodal}. Video data can be used to derive visual modalities like activity, gesture, pose, gaze, and affect \cite{blikstein2016multimodal,noroozi2019multimodal}. 

In the post-COVID era, multimodal learning and training studies experienced notable shifts. As research moved from virtual to physical settings, the use of participant-produced artifacts increased while reliance on environment log data declined. This transition, along with the rise of LLMs, enabled richer forms of textual feedback not dependent on rule-based systems derived from logs. Surveys and interviews also became more prominent, reflecting a growing emphasis on stakeholder agency in system design and validation rather than purely technological advancement~\cite{cohn2025personalizing,cohn2025exploring}. The most striking shift was in textual input: Corpus~A included only one study using raw text~\cite{666050348}, while nearly one-third of Corpus~B papers captured text as a primary data source~\cite{190066185,1441411748,2429627610,3537775194,3313249608}---almost entirely due to LLM-mediated interactions.

Research goals, target modalities, and participant interaction structures shape data collection methods. For example, studying socially shared regulation of learning requires both environment log data to capture cognitive activity and discourse data to analyze metacognitive and social processes~\cite{snyder2025using,snyder2023analyzing,cohn2024human}. In contrast, CPR tutor training relies on motion data and physiological signals (e.g., EMG, accelerometers, gyroscopes) to evaluate chest compression performance~\cite{di2020real}. 

However, some modalities pose challenges to adoption. Video and sensor-based methods raise privacy concerns and are often perceived as invasive~\cite{marin2023social,prinsloo2022answer}, while LLM-based systems may raise skepticism due to risks of hallucination, toxicity, and misuse~\cite{harvey2025don,mintz2023artificial}. Self-reported measures, such as interviews and surveys, are valuable but must typically be triangulated with other modalities to ensure reliability~\cite{ober2021linking}. Synchronizing data from multiple data streams is challenging and requires standardization, time alignment, and feature engineering. This process is often time-consuming and requires both domain knowledge and technical expertise, creating additional barriers to adoption.

\subsection{Multimodal Data} \label{subsec:multimodal_data}

Multimodal data form the foundation of MMLA systems: choices about what to capture, how to represent it, and which streams to combine determine the learner states and processes that can be modeled, the inferences that can be made about their behaviors and performance, and how analytics can inform action. In our framework, multimodal data sit at the intersection of learner behavior and analytics, linking observable activity across multiple modalities to higher-level constructs (e.g., collaboration quality, regulation of self- and group-learning, knowledge and skill acquisition) \cite{jarvela_what_2021}. 

These modality choices shape both traditional MMLA pipelines and GenAI-enabled systems by defining system architecture that determines interpretability, robustness, and analytic scope. We adopt a unified taxonomy of five \textbf{modality groups}, which partitions modalities based on how they are derived and the information they convey: (1) natural language, (2) vision, (3) physiological signals, (4) human-centered evidence, and (5) logs. Table~\ref{tab:modalities} presents each modality alongside its corresponding modality group(s), setting the stage for the subsequent sections that examine the constructs and analytic methods enabled by each category.

\begin{table}[htbp]
    \renewcommand{\arraystretch}{1.3}%
    \centering
    \begin{tabularx}{\textwidth}{p{0.18\linewidth}@{\hskip .1in} | X | p{0.162\linewidth}@{\hskip .1in}}
        \toprule
        \textbf{Modality} & \textbf{Description} & \textbf{Modality Group} \\
        \midrule
        Affect & Participant's facial expression, or emotional or affective state \cite{3408664396,1609706685,3093310941}. & NLP, Vision, Physical \\
        Pose & Participant's physical position, location, or body posture \cite{2456887548,1637690235,2181637610}. & Vision, Physical \\
        Gesture & Participant's gestures and body language \cite{818492192,3095923626,85990093}. & Vision \\
        Activity & Participant's observable actions or activities \cite{483140962,3135645357,3783339081}. & Vision, Physical \\
        Prosodic Speech (Pros. Speech) & Elements of speech beyond word meaning, e.g. volume, pauses, and intonation \cite{1118315889,3339002981,2345021698}. & NLP \\
        Transcribed Speech (Trans. Speech) & Textual speech transcribed from audio \cite{3146393211,3796180663,4019205162}. & NLP \\
        Qualitative Observations (Qual. Obs.) & Researcher observations about the participant and study task \cite{3309250332,1847468084,3398902089}. & Human-centered \\
        Logs & Participant's environment actions and system state data \cite{1886134458,2070224207,4278392816}. & Logs \\
        Gaze & Participant's eye gaze, e.g., movement, direction and focus \cite{1019093033,1581261659,1598166515}. & Vision, Physical \\
        Interview & Notes from interviews between researchers and participants \cite{86191824,3448122334,1296637108}. & Human-centered \\
        Survey & Participant's responses to surveys/questionnaires \cite{1374035721,957160695,123412197}. & Human-centered \\
        Pulse & The participant's pulse, indicating their heart rate \cite{3660066725,3856280479,433919853}. & Physical \\
        EDA & Participant's electrodermal activity \cite{205660768,147203129,2000036002}. & Physical \\
        Temperature (Temp.) & Participant's body temperature \cite{2000036002,1877483551,123412197}. & Physical \\
        Blood Pressure (BP) & Participant's blood pressure \cite{123412197,3856280479,433919853}. & Physical \\
        EEG & Participant's electroencephalography activity \cite{4278392816,2000036002,123412197}. & Physical \\
        Fatigue & The level of fatigue experienced during the activity \cite{3660066725,3856280479}. & Vision, Physical \\
        EMG & Participant's electromyography activity \cite{di2020real,1763513559}. & Physical \\
        Participant Produced Artifacts (PPA) & Artifacts produced by the participant during the study, e.g., pre/post-tests \cite{2634033325,4277812050,2155422499}. & Human-centered \\
        Researcher Produced Artifacts (RPA) & Artifacts produced by the researcher about the study and participants, e.g., field notes \cite{3809293172,4035649049,2609260641}. & Human-centered \\
        Spectrogram (Spect.) & Representation of audio frequencies in the form of a spectrogram \cite{3754172825}. & NLP \\
        Text & Participant's raw text data generated in the study environment \cite{666050348}. & NLP \\
        Pixel & RGB pixel values from cameras or sensors \cite{3135645357}. & Vision \\
        \bottomrule
    \end{tabularx}
    \caption{Modalities, their definitions, and the modality groups they fall into (detailed in Section~\ref{subsec:multimodal_data}).}
    \label{tab:modalities}
\end{table}

\begin{figure}[htbp]
    \centering
    \begin{minipage}[t]{0.48\textwidth} 
        \centering
        \includegraphics[width=\textwidth]{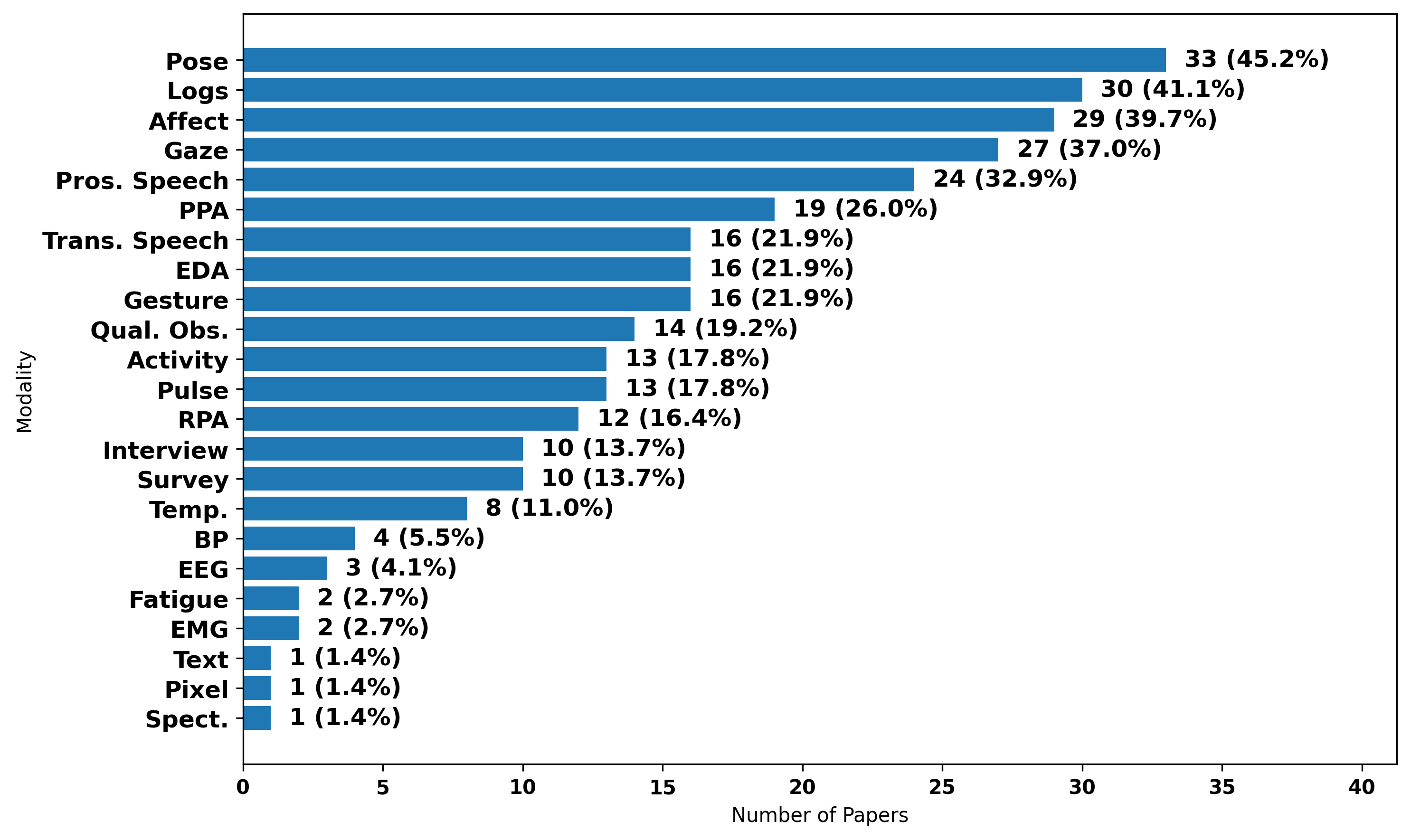}
        \caption{Corpus A modalities distribution.}
        \label{fig:modalities_A}
    \end{minipage}
    \hfill
    \begin{minipage}[t]{0.48\textwidth} 
        \centering
        \includegraphics[width=\textwidth]{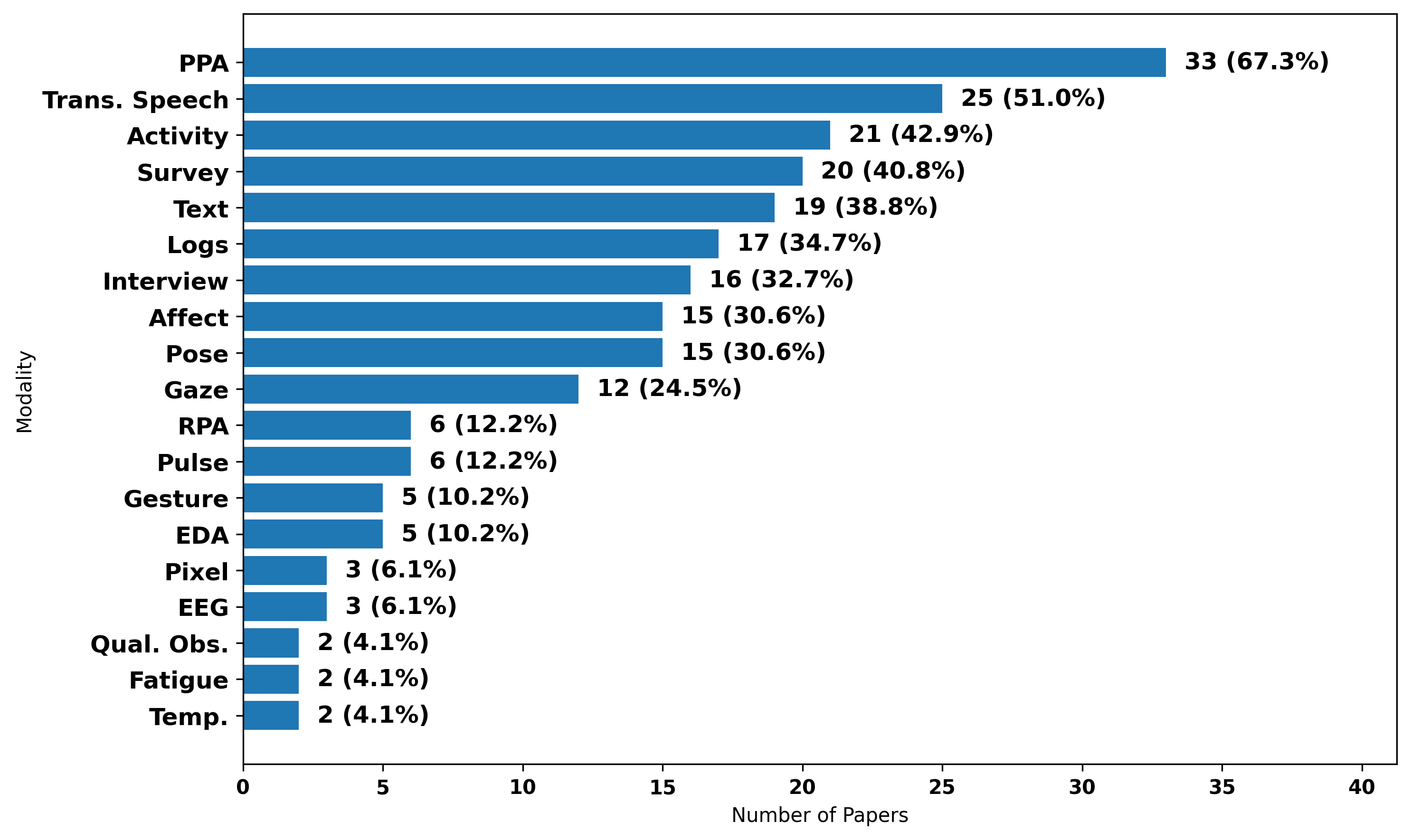}
        \caption{Corpus B modalities distribution.}
        \label{fig:modalities_B}
    \end{minipage}
    \label{fig:modalities_distributions}
    \Description[Modality distributions.]{Modality distributions.}
\end{figure}

Figures~\ref{fig:modalities_A} and~\ref{fig:modalities_B} show the modality distributions for Corpora A and B, largely reflecting trends in data collection media. Prior to the development of LLMs, COVID-era studies (Corpus A) focused on individual modalities such as pose, logs, affect, gaze, and prosodic speech. These modalities were often collected in virtual settings using tools like microphones, webcams, and trace data.

In contrast, Corpus~B reflects a shift toward in-person, multi-party, human-centered studies integrating LLMs and sensor-rich physical environments. This newer corpus emphasizes participant-produced artifacts, transcribed speech, physical activity, surveys, and raw textual inputs, captured through 3-D video, lapel microphones, and student-created materials. Several modalities present in Corpus~A---such as prosodic speech, spectrogram, EMG, and blood pressure---are notably absent in Corpus~B, underscoring a broader shift from physiological signal-based approaches toward artifact- and LLM-centric methods. Importantly, many post-LLM systems employ large language models not as end-to-end multimodal architectures, but as analytic or interpretive layers operating on representations extracted from other modalities.

Approximately two-thirds of papers in both corpora use 3--5 distinct modalities to guide their research (A: 49/73 \cite{1609706685, 2456887548}; 67\%, B: 34/49 \cite{328477558,190066185}; 69\%). This reflects a methodological balance: the number is sufficient to enable triangulated inferences across heterogeneous signals (e.g., aligning logs with gaze, or artifacts with surveys), while remaining tractable in terms of data collection, synchronization, and model complexity. This norm becomes especially relevant when incorporating more sophisticated analytic systems, such as LLM-based agents, which must operate within practical limits on data richness and annotation effort. The following subsections detail how the modalities within each modality group are operationalized in practice.

\subsubsection{Natural Language}\label{subsubsec:language}

Natural language captures how learners and trainees speak, write, and interact with peers, instructors---and increasingly, LLM-based systems---across modalities such as prosodic and transcribed speech, raw text, and affect derived from language or speech (see Table~\ref{tab:modalities}). Because much teaching, collaboration, feedback, and assessment are inherently language-based, NLP signals often encode rich information about learners' metacognition (e.g., goal setting \cite{cohn2025theory}, planning \cite{fonteles2024aied}, and reflective behaviors \cite{snyder2023analyzing}) as well as collaborative processes such as information pooling and consensus building \cite{snyder2024investigating}.

Natural language is also frequently used to contextualize other modalities, including gaze, posture, and interaction logs. For example, \citet{snyder2025using} employed Markov modeling to infer students' metacognitive states (planning, enacting, monitoring, and reflecting) during collaborative problem solving by integrating environment log data and collaborative discourse, enabling ChatGPT-generated summaries of collaboration to be grounded in students' actions within the learning environment. \citet{zhou2024detecting} used video and conversation data to automatically detect gaze, nonverbal speech, and resource-management behaviors during group learning. Analysis revealed distinctive interaction patterns, including loops between gaze-linked referring/following and resource-management behaviors. These patterns differentiated groups based on their shared understanding and collaborative learning outcomes.

The use of NLP has skyrocketed in recent years, increasing from 35/73 papers (48\%) in Corpus~A---where prosodic speech was the primary NLP modality---to 40/49 (82\%) in Corpus~B~\cite{962997360, 1161441004}---where text and transcribed speech were the primary modalities. This shift is closely tied to studies that deploy LLM-enabled systems, such as GPT-based learning aids~\cite{1161441004}, GenAI-supported multimodal composing~\cite{780281159}, and multimodal LLMs for assessment~\cite{3224774131}. Natural language features such as raw text, word embeddings, term frequency, loudness, and pitch are consistently reported as informative, with strong associations with predictive outcomes such as learner productivity, performance, and collaborative confusion or conflict. Collaborative settings, in particular, highlight these features as frequently among the most predictive, especially when combined with other multimodal signals~\cite{cohn2025personalizing}.

While NLP outputs across Corpora~A and~B are broadly similar, their methodological approaches differ substantially. Corpus~A predominantly uses traditional machine learning techniques such as SVMs~\cite{3754172825,85990093,1637690235} and logistic regression~\cite{957160695,3796180663,1576545447}, often supplemented by qualitative methods (e.g., transcript coding, case studies) to interpret learner discourse and interaction~\cite{2497456347,1847468084,1296637108}. In contrast, Corpus~B centers on LLM-enabled pipelines, particularly for analyzing dialogue and delivering feedback~\cite{3224774131,3313249608,3537775194}. These systems enable free-flowing conversations that were not feasible prior to the LLM era. In some cases, multimodal LLMs are also used for assessment~\cite{3224774131}. Here, LLMs and GenAI tools serve dual roles: they act as interactive components within the learning environment and as generators of textual data for downstream multimodal analysis, with other modality groups (e.g., vision, logs, physiological signals) offering contextual signals around these language-based interactions.

The natural language challenges faced by MMLA researchers also differ considerably between corpora. Corpus~A, with its reliance on traditional machine learning, frequently cites issues such as small, imbalanced datasets~\cite{957160695,3783339081,3796180663,cochran2022improving,cochran2023improving,cochran2023bimproving}, which hinder the training and adaptation of deep learning and transformer models~\cite{cohn2020bert,cochran2023using,957160695,3796643912,3448122334}. The corpus's emphasis on audio data introduces additional challenges, particularly the time-intensive nature of feature extraction. Tools such as NLTK~\cite{nltk}, openSMILE~\cite{eyben2010opensmile}, and TAACO~\cite{crossley2016tool,crossley2019tool} can aid this process, but often yield large and difficult-to-interpret feature spaces~\cite{3135645357}, while manual preprocessing and feature engineering further constrain scalability~\cite{32184286,3796180663}. Conversational agents are rare in Corpus~A; when used, they typically deliver static messages or summative feedback rather than engaging in multi-turn interaction~\cite{3093310941,3783339081,1576545447}. While using raw text with LLMs (like in Corpus B) mitigates many of these issues, others arise such as transcription errors from automatic speech recognition (ASR) in noisy classrooms~\cite{1118315889,666050348,32184286} and concerns surrounding LLMs like adverse interactions with students and how to effectively evaluate LLM output \cite{jiao2025llms, chang2024survey}.

\subsubsection{Vision}\label{subsubsec:vision}

Vision-based modalities offer continuous insight into how learners move, attend, react, and interact in learning and training environments. Cameras and eye trackers capture signals such as affect, pose, gesture, activity, gaze, and fatigue\footnote{We define \textbf{pixel} data in Table~\ref{tab:modalities}, but omit further discussion due to its limited use across both corpora.}, which are often inaccessible through logs or language alone. These visual signals are critical for modeling non-verbal behavior and engagement. In MMLA, they support both model-based approaches (e.g., convolutional neural networks~\cite{1637690235}) and qualitative methods (e.g., interaction analysis~\cite{780281159}), and are frequently triangulated with other modalities such as logs and speech~\cite{780281159}.

The rise of multimodal LLMs such as GPT~\cite{achiam2023gpt} and Gemini~\cite{team2023gemini} has broadened the role of vision-language models (VLMs) in MMLA, shifting their use from traditional tasks like classification and coding to more complex applications in interpretation and sense-making. For example, \citet{yan2024vizchat} use GPT-4V to interpret screenshots of nursing students' learning analytics dashboards, enabling the system to ``see'' and reason about visual elements such as charts and graphs. This visual understanding is integrated with retrieval-augmented generation (RAG), enabling the chatbot to produce explanations grounded in both the dashboard's visual structure and its educational data context. Unlike traditional machine learning pipelines, VLM-based multimodal integration requires minimal feature engineering. While neural networks often require pixel data to be normalized or transformed into visual embeddings aligned with textual representations \cite{huang2020pixel}, VLMs can directly accept raw text and image inputs from the user.

There has been a marked shift away from vision modalities in recent years. Whereas three of Corpus A's top four modalities were vision-based (pose \cite{3339002981}, affect \cite{1847468084}, and gaze \cite{2497456347}), none appears among the top seven in Corpus B. Except for the activity modality, the percentage of papers using each vision modality declined\footnote{We ignore the fatigue modality due to underrepresentation: one paper in Corpus A and two in Corpus B.} (see Figures~\ref{fig:modalities_A} and~\ref{fig:modalities_B}). Overall, vision-focused papers fell from 59/73 (81\%) in Corpus A to 27/49 (55\%) in Corpus B. We hypothesize that the rise in activity papers (A: 13/73; 18\% \cite{3660066725,1296637108}, B: 21/49; 43\% \cite{425012016,780281159}) reflects a return to in-person research after COVID.

Vision modality group implementations were broadly consistent across corpora, favoring quantitative, model-based analyses to classify attributes such as pose~\cite{1118315889, 2429627610}, gaze~\cite{3339002981, 3313249608}, and affect~\cite{1374035721, 1844320601}. These modalities were typically used in traditional machine learning or shallow deep learning pipelines to infer learner states such as engagement, collaboration quality, or skill level, often supplemented by qualitative interpretation of predicted classes \cite{1500258376, 2005607968}. Vision data functioned both as features (e.g., using gaze to predict engagement~\cite{1935812764}) and as prediction targets (e.g., deriving affect from image data~\cite{1500258376}). In Corpus~B, vision was often one component of larger multimodal pipelines involving sensors and logs~\cite{261302708}, providing contextual grounding for downstream analytics or dashboards~\cite{cohn2024multimodal}. While LLM usage increased overall, vision integration with multimodal LLMs was rare; instead, vision inputs like gaze were used to inform or condition LLM-based interactions (e.g., as context for ChatGPT~\cite{1844320601}).

As with natural language, vision-based modalities often contributed significant predictive power to multimodal pipelines. For instance, \citet{acosta2024multimodal} demonstrated that integrating trace log data with vision features such as facial action units, head pose, and gaze---extracted using OpenFace~\cite{baltruvsaitis2016openface}---led to more accurate predictions of collaborative satisfaction than any individual modality or subset. Student survey responses served as ground truth, and two specific facial action units emerged as the most predictive features across both high- and low-performing modality combinations. Similarly, \citet{3754172825} used early fusion to combine video features (e.g., facial expressions, body movements, inter-learner distance), linguistic features (text embeddings), and audio signals (e.g., speaking time, pitch) to predict \textit{impasse}, i.e., moments of stalled progress during collaborative problem solving due to conflicting ideas. Their results highlighted facial muscle movements as particularly strong predictors of impasse, underscoring the importance of visual signals in capturing nuanced social dynamics.

However, vision-based multimodal analytics face several practical and methodological challenges. Many learning and training environments lack controlled lighting, fixed camera setups, or specialized hardware (e.g., eye trackers), limiting the feasibility of fine-grained gaze or pose analysis~\cite{ashwin2025challenges,colyer2018review,worsley2012multimodal}. Additionally, small and noisy datasets often lead researchers to rely on pre-extracted features rather than raw pixel data, which can obscure model assumptions and reduce adaptability across tasks or domains. For instance, commercial tools like iMotions provide real-time emotion tracking from facial muscle movements. Yet, the inferred states (e.g., joy, anger, fear) are typically assumed as ground truth without independent validation. Synchronizing and fusing visual data with other modalities, such as natural language, logs, or physiological signals, remains complex and time-consuming, with missing data and differing temporal resolutions further complicating joint modeling. Additionally, there is growing concern that opaque vision components may introduce bias or misinterpret learner behaviors, particularly for underrepresented populations or non-standard learning contexts~\cite{ashwin2025challenges}.

\subsubsection{Physiological Signals}\label{subsubsec:sensors}

Physiological signal-based modalities capture learners' physiological and motion-related traces, including affect, pose, activity, gaze, pulse, electrodermal activity (EDA), body temperature, blood pressure, electroencephalography (EEG), fatigue, and electromyography (EMG). These modalities link learners' observable actions with their internal states, enabling the interpretation of engagement, cognitive load, stress levels, and coordination in both learning and training contexts. Unlike vision-based data, physiological signal modalities are typically used as primary features in predictive models rather than as target outputs.

Physiological signal modalities appear more frequently in Corpus~B (23/49; 47\%~\cite{4089325423, 2166765216}) than Corpus~A (A: 20/73; 27\%~\cite{205660768, 1609706685}), with notable differences in how they are deployed. In Corpus~A, physiological signals are primarily tied to the EDA modality (16/20; 80\%~\cite{3095923626, 4278392816}), and are disproportionately used in training contexts. Although training constitutes 22\% (16/73) of Corpus A studies, it accounts for 40\% of those using physiological signals (8/20)~\cite{1609706685, martinez-maldonado_data_2020}). In contrast, physiological signal use in Corpus~B shifts toward motion-oriented modalities like activity and pose---each of which appears in 12/23 (52\%) of physiological signal-enabled studies ~\cite{261302708, martinez2023lessons}. 

In practice, this leads to different methodological approaches across the corpora. In Corpus A, physiological signals are commonly used for predictive modeling and to examine their relationships with learning behaviors and outcomes. Signal processing converts raw data streams (such as EDA, pulse, and accelerometer readings) into interpretable metrics, including learning gains, team dynamics, and shared arousal. These physiological signals support offline analyses, such as identifying patterns associated with performance, and in-time feedback during training, enabling timely interventions. Additionally, these features are often integrated with other modalities (such as visual data, logs, and language) to provide context for engagement, stress, and coordination.

Corpus B emphasizes interaction-rich environments, such as simulations and collaborative tasks, helping to capture arousal and cognitive load during the learning process. While the methods employed may vary, the physiological signals in Corpus B more clearly bridge the gap between ``\textit{vision-like}'' behavioral traces that focus on students' observable actions and their cognitive and emotional states, thereby reinforcing their integrative role in connecting what learners are doing, how they move, and how their bodies respond \cite{nguyen2015don}.

The wide range of modalities derived from physiological and motion-based sensors has yielded diverse and insightful findings across multimodal learning and training pipelines. For example, \citet{que2025using} combined gaze data from EyeLink 1000 Plus eye trackers with heart rate, inter-beat intervals, and electrodermal activity from an Empatica E4 wristband to predict three types of cognitive load during an English as a Second Language (ESL) reading task: \textit{extraneous load} (avoiding irrelevant information while learning), \textit{intrinsic load} (reflecting the complexity of learning material), and \textit{germane load} (resources available for processing intrinsic load, e.g., comprehension). They found that extraneous load was predicted by increased fixation count and lower mean heart rate; intrinsic load by increased fixation count and mean saccade amplitude; and germane load by increased fixation count and heart rate variability.

While the above example is illustrative, many other studies demonstrate that physiological signals do not yield generalizable findings across contexts: different sensors work best in different environments, for different tasks, and with different populations of learners and trainees. Moreover, integrating and interpreting heterogeneous sensor streams remains technically challenging, and reliance on specialized hardware, such as eye trackers and wristbands, raises practical concerns about cost, scalability, and privacy. Teachers and students have also emphasized the importance of understanding how machine learning models generate predictions~\cite{cohn2024chain}, highlighting the need for interpretable, human-centered, explainable AI (XAI) approaches when using physiological signals. Without these, stakeholders may struggle to trust or act on insights derived from these modalities~\cite{schoonderwoerd2021human,rojat2021explainable}.

\subsubsection{Human-Centered}\label{subsubsec:human}

Human-centered modalities include qualitative observations~\cite{1469065963, 2995141815}, interviews~\cite{2609260641, 3537775194}, surveys~\cite{518268671, 1285699194}, and artifacts produced by participants~\cite{1196965665, 2634033325} or researchers~\cite{853680639, 2737776963}. These modalities anchor multimodal learning and training analytics in the lived experiences of learners and the perspectives of educators and researchers, offering insights into how participants perceive, interpret, and reflect on tasks---insights that are often inaccessible through sensor or log-based data streams alone. They are frequently used to complement quantitative findings with rich detail (e.g., via case studies or error analyses), and are often treated as ground truth in predictive modeling or for correlating learning behaviors with outcomes. Human-centered data are crucial for validating inferences from other modalities, improving the interpretability of model outputs for stakeholders, and understanding how multimodal systems are experienced and perceived in practice.

While both corpora incorporated human-centered data, its prevalence rose sharply from Corpus~A to Corpus~B (A: 45/73; 61\%~\cite{1598166515,3308658121}, B: 46/49; 94\%~\cite{2166765216,martinez2023lessons}), reflecting the community's growing emphasis on stakeholder agency in the design and evaluation of intelligent systems. In earlier work, methodological studies in educational AI often lacked input from teachers, students, or learning scientists~\cite{cochran2023improving}. With the emergence of LLMs, however, user-centered approaches, such as participatory design and co-design~\cite{hutchins2024co,sarmiento2022participatory}, have become critical, as trust in GenAI systems hinges on their perceived safety, effectiveness, and alignment with stakeholder needs \cite{cohn2025theory}. Across both corpora, participant-produced artifacts were the most frequent human-centered modality, with a stronger emphasis in Corpus~B (33/46; 72\%)~\cite{3313249608,141378338} versus A (19/45; 42\%)~\cite{205660768,1886134458}). 

Human-centered data are most often used for model-free, qualitative analysis, but it can also be annotated for quantitative purposes. One noteworthy approach in multimodal learning and training research is \textit{quantitative ethnography}~\cite{shaffer2017quantitative}, which involves applying qualitative coding to data and then extracting quantitative features for analysis. This enables the study of complex human behavior through techniques such as network analysis. For example, \citet{sung2024beyond} employed multimodal \textit{epistemic network analysis\footnote{ENA transforms coded qualitative data into visual networks that reveal how coded concepts co-occur over time. Data segments (e.g., collaborative discourse turns) are coded according to a theoretical framework; nodes represent codes, and edges reflect their co-occurrence within a pre-defined window. Edge thickness encodes co-occurrence frequency, enabling temporal comparisons across groups and assessment of learning processes.}} (ENA)~\cite{shaffer2016tutorial} during a guided reading study in a college biology course to examine how sequences of students' self-regulated learning behaviors differed between mastery and non-mastery groups. Think-aloud data was analyzed to identify self-regulation strategies (e.g., monitoring, assessing, summarizing), while environment log data was used to differentiate between in-class and out-of-class engagement with guided reading questions. Quiz scores served as measures of learning outcomes. In both groups, monitoring-related verbalizations frequently co-occurred with other self-regulated learning codes; however, the co-occurring codes differed by group: in the mastery group, monitoring was more often paired with domain-specific strategies, whereas in the non-mastery group, it was more often paired with domain-general strategies. This discrepancy led the authors to conduct follow-up qualitative analyses to better understand the contextual nuances of the students' learning processes.

The challenges surrounding human-centered modalities stem largely from the inherent subjectivity of qualitative data and analysis. Observations, participant-produced artifacts, and self-reported measures (e.g., surveys and interviews) can introduce coder bias as well as cultural and linguistic biases~\cite{mehra2015bias,noble2015bias}, which may propagate into downstream models and compromise generalizability. Furthermore, the manual processes involved in data collection, coding, and interpretation are labor-intensive and difficult to scale. Compounding these challenges is the limited standardization of coding schemes across studies, particularly for artifacts, interviews, and observational data, which hinders replication and cross-study comparison.

\subsubsection{Logs}\label{subsubsec:logs}

Environment logs capture learners' and trainees' interactions with digital tools, platforms, and learning environments. In MMLA, these time-stamped traces (e.g., clicks, navigation, tool use) provide a behavioral record that can be aligned with other modalities to infer cognitive strategies, engagement, and progress in problem-solving. While modalities like language or vision may reveal what participants are thinking or feeling, log data indicate what they are actually doing. These streams are highly integrative, often providing context for interpreting focal modalities such as natural language and vision. Log data are used similarly across both corpora in terms of frequency and methodology (A: 30/73; 40\% \cite{2456887548,123412197}, B: 17/49; 35\% \cite{190066185,cohn2024multimodal}).

Studies often use log data in supervised machine learning contexts, extracting features such as mouse clicks, click frequencies, click sequences, and inactivity (i.e., no clicks) as inputs for models like logistic regression, support vector machines (SVM), and random forests to predict outcomes like task performance and engagement~\cite{147203129,483140962,1019093033}. Statistical analyses are also common, linking log-derived metrics to learning gains and behavioral patterns, for example, by mining clickstream sequences to infer cognitive strategies such as constructing, debugging, and assessing~\cite{snyder2024understanding}.

LLMs have broadened the interpretive scope of log data by enabling its direct integration into prompts as contextualized natural language. For example, \citet{fonteles2026JLI} used late fusion in an embodied learning setting: gaze, speech, and log data were first classified with unimodal deep learning models, then combined as text-based input to an LLM to interpret students' socially shared regulation. Similarly, \citet{cohn2025personalizing} translated students' block-based programming actions into natural language to contextualize collaborative discourse during RAG with an LLM-based pedagogical agent. Incorporating log data to situate discourse within the learning environment improved semantic alignment and retrieval performance relative to using discourse alone. Students also reported positive interactions with the agent, indicating its potential to enhance engagement and support in collaborative learning.

The main drawback of environmental log data is that it is usually only applicable in digital settings, such as virtual or blended environments, but not in fully physical ones. Time alignment and temporal granularity present challenges for multimodal fusion. Researchers often need to reconcile different sampling rates, synchronize events across various modalities, and manage high-dimensional time series. Additionally, log-based models frequently struggle to generalize across different systems, partly due to the limited adoption of interoperability standards, such as xAPI and LMS-based\footnote{Learning Management System} logging. The engineering costs can also be quite high, as building robust logging infrastructures and analysis pipelines demands significant software development effort, which can impede both reuse and scalability.

\subsection{Learning Analytics} \label{subsec:learning_analytics}

Learning analytics involves transforming data into actionable insights to better understand how students learn and train. This module connects the diverse data streams generated during learning and training activities with the inferences researchers draw to understand learner behavior and deliver more effective feedback. It consists of two main components: \textbf{data fusion}, which focuses on integrating diverse data streams, and \textbf{analysis}, which centers on interpreting this data. In the following subsections, we will explore both components in detail.

\subsubsection{Data Fusion}
\label{subsec:data_fusion_insights}

Data fusion is essential for leveraging multiple data sources to enhance our understanding of learning and training. Only through fusion can we construct unified representations of learners and trainees that surpass the explanatory power of unimodal approaches. Just as humans rely on multiple senses to understand the world, data fusion allows researchers to integrate diverse modalities to better capture the conditions under which learners struggle, improve, and progress.

The conventional classification of fusion methods in MMLA, as defined by \citet{chango2022review}, includes three types: \textit{early, late}, and \textit{hybrid} fusion. Early (feature-level) fusion merges raw data from different sources at the initial processing stage. While it captures inter-modal interactions effectively, it faces challenges related to data heterogeneity and model complexity. Late (decision-level) fusion processes each modality independently before integrating results, enabling modality-specific insights but often overlooking inter-modal dynamics. Hybrid fusion blends these approaches, fusing data at multiple stages to exploit both inter-modal synergies and unimodal depth. However, this increases pipeline complexity and requires careful feature selection and synchronization.

We argue that this three-way classification does not adequately reflect the complexity of modern multimodal analysis. Our review revealed difficulties in categorizing fusion practices due to inconsistencies in defining ``\textit{raw}'' versus ``\textit{processed}'' features. For example, skeletal joint position data from a Microsoft Kinect may be considered raw by some since it is directly available from the device, but processed by others, since it is computed internally by the Kinect system from raw depth data.

\begin{figure}[htbp]
    \centering
    \includegraphics[width=0.7\textwidth]{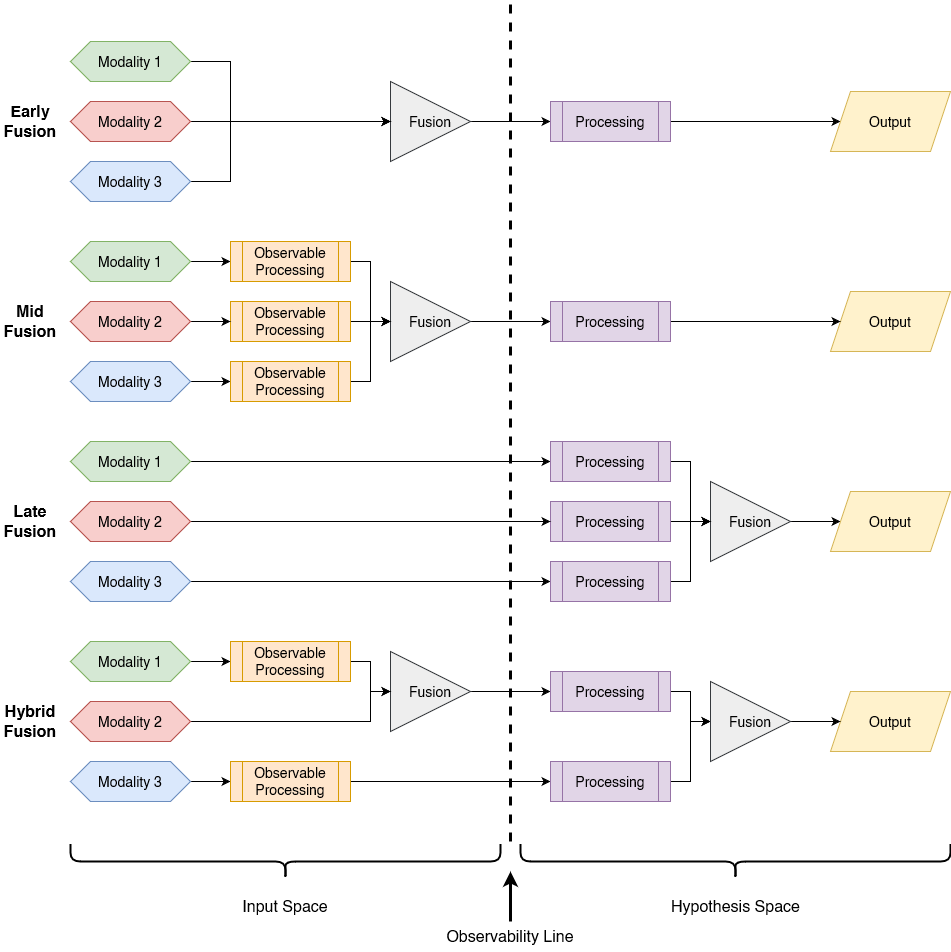}
    \caption{Multimodal data fusion scheme according to when fusion is performed relative to the observability line.}
    \label{fig:DataFusion}
    \Description[Multimodal data fusion scheme]{Multimodal data fusion scheme depending on when fusion is performed.}
\end{figure}

To resolve such ambiguity, we adopt and formalize the notion of \textit{mid fusion}, drawing from the concept of the \textit{observability line} proposed by \citet{di2018signals} that separates the \textit{input space} (i.e., observable evidence) from the \textit{hypothesis space} (i.e., inferred constructs). While the authors note that the boundary between observable and unobservable features is conceptual and context-dependent, we use this distinction to define four fusion categories that are summarized in Table~\ref{tab:data-fusion-categories} and illustrated in Figure~\ref{fig:DataFusion}.

\begin{table}[htbp]
    \renewcommand{\arraystretch}{1.3}%
    \centering
    \begin{tabular}{p{0.12\linewidth}@{\hskip .1in} | @{\hskip .1in}p{0.78\linewidth}@{\hskip .1in}}
        \toprule
        \textbf{Category} & \textbf{Description} \\
        \midrule
        Early Fusion & Draws inferences and computes analytics from multiple sources of raw, directly observable data at the earliest stage of processing before any modality-specific analysis  \cite{3095923626,205660768,1315379489}. \\
        Mid Fusion & Represents a compromise that mixes early and late fusion for analysis by combining processed, observable features generated from individual sources with analysis using other sources of data within the input space \cite{1576545447,1581261659,1598166515}. \\
        Late Fusion & Analysis is performed on individual modalities, and the inferences (abstracted and unobservable) are combined to generate outcomes at a later stage, i.e., in the hypothesis space \cite{3408664396,2836996318,2497456347}. \\
        Hybrid Fusion & Combines the strengths of both early and late fusion methods. Data from various sources are combined at multiple stages of processing \cite{818492192,2456887548,3135645357}. \\
        Other & Studies that do not fit into the early, mid, late, or hybrid categories, or where the fusion point was not specified, fusion was not performed, or fusion was performed qualitatively through observation \cite{1847468084,3398902089,86191824}. \\
        \bottomrule
    \end{tabular}
    \caption{Data fusion approaches.}
    \label{tab:data-fusion-categories}
\end{table}

For instance, Kinect raw pixel or depth data fit early fusion; skeletal joint position data---processed but still observable---fit mid fusion. We do not consider standard feature engineering practices, such as normalization, standardization, binning, or one-hot encoding, as constituting ``processed'' data for the purposes of our taxonomy. Data subjected only to these transformations would still fall under early fusion, and inferred constructs that are not observable, like planning and motivation, align with late fusion. Although the boundary between categories remains flexible, introducing mid fusion helps clarify methodological ambiguity and highlights distinctions among MMLA sub-communities in terms of their fusion practices. 

Fusion strategies demonstrated a notable consistency across both corpora, with mid fusion being the most commonly used approach. In Corpus A, 27 out of 73 papers (37\%) utilized mid fusion~\cite{433919853, 483140962}, while in Corpus B, 19 out of 49 papers (39\%) followed suit~\cite{227355655, 1040787959}. This indicates a tendency to process features before integrating modalities. Hybrid fusion was also present in both corpora, though it appeared less frequently in Corpus B (Corpus A: 19/73; 26\%~\cite{1019093033, 1426267857}; Corpus B: 8/49; 16\%~\cite{2668965770, 2737776963}). In contrast, standalone instances of early or late fusion---outside of hybrid contexts---were rare in both datasets. It is important to note that a significant portion of the papers either did not perform fusion, did not disclose their fusion methodology, or employed alternative non-canonical strategies (e.g., ``qualitative'' fusion, which involves considering multiple modalities simultaneously during qualitative analysis). These instances accounted for 27\% of papers in Corpus A (20/73), and 31\% in Corpus B (15/49)~\cite{martinez-maldonado_data_2020, 1345598079}.

The ways in which fusion was implemented varied widely across studies. Mid fusion was particularly common in settings that used devices such as the Microsoft Kinect. Hybrid pipelines were often preferred in studies that incorporated three or more modalities, likely because of their flexibility in handling complex, heterogeneous data. No discernible pattern emerged between fusion type and the nature of input or output variables: both mid and hybrid fusion approaches were used to combine input features such as discourse embeddings, prosody, affect, behavior traces from physiological sensors, and log data. These combinations were frequently used to predict collaboration and learning quality, or to support students and teachers through real-time feedback and multimodal dashboards~\cite{151988148,1278817005,1731146538}. 

While less common, fusion with multimodal LLMs was explored in Corpus~B to enable end-to-end interpretation of complex, multimodal artifacts. For example, \citet{whitehead2025utilizing} used GPT-4o to annotate students' posture during collaborative physics tasks by fusing cropped video frames with expert-defined textual prompts and a coding scheme. Fusion occurred at inference time within the model, which produced categorical posture annotations (e.g., sitting, leaning) as tabular outputs for downstream analysis. Results showed high test-retest reliability and strong agreement with human raters for simpler behaviors, though accuracy declined for more context-dependent postures. Additionally, the authors noted that performance in this context is heavily reliant on data quality, ``careful prompt engineering,'' and human validation.

Several non-LLM challenges related to fusion emerge as well---perhaps none more significant than the alignment, integration, and deployment of heterogeneous data sources in real-world settings (i.e., cross-modal interaction). These challenges include reconciling disparate sampling rates, addressing inconsistent data quality, and managing missing values, all of which complicate synchronization and modeling. Fusion pipelines often demand extensive preprocessing, manual calibration, and domain expertise to ensure that signals are both temporally aligned and semantically coherent---requirements that are especially difficult to fulfill in real-time or online learning environments. Consequently, despite methodological advancements, the practical barriers to achieving robust, generalizable fusion remain a central bottleneck for MMLA research and its broader implementation.

\subsubsection{Analysis} \label{subsec:analysis_approaches_and_methods}

Analysis is how researchers transform multimodal traces into evidence about learning and training. The research questions determine which forms of analysis are appropriate (e.g., supervised vs. unsupervised, qualitative vs. quantitative, temporal vs. static), depending on the types of insights researchers hope to gain. We classify \textbf{analysis approaches} as either \textbf{model-based} or \textbf{model-free}. Model-based analysis relies on formal models to uncover the underlying structure of the data and the interrelationships between variables. These models often involve mathematical formulations, such as machine learning functions, or computational simulations that encode theoretical assumptions about learning processes. In contrast, model-free approaches avoid such assumptions, instead using empirical statistics (e.g., correlations) or qualitative analyses to identify patterns and relationships directly from the data. 

Similarly, we use the term \textbf{analysis method} to refer to the specific techniques employed to derive insights from multimodal data in learning and training contexts. These methods, which are summarized in Table~\ref{tab:analysis_methods}, range from supervised and unsupervised machine learning (e.g., classification, clustering) to qualitative approaches and network-based analyses. It is important to note that there is no one-to-one mapping between analysis approaches and methods, as both model-based and model-free approaches can employ a variety of methods.

\begin{table}[htbp]
    \renewcommand{\arraystretch}{1.3}%
    \centering
    \begin{tabular}{p{0.155\linewidth}@{\hskip .1in} | @{\hskip .1in}p{0.78\linewidth}@{\hskip .1in}}
        \toprule
        \textbf{Method} & \textbf{Definition} \\
        \midrule
        Classification & Assigning pre-defined labels to input data based on feature analysis through supervised learning (often via deep learning approaches) \cite{3408664396,3339002981,2456887548}. \\
        Regression & Predicting continuous numerical values through supervised learning to understand input-output relationships \cite{1118315889,1609706685,2836996318}. \\
        Clustering & Grouping data based on patterns or similarities using unsupervised learning \cite{818492192,1326191931,3809293172}. \\
        Qualitative & Manually examining and interpreting data to uncover patterns or themes \cite{1847468084,3398902089,86191824}. \\
        Statistical & Using statistical methods (e.g., correlation) to analyze data and draw conclusions \cite{3796180663,2634033325,518268671}. \\
        Network analysis & Studying relationships and interactions using graph-based approaches \cite{2345021698,4019205162,1426267857}. \\
        Pattern Extraction & Identifying meaningful patterns or structures within data, including techniques like Markov analysis and sequence mining \cite{123412197,1469065963,1345598079}. \\
        \bottomrule
    \end{tabular}
    \caption{Analysis methods.}
    \label{tab:analysis_methods}
\end{table}

There is a notable shift from model-based analysis in Corpus~A (57/73; 78\%~\cite{3448122334, 4278392816}) to model-free approaches in Corpus~B (33/49; 67\%~\cite{2005607968, 2280467946}). Model-based analyses in both corpora primarily involve supervised learning methods, such as classification and regression, often supplemented by statistical techniques (e.g., correlation analysis). These studies typically use input features derived from speech, video, log, and sensor data to predict outcome variables such as performance or engagement~\cite{3408664396,2456887548,1118315889}. They focus primarily on individual learners, reflecting the difficulty of capturing complex social dynamics within formal, parameterized models.

In contrast, model-free approaches take a more exploratory stance, employing qualitative, clustering, statistical, and pattern-extraction techniques. Qualitative methods (e.g., interaction analysis) draw on theory and observation to interpret multimodal traces~\cite{1847468084}, while statistical and pattern-based approaches highlight relationships between behavior and outcomes (e.g., correlations between strategies and learning gains). These methods are especially prevalent in collaborative learning settings, where they are used to unpack social signals and discourse~\cite{2345021698,4019205162}.

For example, \citet{2005607968} used k-means clustering to identify collaborative patterns in undergraduate pair programming using standardized ratings of process quality (9 dimensions) and programming outcomes (4 dimensions). The resulting clusters differed meaningfully in both collaboration quality and performance. High-performing pairs engaged in knowledge construction, consensus-building discourse, positive affect, and iterative loops between talk and code adjustment. In contrast, lower-performing clusters were marked by self-talk, fragmented regulation, excessive debugging, and weaker coordination between discourse and behavior. Clusters were labeled consensus-achieved, argumentation-driven, individual-oriented, and trial-and-error, with the consensus-achieved group showing the strongest outcomes. Here, clustering functioned as a mid-level segmentation step, enabling data-driven insights into how multimodal interactions help explain relationships between collaboration and learning outcomes without relying on prior assumptions.

While both model-based and model-free methods are valuable across both corpora, each comes with inherent trade-offs. A persistent tension exists between the predictive strength and structure offered by model-based approaches---allowing researchers to leverage domain knowledge to define variable relationships that effectively guide analysis---and the interpretive richness and flexibility of model-free analyses that allows for the discovery of unanticipated insights. Choosing between them is not always straightforward. A balanced and often beneficial strategy is to employ both approaches in tandem: model-based analysis to test hypothesized relationships, while model-free methods to reveal latent patterns.

\subsection{Feedback}\label{subsec:feedback}

In multimodal learning and training environments, feedback emerges when systems are deployed in real-world contexts (e.g., classrooms), typically taking one of two forms. \textbf{Direct feedback} refers to feedback explicitly provided to the user by the system---such as a pedagogical agent assisting a student--- to improve performance or other learning metrics. \textbf{Indirect feedback}, conversely, is not intended for the end user but is derived from analysis of system use or learner behavior. It informs researchers and developers on how to refine their systems. Such feedback may arise from observing user-system interactions or analyzing outcomes across learner populations, ultimately leading to deeper insights that can be used to improve systems. Both types of feedback are essential for advancing MMLA and helping close the loop between methodological innovation and applied practice.

Every paper in Corpora~A and~B incorporated indirect feedback in some capacity~\cite{1637690235, 603534886}, highlighting the importance of using authentic studies with human subjects to refine system behavior. By contrast, the extent to which direct feedback was employed varied considerably across the two corpora. In the pre-LLM era, only 41 of 73 papers (43.8\%) provided direct feedback to users, compared to 30 of 49 papers (61.2\%) in the post-ChatGPT era~\cite{1576545447, martinez-maldonado_data_2020}. The LLM era has also enabled significantly more dynamic forms of direct feedback: learners and trainees can now engage in \textit{dialogic} interactions, receiving feedback through rich exchanges with LLM systems that retain conversation history and support stateful interaction~\cite{328477558, 3304069824}.

The way multimodality is employed to deliver direct feedback differs substantially across the two corpora. For example, before LLMs, \citet{85990093} introduced the \textit{Virtual Debate Coach}, which monitors trainees' speech, prosody, posture, and gestures through multimodal sensing and analysis. The system extracts features such as filled pauses, speech pitch, and gestures derived from 3-D video coordinates to train an SVM classifier that estimates debaters' confidence levels. Feedback is then generated using predefined rules and expert-informed strategies. 

Researchers used indirect feedback in this case to extend the system's machine learning capabilities by enabling automatic detection and interpretation of behavioral variation, as well as assessment of debater proficiency. Direct feedback was provided both formatively and summatively to help students improve their performance and confidence; however, student-agent interactions are stateless, lacking dialogue-state tracking, turn-level language modeling, or access to conversational history. This represents a canonical pre-LLM feedback paradigm in which multimodal features are manually engineered and combined with rule-based or heuristic logic to produce feedback within a discrete response space.

By contrast, multimodal LLMs operate in a continuous space and can process heterogeneous data directly, without requiring engineered features. \citet{3224774131} employed GPT-4o to automatically score and generate explanatory feedback on students' multimodal science assessments. The authors demonstrated that LLMs can ingest handwritten student assessments---including textual and visual content as a single input---with over 90\% transcription accuracy, achieving grading alignment comparable to human raters (Cohen's $k = 0.84$). Feedback quality was further enhanced through prompt engineering with few-shot exemplars, yielding responses that were more accurate and better aligned with teacher feedback.

While their system provided direct feedback to students in the form of a score and an accompanying explanation, they also used indirect feedback to improve system design through thematic error analysis. As the authors themselves note, the findings ``present opportunities for designing learning analytics systems that allow for iterative evaluation and modification [of LLMs'] assessment output''~\cite{3224774131}. Their analysis revealed that the LLM (1) failed to evaluate the depth of students' responses accurately, (2) hallucinated information not present in the prompt, and (3) exhibited inaccurate numerical reasoning. These were identified as the most critical issues to address in future iterations.

Additionally, the ease of deploying LLM-based feedback systems at scale (e.g., via API calls to OpenAI) has contributed to the emergence of multimodal dashboards and tools that serve as feedback layers for teachers and students, supporting guided reflection and debriefing rather than functioning solely as research instruments~\cite{261302708, 1040787959}. In parallel, the multimodal capabilities of enterprise LLMs such as ChatGPT, Claude, and Gemini have facilitated the integration of GenAI-based systems with logs, artifacts, and other multimodal traces to generate personalized, data-driven feedback. These systems are designed to support self-directed learning, enhance engagement, and improve learner performance~\cite{1441411748, 2429627610, 3537775194}.

However, the rise of LLM-based feedback systems has introduced several challenges for multimodal learning and training. Human feedback often outperforms or qualitatively differs from AI-generated feedback, particularly in complex tasks \cite{fonteles2026JLI}. This gap highlights ongoing design tensions around trust, interpretability, and the roles of human and AI actors in direct feedback ecosystems~\cite{425012016, 2846172025}.

In addition, the innate fusion capabilities afforded by contemporary multimodal LLMs often come at the expense of user control and transparency. While feature engineering is time-intensive, it enables researchers to evaluate which features contribute to model performance. In contrast, LLMs typically accept a single multimodal input~\cite{3224774131}, internally extracting features that are neither observable nor modifiable by users, and whose influence can only be evaluated indirectly through techniques such as perturbation analysis.

Recent work has shown promising results using multimodal late fusion with LLMs for direct feedback by first distilling each modality into text and then leveraging the LLM to perform textual fusion~\cite{fonteles2026JLI,fonteles2025exploring} before feedback. However, this approach relies heavily on prompt engineering. Most studies employ ad hoc prompting strategies, with limited attention to systematically aligning generated feedback with established pedagogical principles~\cite{cohn2025theory}. This gap is often attributed to the absence of established learning frameworks in software engineering pipelines~\cite{cohn2024chain}.

\subsection{Summary}

The four framework components---Environment, Multimodal Data, Learning Analytics, and Feedback---collectively illustrate how multimodality is used in learning and training environments. The environment determines which modalities can be captured and in what context, setting the stage for meaningful data collection. These interactions yield rich multimodal data streams, each offering unique windows into learning and training processes. Learning analytics fuses the heterogeneous data for analysis to extract insights, uncover patterns, and make inferences about learning and performance. These insights are used to generate feedback, either directly to learners and trainees or indirectly to researchers, engineers, and system designers to inform theory and improve educational tools. Across all four components, multimodality is the connective tissue that enables holistic, context-aware, and actionable understandings of learning and training in complex environments.

However, the approach to multimodal learning and training research differs markedly between Corpus~A and Corpus~B. Table~\ref{tab:pre_post_llm_comparison} outlines key methodological shifts from pre-LLM multimodal learning analytics to more recent GenAI-enabled practices, highlighting how large transformer-based models have redefined data requirements, fusion strategies, and analytic workflows. Although researchers in Corpus B continue to apply and refine traditional methods established in Corpus A, the rapid adoption of LLMs and GenAI signals a clear and ongoing paradigm shift in the field.

\begin{table*}[t]
    \centering
    \small
    \begin{tabular}{p{3.2cm} p{4.5cm} p{6.2cm}}
        \toprule
        \textbf{Dimension} &
        \textbf{Pre-LLM MMLA Methods (2017--2022)} &
        \textbf{Post-LLM / GenAI-Enabled MMLA Methods (Late 2022--Present)} \\
        \midrule
        
        Feature Engineering &
        Predominantly manual and domain-specific feature extraction (e.g., handcrafted gaze metrics, prosodic features, rule-based textual features). &
        Reduced reliance on manual feature engineering through pretrained representations and prompt-based abstraction, though handcrafted features remain common in applied settings. \\
        
        Model Architectures &
        Classical machine learning (e.g., SVMs, random forests) and task-specific deep learning models (e.g., CNNs, LSTMs). &
        Increasing use of transformer-based foundation models (e.g., LLMs, VLMs, multimodal transformers---particularly GPT-series models), often combined with task-specific components. \\
        
        Fusion Strategies &
        Explicit early, mid, or late fusion pipelines designed and tuned per task. &
        Hybrid fusion approaches combining explicit fusion pipelines with implicit cross-modal reasoning enabled by pretrained models. \\
        
        Data Requirements &
        Substantial labeled datasets are required for model training and validation. &
        Support for reduced annotation through transfer learning and zero- or few-shot inference, depending on task and context. \\
        
        Adaptability Across Tasks &
        Limited generalization; models are typically trained for a single task or environment. &
        Improved cross-task and cross-domain transferability enabled by pretrained models, though adaptation remains context-dependent in applied environments and can require substantial prompt engineering. \\
        
        Handling of Unstructured Data &
        Limited support for open-ended or qualitative data (e.g., discourse, reflection, embodied activity). &
        Improved capacity to process unstructured and open-ended multimodal data, particularly in language-rich and mixed-modality tasks. \\
        
        Human-in-the-Loop Interaction &
        Primarily offline analysis and post-hoc interpretation of multimodal data. &
        Emerging support for interactive and human-in-the-loop analytics, including AI-assisted feedback and sense-making in certain contexts (e.g., assessment). \\
        
        Interpretability and Transparency &
        Relatively interpretable pipelines with explicit features and model logic. &
        Foundation models introduce new interpretability challenges, alongside emerging practices for prompting, validation, and human oversight. \\
        
        Scalability and Deployment &
        Deployment constrained by sensing setups, preprocessing pipelines, and model retraining requirements. &
        Easier prototyping and deployment via APIs and pretrained models, coupled with new constraints related to cost, latency, privacy, and governance. \\
        
        Methodological Constraints &
        Strong dependence on controlled data collection, domain expertise, and context-specific sensing infrastructures. &
        Shift toward software-centric constraints, including model access, computational cost, data privacy, and alignment with institutional policies. \\
        
        \bottomrule
    \end{tabular}
    \caption{Comparison of Pre-LLM (Corpus A) and Post-LLM (Corpus B) Methodological Affordances in Applied Multimodal Learning and Training Analytics}
    \label{tab:pre_post_llm_comparison}
\end{table*}

In the following section, we outline three \textbf{archetypes} of multimodal learning and training research that exemplify how this research is conducted in practice. Rather than introducing new categories, the archetypes synthesize recurring configurations of methods, environments, and analytic goals observed across the corpus. Organizing the literature around these categories helps elucidate the field's diverse goals and recurring methodological approaches. To anchor each archetype, we introduce a representative case study that highlights its core characteristics in context.

\section{Archetypes} \label{sec:archetypes}

Following the analysis in Section~\ref{sec:results}, we reexamined our corpus to classify prevailing research objectives in the application of multimodal methods to learning and training environments. We identified three primary research objectives, termed archetypes: \textbf{Designing and Developing Methods}, \textbf{Analyzing Outcomes}, and \textbf{Exploring Behaviors}. These archetypes, detailed in subsequent subsections, often overlap within studies---i.e., they are not mutually exclusive, and individual studies may instantiate multiple archetypal patterns. For example, method development research may also yield insights into participant behaviors and outcomes. While these archetypes broadly define the field, they are not exhaustive, and some studies may not align precisely with these categories.

\subsection{Designing and Developing Methods} \label{subsec:mdd}

The Designing and Developing Methods archetype encompasses studies that focus on designing, constructing, and evaluating multimodal research methods applicable to learning and training environments. These studies prioritize methodological innovation over the derivation of generalizable findings about a population. Although the developed methods often aim to predict outcomes (Section~\ref{subsec: outcomes}) and discern behaviors (Section~\ref{subsec: behaviors}), the primary focus remains on the methods themselves, not the implications of their findings for study participants. 

These methods are typically quantitative, utilizing supervised learning techniques such as classification \cite{1598166515,2000036002,2070224207} and regression \cite{3051560548,1581261659,2836996318}, and their efficacy is reported using performance metrics such as F1-scores \cite{1019093033,1886134458,2456887548}. Data collection often involves video, audio, and log data \cite{3796643912,3796180663,518268671}; targeting modalities such as affect, pose, prosodic speech, and logs \cite{3408664396,3796643912,3093310941}; employing data fusion techniques like mid or hybrid fusion \cite{566043228,1609706685,3625722965} using model-based approaches \cite{2836996318,1581261659,2070224207}.

Our corpus reveals a broad spectrum of tasks addressed by Designing and Developing Methods research, ranging from engagement detection in educational games \cite{3408664396} to skill classification in sports \cite{3625722965}. The versatility of multimodal methods is evident across diverse settings, domains, instructional levels, and didactic approaches, without a dominant trend in any one area.

\subsubsection{Case Study: Designing and Developing Methods in Psychomotor Skill Training}

A representative example of the Designing and Developing Methods archetype is the \textit{Table Tennis Tutor (T3)} system, developed for automated forehand stroke classification in psychomotor training~\cite{3625722965}. Within our framework, the environment corresponds to a training context centered on motor skill acquisition, involving a novice-oriented table tennis practice scenario with one human trainee and an expert coach. The setting is a casual physical sports environment (i.e., a recreational room), where data are collected using smartphone sensors (accelerometer and gyroscope) placed in the player's pocket and a Microsoft Kinect V2 depth camera positioned at the center of the table. 

Multimodal data are captured using the physiological signals and vision modality groups---including inertial motion and 3-D trajectories of skeletal joints---which are synchronized using a Multimodal Learning Hub~\cite{schneider2018multimodal}. A total of 510 forehand strokes were collected, both correct and incorrect, which were manually annotated using time-aligned video and sensor traces from 33 recorded training sessions. Additionally, the authors conduct an expert interview with the coach to gather design requirements and evaluate the system's acceptability; however, no controlled intervention study involving trainees is implemented.

Within the learning analytics component of our framework, T3 employs mid fusion, in which features from the smartphone and Kinect sensors are temporally aligned and integrated after initial processing. The analysis approach is strictly model-based, employing an LSTM network to perform binary classification of correct versus incorrect strokes. Three model configurations were evaluated: smartphone-only, Kinect-only, and combined sensing. Performance was assessed using standard quantitative metrics aligned with this archetype: accuracy, precision, and recall. The fused multimodal configuration achieved the highest precision (0.73), highlighting the methodological value of combining heterogeneous sensor streams for multimodal analysis. Notably, the primary research contribution lies not in pedagogical insight, but in the comparative evaluation of sensing configurations and the demonstration of a multimodal infrastructure for stroke classification.

Finally, T3 illustrates the feedback component in a manner typical for this archetype: primarily indirect, informing system developers and researchers about sensor placement, fusion strategies, and model performance. While future work discusses integrating direct feedback to learners via automated auditory cues, this functionality remains mainly conceptual at this stage. This aligns closely with the defining characteristics of the \textit{Designing and Developing Methods} archetype: the study prioritizes method construction, data fusion, and neural classification performance over validation of downstream learning impact, thereby grounding our framework's multimodal pipeline in an authentic yet method-centric training scenario.

\subsection{Analyzing Outcomes} \label{subsec: outcomes}

The Analyzing Outcomes archetype centers on identifying relationships between multimodal signals and specific outcome metrics (e.g., learning gains). Unlike the Designing and Developing Methods archetype, this approach prioritizes understanding the system's impact on learning or training, i.e., seeking findings generalizable across broader populations. Studies commonly employ supervised learning techniques, including classification \cite{1326191931, 1877483551, 2070224207} and regression \cite{1581261659, 2836996318, 3093310941}, alongside insights derived from statistical trends and unsupervised analyses \cite{3796180663, 4278392816, 2000036002}.

This archetype has been applied in diverse learning and training contexts, examining constructs such as attention and engagement \cite{1315379489, 3448122334, 1581261659}, task accuracy and performance \cite{3625722965, 1576545447, 1886134458}, learning outcomes \cite{1763513559, 4277812050, 433919853}, and collaborative dynamics \cite{3754172825, 3309250332, 1637690235}. Despite variation in settings, consistent use of common outcome variables has enabled the field to produce findings with cross-context relevance. However, the emphasis on outcomes can obscure the underlying complexity of learning processes, potentially leading to interventions optimized for high-achieving learners while overlooking variability in individual needs and trajectories \cite{Fonteles2024Underserved}.

\subsubsection{Case Study: Analyzing Outcomes in Embodied Math Learning with GenAI Feedback}

A representative example of the Analyzing Outcomes archetype is the study by \citet{2846172025}, which examines the impact of LLM-generated formative feedback on students' learning and cognitive engagement in an embodied mathematics environment using a body-scale number line. The environment is situated within a learning context in a physical classroom in Norway, involving middle-school students (ages 11-13) solving integer arithmetic problems through full-body movement. The setting is a multisensory space featuring floor and wall projections, with sensors comprising mobile eye-tracking glasses, multiple video cameras, and system-level motion tracking.

The resulting multimodal data span vision (eye gaze, fixation patterns), physiological sensors (pupil dilation, body movement), and logs (task attempts, correctness, and AI-generated feedback). The study adopts a between-groups design comparing human (teacher) feedback with LLM feedback generated by GPT-4, directly aligning multimodal sensing with interpretable outcome variables.

For learning analytics, the study employs hybrid fusion by synchronizing gaze, pupil, and interaction log data to construct composite indicators of learning and engagement. The analysis approach is primarily model-free, relying on statistical hypothesis testing (t-tests with normalization and Welch correction) rather than predictive modeling. Core outcome variables include task performance (as a proxy for learning gain), cognitive load inferred from pupillary oscillations, time spent on areas of interest (AOIs), and transitions between AOIs. While no significant differences in task performance were observed between conditions, significant effects were found for cognitive load and information processing strategies. Students receiving GenAI feedback exhibited lower cognitive load and more balanced visual processing than those receiving teacher feedback. 

Finally, the feedback component illustrates both direct and indirect roles within our framework. Direct feedback is delivered either by the human teacher through multimodal dialogue and gestures, or by GenAI through structured, text-based formative hints generated from real-time system logs. Indirect feedback is derived from post hoc multimodal analysis, providing researchers with insights into how different feedback modalities influence learners' attention, verification strategies, and cognitive effort.

The study's primary goal is not to optimize the feedback-generation method itself (as in the Designing and Developing Methods archetype), but rather to evaluate its impact on learning-relevant outcomes. This positions it as a canonical instantiation of the Analyzing Outcomes archetype: multimodal data and analytics are used primarily to explain differences in learning gains, engagement, and cognitive load across experimental conditions, supporting population-level inferences about the pedagogical implications of GenAI feedback.

\subsection{Exploring Behaviors} \label{subsec: behaviors}

The Exploring Behaviors archetype investigates human behavior and experiences in learning and training contexts by employing an exploratory approach to uncover influencing factors. This research examines a variety of human signals that vary temporally, socially, and spatially, and are tailored to specific learning objectives. Unlike other archetypes, it often incorporates qualitative observations \cite{3809293172, 3660066725, 86191824, 3856280479} and employs data exploration techniques such as correlation analysis \cite{2345021698, 518268671} and pattern recognition \cite{1469065963, 3308658121, 4019205162, 818492192}.

Data fusion in this context is typically qualitative \cite{3398902089, 666050348, 3146393211}, involving the manual integration of multimodal data sources. This approach enables triangulation of student and trainee behaviors, providing a richer context to researchers, statistical analyses, or data visualizations to facilitate deeper insights into the behaviors under study.

Exploring Behaviors research aims to fill knowledge gaps in learning theory and technological applications by investigating human behavior in educational contexts. \citet{3308658121} applied a Markov transition model to assess how students' physical behaviors during a collaborative programming task correlate with collaboration quality, task performance, and learning gains. \citet{2345021698} utilized correlation analysis alongside social network metrics and annotated behaviors to investigate collaborative dynamics in a software engineering course. \citet{3809293172} conducted a qualitative study, using a coding scheme to analyze students' actions, speech, and gestures in embodied learning activities to understand their conceptualization of measurement. 

These studies, often grounded in learning theory, employ multimodal learning analytics to dissect the components of effective learning and collaboration, showcasing the nuanced insights that multimodal methods can provide into these processes. This research spans various sensors, modalities, and settings, with a discernible focus on collaboration.

\subsubsection{Case Study: Exploring Behaviors in Online Collaborative Problem-Solving}

A canonical example of the Exploring Behaviors archetype is the study by \citet{1687167932}, which examines how college students' attention fluctuates during online collaborative problem solving (CPS) using a systemic multimodal approach. The learning environment is an online laboratory session within a computer networking course, where groups of three students collaborate synchronously to conceptualize and solve subnetting problems. The setting consists of a video conferencing platform (Tencent Meeting) with screen-sharing functionality. Sensors include EEG devices worn by all participants, video recordings that capture audio and interaction dynamics, and performance tests.

The multimodal data collected spans physiological signals (attention derived from EEG), human-centered evidence (coded video-recorded actions), and performance-measure logs, illustrating the archetype's emphasis on rich, temporally ordered human signals. The study further distinguishes collaborative patterns---centralized, distributed, and individual---by qualitatively classifying behaviors observable in the video data, as evidenced by coded examples and aligned attention traces.

Learning analytics in this study centers on exploratory modeling rather than prediction, combining correlation analysis, ANOVA, and a Hidden Markov Model (HMM) to uncover latent cognitive states and their transitions. Data fusion is qualitative and manual: researchers triangulate coded social, task-related, and individual behaviors with EEG-derived attention measures to construct a systemic account of collaborative problem-solving processes. 

Behavioral sequences were hand-coded into categorical labels based on detailed video observations. These categories were then used as emission probabilities within the HMM to infer hidden attention states. This approach to fusion exemplifies the archetype's exploratory orientation, in which researchers aim to interpret behavioral mechanisms rather than optimize model performance or predict outcomes.

Findings reveal how attention dynamically interacts with collaborative structures. Centralized groups exhibited synchronized attention patterns; distributed groups showed fluctuating attentional states; and individual groups displayed persistent inattention. These interpretations underscore how exploratory multimodal analytics can surface latent cognitive and social processes that are not observable through performance data alone.

Feedback within this archetype is primarily indirect, offering insight for researchers and educators rather than providing real-time intervention. The multimodal analysis yields explanations for why specific collaborative patterns support or hinder attention. For example, centralized groups exhibit more stable attentional engagement due to coordinated leadership, and distributed groups face challenges in sustaining collective focus. These findings, drawn from integrated behavioral and cognitive data, contribute to theory-building on CPS dynamics and attention, aligning with this archetype's emphasis on understanding human behavior in context.

This case illustrates how multimodal learning analytics can unpack nuanced behavioral processes across temporal, social, and cognitive dimensions, providing a layered understanding of collaborative learning dynamics grounded in qualitative interpretation and exploratory modeling.

\section{Discussion and Conclusions} \label{sec:discussion}

In this review, we examined how multimodal data are collected, fused, and analyzed in applied learning and training contexts, with particular attention to how the field has evolved in response to emerging technologies like GenAI and LLMs. We mapped the methodological landscape across data sources, modality types, fusion strategies, and analysis approaches, highlighting both enduring practices and newer techniques enabled by foundation models. We presented a unified, empirically grounded framework and taxonomy that explain how methodological choices are shaped by learning environments, data, and analytic goals. This allowed us to synthesize persistent methodological challenges that hinder scalability and real-world deployment. We concluded by identifying three methodological archetypes that illustrate how multimodal components are combined in practice to design intelligent systems, assess learning outcomes, and explore learner behaviors. 

Together, these contributions offer a coherent account of the current methodological landscape in multimodal learning and training research. In the following subsections, we reflect on the insights our framework affords, identify research gaps, examine persistent methodological and practical challenges, and outline promising directions for future research. We conclude by presenting broader implications for the design, deployment, and interpretation of multimodal learning and training systems. 

\subsection{Framework Insights and Research Gaps}

Multimodality is not simply an exercise in data aggregation: when designed effectively, these systems can capture patterns that no single modality reveals alone, thus producing insights that are greater than the sum of their parts \cite{3095923626}. Multimodal methods have consistently demonstrated effectiveness in predicting learning outcomes and identifying salient features that drive those outcomes \cite{3339002981,1637690235,3783339081}. Each modality encodes distinct information (e.g., metacognition in collaborative discourse \cite{snyder2025using,snyder2023analyzing}), and when combined thoughtfully, yields more comprehensive and context-rich representations of learning and training. Even when predictive accuracy does not improve, multimodality can surface meaningful patterns, offering greater interpretability and opportunities to design more responsive and effective educational systems \cite{vrzakova_focused_2020,1770989706}. 

Recent advances in LLMs further extend the affordances of multimodal systems by enabling late-fusion classification and prediction through the distillation of individual modalities into semantically meaningful labels, accompanied by human-readable justifications that articulate model reasoning \cite{fonteles2026JLI}. This interpretive capacity introduces new opportunities for transparency and dialogue in learning analytics. When paired with structured prompt engineering and iterative refinement via active learning \cite{cohn2024chain}, LLMs can generate time-aligned behavioral explanations across modalities, allowing researchers and educators to examine outputs through natural language contextualized by other modalities. These capabilities offer a dual benefit: enhancing interpretability and supporting sense-making within human-in-the-loop workflows across diverse environments and tasks (e.g., embodied learning, assessment, feedback).

While our findings highlight the distinct advantages of multimodal approaches, the literature reveals both areas of maturity and critical, underexplored gaps across the four components of our framework. Within the environment component, most research remains concentrated in small-scale classroom settings, typically within K-12 or postsecondary contexts. These studies are often anchored to short-term outcomes or single-session observations, with limited attention to training environments, younger learners in primary education, or adult learners outside the university system. Longitudinal perspectives, which are essential for understanding how learners and trainees evolve over time, are notably scarce.

Challenges surrounding multimodal data collection, synchronization, and alignment continue to limit scalability. Although widely acknowledged, these issues remain unresolved in most real-world studies. Recent efforts have targeted the cross-modal interaction bottleneck through open-source tools like SyncFlow \cite{timalsinasyncflow}, which enable automated collection and alignment of webcam, screen, and audio data via web browsers or standalone applications. However, such tools are rarely adopted in practice and remain in the early stages of development. The scarcity of large, open-source datasets tailored for multimodal learning further constrains replication, benchmarking, and broader methodological progress.

Within learning analytics, there is growing use of LLMs and machine learning pipelines to process multimodal signals; however, the interpretability of system decisions remains a significant challenge. Explanations for model decision-making are often lacking or inaccessible to practitioners. Furthermore, the integration of LLMs in this space is limited, with most studies relying on proprietary GPT-based APIs used out of the box. This leaves a gap in the development of more advanced pipelines and little exploration of open-source alternatives that could promote transparency and trust.

The feedback component reflects early momentum toward LLM-generated formative and explanatory feedback. However, few studies rigorously examine whether such feedback aligns with curricular goals, teachers' expectations, or broader pedagogical intent. Trust and adoption challenges, particularly from educators, remain an impediment to adoption, and little is known about how learners engage with or respond to LLM feedback in authentic settings.

These findings illuminate the current state of multimodal learning and training research and reveal a set of persistent, interrelated gaps and challenges. The following subsections build on this analysis by categorizing the core challenges that hinder progress and outlining future directions to guide the design of more impactful MMLA systems.

\subsection{Challenges and Limitations}

While multimodality has significantly advanced learning and training, several major challenges persist. 

\paragraph{Data.}
Issues with data collection, labeling, and overall scarcity were widely reported across both corpora. Studies involving human subjects often suffer from small $n$, as data collection is time-consuming, resource-intensive, and subject to strict privacy constraints. These factors make it especially difficult to deploy AI systems in authentic classroom settings and collect data at scale---data that is almost never released to the public, particularly if study participants are minors. As a result, many studies rely on limited or demographically narrow datasets, reducing the generalizability and reliability of findings \cite{2609260641,2155422499,1426267857}. Models trained on such data are prone to distributional shifts, introducing biases and compromising predictive validity \cite{32184286}. 

\paragraph{Cross-Modal Interaction.} 
Integrating multimodal data is both complex and resource-intensive, requiring systems to reconcile heterogeneous data representations that vary in structure, granularity, and temporal dynamics \cite{3783339081}. Even minor misalignments can propagate through the pipeline, undermining performance and interpretability. Real-time alignment introduces further difficulty, as modalities may arrive asynchronously, with inconsistent latencies and noise that disrupts temporal and semantic coherence. These challenges are compounded by the field's reliance on mid fusion, where individual modalities are preprocessed into intermediate features before integration, often via enterprise software whose predictions are unverifiable. Yet these predictions are frequently adopted as ground truth for downstream modeling, introducing potential inaccuracies that are often not subjected to human validation.

\paragraph{LLMs.}
While LLMs help alleviate annotation burdens and accelerate data pipelines, their integration introduces a host of new challenges. The size and opacity of enterprise LLMs like ChatGPT complicate efforts to ensure transparency, trustworthiness, and pedagogical alignment. In response, several states and school districts have begun banning or restricting their use in classrooms, citing unresolved ethical, pedagogical, and governance concerns. Training smaller, open models could mitigate some of these issues. However, this is often infeasible due to data scarcity, compute costs, and the technical complexities of self-hosted LLM systems---including concurrency and the management of agent interactions across multiple users. Students' expectations are simultaneously shifting through frequent exposure to GenAI in their personal lives; many now seek immediate, definitive answers, often at the expense of critical thinking \cite{cohn2025personalizing}. This growing dependence risks fostering overreliance on LLM-generated content, potentially undermining pedagogical goals.

\paragraph{Limitations of This Review.}
We acknowledge the limitations of our literature review, including potential reproducibility issues with Google Scholar, the risk of relevant work being excluded during citation graph pruning, and inconsistencies in versioning across published manuscripts. However, based on our comprehensive analysis of the corpus, we are confident that these factors do not compromise the validity of our findings or the inferences drawn. A detailed discussion of these limitations is provided in Appendix~\ref{sec:literature_review_limitations}.

\subsection{Future Work}

Addressing these research gaps and challenges opens several promising directions for future work.

\paragraph{Active Environments.}

Multimodal research in physically active settings remains rare, despite its strong potential to advance activities such as physical training and therapy. Active environments naturally generate rich, multimodal signals, including gesture, gaze, speech, and spatial coordination. These contexts are particularly relevant for domains like embodied learning, rehabilitation, athletics, and hands-on skill training, where understanding nuanced physical actions is critical. Leveraging multimodality in such settings could enable more robust activity recognition, support real-time feedback, and reveal patterns of behavior that are difficult to detect through unimodal approaches.
    
\paragraph{Longitudinal Analyses.}

Longitudinal research remains scarce in multimodal learning and training, with most studies focused on predicting outcomes or identifying correlates at single time points. Few investigate examine how learners or trainees evolve using multimodal data. While longitudinal analyses have been successfully conducted with unimodal or digital trace data \cite{boulton2019student}, multimodal applications remain limited due to challenges in scalability, data standardization, and the complexity of maintaining synchronized multimodal logs over extended periods \cite{lixiang2022scalability}. Yet, such work is essential: longitudinal MMLA could offer richer insights into developmental trajectories, shifts in engagement, and the dynamics of learning or skill acquisition. Addressing these challenges would support more adaptive systems and enable a deeper understanding of progression in authentic, evolving contexts.

\paragraph{Standardization.}

Standardization remains a persistent challenge in multimodal learning and training research. Progress will require the development of shared, open logging formats capable of capturing complex, temporal, and multimodal data, along with stronger adoption of existing norms \cite{lixiang2022scalability}. Such efforts would reduce fragmentation across research teams, improve reproducibility, and enable more scalable, interoperable systems for both research and deployment.
        
\paragraph{Interpretability.}

A significant challenge in multimodal learning and training systems is their limited interpretability, which can undermine trust and hinder pedagogical integration. Emerging techniques, such as using LLMs to generate natural language rationales or visual explanations, offer promising avenues for making system behavior more intelligible to both educators and learners. Interpretability is also key to ensuring pedagogical alignment, allowing educators to verify that system feedback reinforces intended learning objectives rather than contradicting or bypassing them. While multimodal learning and training research has begun to make direct feedback more interpretable to stakeholders by using \textit{chain-of-thought} \cite{wei2022chain} prompting to explain model reasoning \cite{cohn2024chain}, the inner workings of foundation models and complex systems remain largely opaque.

\paragraph{Agentic and Multi-Agent Workflows.}

There is increasing interest in agentic and multi-agent architectures capable of coordinating diverse functions across sensing, reasoning, and feedback. These pipelines allow individual components---such as activity recognition, affect analysis, and instructional scaffolding---to operate semi-autonomously while contributing to a shared objective. Emerging frameworks, such as Anthropic's Agent Skills \cite{anthropic2025agentSkills}, offer modular, reusable capabilities that can be composed to create more adaptable, context-aware pedagogical agents. Similarly, Model Context Protocol (MCP) \cite{anthropic2024modelContext} introduces structured ways to track contextual metadata alongside model behavior, improving traceability and interpretability across agent interactions. These approaches offer promising solutions to long-standing issues of scalability, modularity, and pedagogical alignment in multimodal systems.

\subsection{Implications}

As we look to the future, this review highlights important considerations for the design, implementation, and evaluation of multimodal learning and training systems. As these systems become more advanced, especially with the integration of LLMs, we must assess their impact not only on technical performance but also on their educational value and alignment with effective teaching practices. To fully harness the potential of multimodal approaches, we need integration strategies that are context-aware and aligned with instructional goals.

Multimodal data integration has evolved significantly in the post-LLM era. Where earlier systems required handcrafted features and explicit synchronization across channels, today's pipelines increasingly rely on foundation models to infer structure, relationships, and meaning from raw or weakly-labeled inputs. These powerful models offer new capabilities for generating explanations, scaffolding instruction, and streamlining feedback---as well as accelerating development and lowering the barrier to entry. However, the proliferation of LLMs in multimodal learning and training systems also raises new concerns: model-generated errors can now propagate more easily across modalities, especially when early outputs are treated as ground truth. 

Prior LLM work in intelligent learning environments has shown how early misclassifications can shape future predictions and create degenerate feedback loops \cite{fonteles2026JLI}. As such, maintaining mechanisms for verification, revision, and human-in-the-loop oversight is critical to prevent the compounding of errors. Research has also shown that LLMs can hinder---or even harm---learning when learners over-rely on generated content or accept it uncritically \cite{kosmyna2025your,zhou2025impact}. The ease and fluency of LLM outputs may also disincentivize effortful learning, encouraging passive consumption rather than active engagement. This is particularly problematic when learners become frustrated and expect direct answers, undermining metacognitive development and long-term retention \cite{cohn2025personalizing}.

A significant question remains largely unaddressed: \textit{what if students simply choose not to engage with the systems we design}? While much of the multimodal learning and training literature focuses on technological advancement, it often overlooks the foundational issue of stakeholder uptake. Regardless of how advanced or well-designed an MMLA system may be, its practical effectiveness depends on whether learners attend to and act on the feedback it provides. Prior work shows that LLM-generated feedback can yield substantial learning gains---but only for those who actively engage with it \cite{thomas2025llm}. More broadly, meaningful impact requires collective buy-in from researchers, administrators, teachers, students, and parents. Without this, even the most sophisticated systems risk failing to support learning---or failing to be used at all.

The future of multimodal learning and training depends not just on improving models and architectures, but on rethinking how we integrate these technologies into learning experiences. The goal is not to replace instruction, but to augment it by building systems that are transparent, responsive, and pedagogically grounded. Addressing these implications will require sustained collaboration across disciplines, careful attention to learner behavior, and a commitment to designing systems that serve educational values as much as technological advancement.



\bibliographystyle{ACM-Reference-Format}
\bibliography{references, zotero_references, uuid_references}

@INPROCEEDINGS{10628321,
  author={Thiannoi, Thanyalak and Junpeng, Putcharee and Intharah, Thanapong},
  booktitle={2024 International Technical Conference on Circuits/Systems, Computers, and Communications (ITC-CSCC)}, 
  title={Efficiency of Decision Tree Depth to Diagnose Mathematical Procedures in Number and Algebra for Seventh-grade Students}, 
  year={2024},
  volume={},
  number={},
  pages={1-6},
  doi={10.1109/ITC-CSCC62988.2024.10628321}}

@article{zedelius2019beyond,
  title={Beyond subjective judgments: Predicting evaluations of creative writing from computational linguistic features},
  author={Zedelius, Claire M and Mills, Caitlin and Schooler, Jonathan W},
  journal={Behavior research methods},
  volume={51},
  number={2},
  pages={879--894},
  year={2019},
  publisher={Springer}
}

@article{gansle2002moving,
  title={Moving beyond total words written: The reliability, criterion validity, and time cost of alternate measures for curriculum-based measurement in writing},
  author={Gansle, Kristin A and Noell, George H and VanDerHeyden, Amanda M and Naquin, Gale M and Slider, Natalie J},
  journal={School Psychology Review},
  volume={31},
  number={4},
  pages={477--497},
  year={2002},
  publisher={Taylor \& Francis}
}

@article{jiang2022empirical,
  title={An empirical analysis of high school students' practices of modelling with unstructured data},
  author={Jiang, Shiyan and Nocera, Amato and Tatar, Cansu and Yoder, Michael Miller and Chao, Jie and Wiedemann, Kenia and Finzer, William and Ros{\'e}, Carolyn P},
  journal={British Journal of Educational Technology},
  volume={53},
  number={5},
  pages={1114--1133},
  year={2022},
  publisher={Wiley Online Library}
}

@article{jarvela2023predicting,
  title={Predicting regulatory activities for socially shared regulation to optimize collaborative learning},
  author={J{\"a}rvel{\"a}, Sanna and Nguyen, Andy and Vuorenmaa, Eija and Malmberg, Jonna and J{\"a}rvenoja, Hanna},
  journal={Computers in Human Behavior},
  volume={144},
  pages={107737},
  year={2023},
  publisher={Elsevier}
}

@article{schoonderwoerd2021human,
  title={Human-centered XAI: Developing design patterns for explanations of clinical decision support systems},
  author={Schoonderwoerd, Tjeerd AJ and Jorritsma, Wiard and Neerincx, Mark A and Van Den Bosch, Karel},
  journal={International Journal of Human-Computer Studies},
  volume={154},
  pages={102684},
  year={2021},
  publisher={Elsevier}
}

@article{rojat2021explainable,
  title={Explainable artificial intelligence (xai) on timeseries data: A survey},
  author={Rojat, Thomas and Puget, Rapha{\"e}l and Filliat, David and Del Ser, Javier and Gelin, Rodolphe and D{\'\i}az-Rodr{\'\i}guez, Natalia},
  journal={arXiv preprint arXiv:2104.00950},
  year={2021},
  pages={N/A},
  numpages={N/A},
  number={N/A},
  volume={N/A}
}

@article{kitchenham2004procedures,
  title={Procedures for performing systematic reviews},
  author={Kitchenham, Barbara},
  journal={Keele, UK, Keele University},
  volume={33},
  number={2004},
  pages={1--26},
  year={2004}
}

@article{slade2013learning,
  title={Learning analytics: Ethical issues and dilemmas},
  author={Slade, Sharon and Prinsloo, Paul},
  journal={American Behavioral Scientist},
  volume={57},
  number={10},
  pages={1510--1529},
  year={2013},
  publisher={SAGE Publications Sage CA: Los Angeles, CA}
}

@article{kitto2019practical,
  title={Practical ethics for building learning analytics},
  author={Kitto, Kirsty and Knight, Simon},
  journal={British Journal of Educational Technology},
  volume={50},
  number={6},
  pages={2855--2870},
  year={2019},
  publisher={Wiley Online Library}
}

@article{baltruvsaitis2018multimodal,
  title={Multimodal machine learning: A survey and taxonomy},
  author={Baltru{\v{s}}aitis, Tadas and Ahuja, Chaitanya and Morency, Louis-Philippe},
  journal={IEEE transactions on pattern analysis and machine intelligence},
  volume={41},
  number={2},
  pages={423--443},
  year={2018},
  publisher={IEEE}
}

@article{khoo2024machine,
  title={Machine learning for multimodal mental health detection: a systematic review of passive sensing approaches},
  author={Khoo, Lin Sze and Lim, Mei Kuan and Chong, Chun Yong and McNaney, Roisin},
  journal={Sensors},
  volume={24},
  number={2},
  pages={348},
  year={2024},
  publisher={MDPI}
}

@article{shaik2024survey,
  title={A survey of multimodal information fusion for smart healthcare: Mapping the journey from data to wisdom},
  author={Shaik, Thanveer and Tao, Xiaohui and Li, Lin and Xie, Haoran and Vel{\'a}squez, Juan D},
  journal={Information Fusion},
  volume={102},
  pages={102040},
  year={2024},
  publisher={Elsevier}
}

@article{mondal2025survey,
  title={A survey of multimodal event detection based on data fusion},
  author={Mondal, Manuel and Khayati, Mourad and Sandlin, H{\^o}ng-{\^A}n and Cudr{\'e}-Mauroux, Philippe},
  journal={The VLDB Journal},
  volume={34},
  number={1},
  pages={9},
  year={2025},
  publisher={Springer}
}

@article{cengiz2025survey,
  title={A survey on data fusion approaches in IoT-based smart cities: Smart applications, taxonomies, challenges, and future research directions},
  author={Cengiz, Berna and Adam, Iliyasu Yahya and Ozdem, Mehmet and Das, Resul},
  journal={Information Fusion},
  volume={121},
  pages={103102},
  year={2025},
  publisher={Elsevier}
}

@article{teoh2024advancing,
  title={Advancing healthcare through multimodal data fusion: a comprehensive review of techniques and applications},
  author={Teoh, Jing Ru and Dong, Jian and Zuo, Xiaowei and Lai, Khin Wee and Hasikin, Khairunnisa and Wu, Xiang},
  journal={PeerJ Computer Science},
  volume={10},
  pages={e2298},
  year={2024},
  publisher={PeerJ Inc.}
}

@article{gaw2022multimodal,
  title={Multimodal data fusion for systems improvement: A review},
  author={Gaw, Nathan and Yousefi, Safoora and Gahrooei, Mostafa Reisi},
  journal={Handbook of Scholarly Publications from the Air Force Institute of Technology (AFIT), Volume 1, 2000-2020},
  pages={101--136},
  year={2022},
  publisher={CRC Press}
}

@article{hussain2024comprehensive,
  title={A comprehensive review on deep learning-based data fusion},
  author={Hussain, Mazhar and O’Nils, Mattias and Lundgren, Jan and Mousavirad, Seyed Jalaleddin},
  journal={IEEE Access},
  year={2024},
  publisher={IEEE}
}

@article{zhao2024deep,
  title={Deep multimodal data fusion},
  author={Zhao, Fei and Zhang, Chengcui and Geng, Baocheng},
  journal={ACM computing surveys},
  volume={56},
  number={9},
  pages={1--36},
  year={2024},
  publisher={ACM New York, NY}
}

@article{mu2020multimodal,
  title={Multimodal data fusion in learning analytics: A systematic review},
  author={Mu, Su and Cui, Meng and Huang, Xiaodi},
  journal={Sensors},
  volume={20},
  number={23},
  pages={6856},
  year={2020},
  publisher={MDPI}
}

@article{crescenzi2020multimodal,
  title={Multimodal Learning Analytics research with young children: A systematic review},
  author={Crescenzi-Lanna, Lucrezia},
  journal={British Journal of Educational Technology},
  volume={51},
  number={5},
  pages={1485--1504},
  year={2020},
  publisher={Wiley Online Library}
}

@INPROCEEDINGS{8433495,
  author={Shankar, Shashi Kant and Prieto, Luis P. and Rodríguez-Triana, María Jesús and Ruiz-Calleja, Adolfo},
  booktitle={2018 IEEE 18th International Conference on Advanced Learning Technologies (ICALT)}, 
  title={A Review of Multimodal Learning Analytics Architectures}, 
  year={2018},
  volume={},
  number={},
  pages={212-214},
  doi={10.1109/ICALT.2018.00057}}

@article{alwahaby2022evidence,
  title={The evidence of impact and ethical considerations of multimodal learning analytics: A systematic literature review},
  author={Alwahaby, Haifa and Cukurova, Mutlu and Papamitsiou, Zacharoula and Giannakos, Michail},
  journal={The multimodal learning analytics handbook},
  pages={289--325},
  year={2022},
  publisher={Springer}
}

@article{chango2022review,
  title={A review on data fusion in multimodal learning analytics and educational data mining},
  author={Chango, Wilson and Lara, Juan A and Cerezo, Rebeca and Romero, Crist{\'o}bal},
  journal={Wiley Interdisciplinary Reviews: Data Mining and Knowledge Discovery},
  volume={12},
  number={4},
  pages={e1458},
  year={2022},
  publisher={Wiley Online Library}
}

@article{di2018signals,
  title={From signals to knowledge: A conceptual model for multimodal learning analytics},
  author={Di Mitri, Daniele and Schneider, Jan and Specht, Marcus and Drachsler, Hendrik},
  journal={Journal of Computer Assisted Learning},
  volume={34},
  number={4},
  pages={338--349},
  year={2018},
  publisher={Wiley Online Library}
}

@incollection{ochoa2022multimodal,
  title={Multimodal learning analytics-Rationale, process, examples, and direction},
  author={Ochoa, Xavier and Lang, C and Siemens, G and Wise, A and Gasevic, D and Merceron, A},
  booktitle={The Handbook of Learning Analytics},
  pages={54--65},
  year={2022},
  publisher={Soc. Learn. Analytics Res. Beaumont, Alberta, Canada}
}

@misc{WhatIsMMLA,
	author = {Society for Learning Analytics Research (SOLAR)},
	title = {{W}hat is {L}earning {A}nalytics?},
	howpublished = {\url{https://www.solaresearch.org/about/what-is-learning-analytics/}},
    year= {N/A},
	note = {[Accessed 07-02-2024]},
}

@article{ochoa2017multimodal,
  title={Multimodal learning analytics},
  author={Ochoa, Xavier and Lang, AWDG Charles and Siemens, George},
  journal={The handbook of learning analytics},
  volume={1},
  pages={129--141},
  year={2017},
  publisher={Society for Learning Analytics Research Beaumont, AB, Canada}
}

@article{blikstein2016multimodal,
  title={Multimodal learning analytics and education data mining: Using computational technologies to measure complex learning tasks},
  author={Blikstein, Paulo and Worsley, Marcelo},
  journal={Journal of Learning Analytics},
  volume={3},
  number={2},
  pages={220--238},
  year={2016}
}

@inproceedings{blikstein2013multimodal,
  title={Multimodal learning analytics},
  publisher={Association for Computing Machinery},
  address={New York, NY, USA},
  author={Blikstein, Paulo},
  booktitle={Proceedings of the third international conference on learning analytics and knowledge},
  pages={102--106},
  year={2013}
}

@article{leelawong2008designing,
  title={Designing learning by teaching agents: The Betty's Brain system},
  author={Leelawong, Krittaya and Biswas, Gautam},
  journal={International Journal of Artificial Intelligence in Education},
  volume={18},
  number={3},
  pages={181--208},
  year={2008},
  publisher={IOS Press}
}

@inproceedings{rodriguez2018teacher,
  title={The teacher in the loop: Customizing multimodal learning analytics for blended learning},
  publisher={Association for Computing Machinery},
  address={New York, NY, USA},
  author={Rodr{\'\i}guez-Triana, Mar{\'\i}a Jes{\"u}s and Prieto, Luis P and Mart{\'\i}nez-Mon{\'e}s, Alejandra and Asensio-P{\'e}rez, Juan I and Dimitriadis, Yannis},
  booktitle={Proceedings of the 8th international conference on learning analytics and knowledge},
  pages={417--426},
  year={2018}
}

@inproceedings{Hutchins2023,
author = {Hutchins, Nicole and Biswas, Gautam},
title = {Using Teacher Dashboards to Customize Lesson Plans for a Problem-Based, Middle School STEM Curriculum},
year = {2023},
isbn = {9781450398657},
publisher = {Association for Computing Machinery},
address = {New York, NY, USA},
url = {https://doi.org/10.1145/3576050.3576100},
doi = {10.1145/3576050.3576100},
booktitle = {LAK23: 13th International Learning Analytics and Knowledge Conference},
pages = {324–332},
numpages = {9},
location = {Arlington, TX, USA},
series = {LAK2023}
}

@article{maseleno2018demystifying,
  title={Demystifying learning analytics in personalised learning},
  author={Maseleno, Andino and Sabani, Noraisikin and Huda, Miftachul and Ahmad, Roslee Bin and Jasmi, Kamarul Azmi and Basiron, Bushrah},
  journal={International Journal of Engineering and Technology (UAE)},
  year={2018},
  volume={7},
  pages={1124-1129},
  publisher={IJET Publsiher}
}

@article{Zilvinskis2017,
author = {Zilvinskis, John and Willis III, James and Borden, Victor M. H.},
title = {An Overview of Learning Analytics},
journal = {New Directions for Higher Education},
volume = {2017},
number = {179},
pages = {9-17},
doi = {https://doi.org/10.1002/he.20239},
url = {https://onlinelibrary.wiley.com/doi/abs/10.1002/he.20239},
eprint = {https://onlinelibrary.wiley.com/doi/pdf/10.1002/he.20239},
year = {2017}
}

@article{thomas2003general,
    author = {David R. Thomas},
    title ={A General Inductive Approach for Analyzing Qualitative Evaluation Data},
    journal = {American Journal of Evaluation},
    volume = {27},
    number = {2},
    pages = {237-246},
    year = {2006},
    doi = {10.1177/1098214005283748},
    URL = {https://doi.org/10.1177/1098214005283748},
    eprint = {https://doi.org/10.1177/1098214005283748}
}

@article{snyder2025using,
  title={Using collaborative interactivity metrics to analyze students' problem-solving behaviors during STEM+ C computational modeling tasks},
  author={Snyder, Caitlin and Cohn, Clayton and Fonteles, Joyce Horn and Biswas, Gautam},
  journal={Learning and Individual Differences},
  volume={121},
  pages={102724},
  year={2025},
  publisher={Elsevier}
}

@inproceedings{snyder2023analyzing,
  title={Analyzing students collaborative problem-solving behaviors in synergistic STEM+ C learning},
  author={Snyder, Caitlin and Hutchins, Nicole M and Cohn, Clayton and Fonteles, Joyce Horn and Biswas, Gautam},
  booktitle={Proceedings of the 14th Learning Analytics and Knowledge Conference},
  pages={540--550},
  year={2024}
}

@inproceedings{vatral2023prediction,
  title={Prediction of Students’ Self-confidence Using Multimodal Features in an Experiential Nurse Training Environment},
  author={Vatral, Caleb and Lee, Madison and Cohn, Clayton and Davalos, Eduardo and Levin, Daniel and Biswas, Gautam},
  booktitle={International Conference on Artificial Intelligence in Education},
  pages={266--271},
  year={2023},
  organization={Springer}
}

@article{sharma2020multimodal,
  title={Multimodal data capabilities for learning: What can multimodal data tell us about learning?},
  author={Sharma, Kshitij and Giannakos, Michail},
  journal={British Journal of Educational Technology},
  volume={51},
  number={5},
  pages={1450--1484},
  year={2020},
  publisher={Wiley Online Library}
}

@inproceedings{cochran2023bimproving,
  title={Improving NLP model performance on small educational data sets using self-augmentation},
  author={Cochran, Keith and Cohn, Clayton and Hastings, Peter},
  year={2023},
  booktitle = {Proceedings of the 15th International Conference on Computer Supported Education, {CSEDU} 2023, Prague, Czech Republic, April 21-23, 2023, Volume 1},
  url= {https://doi.org/10.5220/0011857200003470},
  publisher={scitepress},
  doi= {10.5220/0011857200003470},
}

@inproceedings{cochran2023improving,
  title={Improving Automated Evaluation of Student Text Responses Using GPT-3.5 for Text Data Augmentation},
  author={Cochran, Keith and Cohn, Clayton and Rouet, Jean Francois and Hastings, Peter},
  booktitle={International Conference on Artificial Intelligence in Education},
  pages={217--228},
  year={2023},
  organization={Springer},
  publisher={N/A},
  address = {N/A}
}

@inproceedings{cochran2022improving,
  title={Improving automated evaluation of formative assessments with text data augmentation},
  author={Cochran, Keith and Cohn, Clayton and Hutchins, Nicole and Biswas, Gautam and Hastings, Peter},
  booktitle={International Conference on Artificial Intelligence in Education},
  pages={390--401},
  year={2022},
  organization={Springer},
  publisher={N/A},
  address = {N/A}
}

@book{cohn2020bert,
  title={BERT efficacy on scientific and medical datasets: a systematic literature review},
  author={Cohn, Clayton},
  year={2020},
  publisher={DePaul University},
  address = {N/A}
}

@misc{nltk,
  author = {Loper, Edward and Bird, Steven},
  copyright = {Assumed arXiv.org perpetual, non-exclusive license to distribute this article for submissions made before January 2004},
  doi = {10.48550/ARXIV.CS/0205028},
  publisher = {arXiv},
  title = {NLTK: The Natural Language Toolkit},
  url = {https://arxiv.org/abs/cs/0205028},
  year = 2002
}

@inproceedings{eyben2010opensmile,
  title={Opensmile: the munich versatile and fast open-source audio feature extractor},
  author={Eyben, Florian and W{\"o}llmer, Martin and Schuller, Bj{\"o}rn},
  booktitle={Proceedings of the 18th ACM international conference on Multimedia},
  pages={1459--1462},
  year={2010},
  publisher={N/A},
  address = {N/A}
}

@article{crossley2019tool,
  title={The Tool for the Automatic Analysis of Cohesion 2.0: Integrating semantic similarity and text overlap},
  author={Crossley, Scott A and Kyle, Kristopher and Dascalu, Mihai},
  journal={Behavior research methods},
  volume={51},
  pages={14--27},
  year={2019},
  publisher={Springer}
}

@article{crossley2016tool,
  title={The tool for the automatic analysis of text cohesion (TAACO): Automatic assessment of local, global, and text cohesion},
  author={Crossley, Scott A and Kyle, Kristopher and McNamara, Danielle S},
  journal={Behavior research methods},
  volume={48},
  pages={1227--1237},
  year={2016},
  publisher={Springer}
}

@article{team2023gemini,
  title={Gemini: a family of highly capable multimodal models},
  author={Team, Gemini and Anil, Rohan and Borgeaud, Sebastian and Wu, Yonghui and Alayrac, Jean-Baptiste and Yu, Jiahui and Soricut, Radu and Schalkwyk, Johan and Dai, Andrew M and Hauth, Anja and others},
  journal={arXiv preprint arXiv:2312.11805},
  year={2023},
  number = {N/A},
  volume = {N/A},
  numpages = {N/A},
  pages = {N/A}
}

@ARTICLE{wei2022chain,
       author = {{Wei}, Jason and {Wang}, Xuezhi and {Schuurmans}, Dale and {Bosma}, Maarten and {Ichter}, Brian and {Xia}, Fei and {Chi}, Ed and {Le}, Quoc and {Zhou}, Denny},
        title = "{Chain-of-Thought Prompting Elicits Reasoning in Large Language Models}",
      journal = {arXiv e-prints},
         year = 2022,
        month = jan,
          eid = {arXiv:2201.11903},
        pages = {arXiv:2201.11903},
          doi = {10.48550/arXiv.2201.11903},
archivePrefix = {arXiv},
       eprint = {2201.11903},
 primaryClass = {cs.CL},
       adsurl = {https://ui.adsabs.harvard.edu/abs/2022arXiv220111903W},
     number = {N/A},
     volume = {N/A},
   numpages = {N/A}
}

@inproceedings{chinh2019ways,
author = {Chinh, Bonnie and Zade, Himanshu and Ganji, Abbas and Aragon, Cecilia},
title = {Ways of Qualitative Coding: A Case Study of Four Strategies for Resolving Disagreements},
year = {2019},
isbn = {9781450359719},
publisher = {Association for Computing Machinery},
address = {New York, NY, USA},
url = {https://doi.org/10.1145/3290607.3312879},
doi = {10.1145/3290607.3312879},
booktitle = {Extended Abstracts of the 2019 CHI Conference on Human Factors in Computing Systems},
pages = {1–6},
numpages = {6},
location = {Glasgow, Scotland Uk},
series = {CHI EA '19}
}

@article{braun2006using,
  title={Using thematic analysis in psychology},
  author={Braun, Virginia and Clarke, Victoria},
  journal={Qualitative research in psychology},
  volume={3},
  number={2},
  pages={77},
  year={2006},
  publisher={Taylor \& Francis Ltd.}
}

@article{koedinger1997intelligent,
  title={Intelligent tutoring goes to school in the big city},
  author={Koedinger, Kenneth R and Anderson, John R and Hadley, William H and Mark, Mary A and others},
  journal={International Journal of Artificial Intelligence in Education},
  volume={8},
  number={1},
  pages={30--43},
  year={1997}
}

@software{cholewiak2021scholarly,
  author  = {Cholewiak, Steven A. and Ipeirotis, Panos and Silva, Victor and Kannawadi, Arun},
  title   = {{SCHOLARLY: Simple access to Google Scholar authors and citation using Python}},
  year    = {2021},
  doi     = {10.5281/zenodo.5764801},
  license = {Unlicense},
  url = {https://github.com/scholarly-python-package/scholarly},
  version = {1.5.1},
  organization={N/A}
}

@misc{gscholar,
	author = {Bastian Venthur},
	title = {{G}it{H}ub - venthur/gscholar: {Q}uery {G}oogle {S}cholar with {P}ython},
	howpublished = {\url{https://github.com/venthur/gscholar}},
	year = {2010},
	note = {[Accessed 08-02-2024]},
}

@misc{serpapi,
	author = {SerpApi},
	title = {{G}oogle {S}cholar {A}{P}{I}},
	howpublished = {\url{https://serpapi.com/google-scholar-api}},
	year = {N/A},
	note = {[Accessed 08-02-2024]},
}

@misc{spacy_fastlang,
	author = {{T}homas {T}hiebaud"},
	title = {{S}pacy {F}ast{L}ang},
	howpublished = {\url{https://spacy.io/universe/project/spacy_fastlang}},
	year = {2020},
	note = {[Accessed 08-02-2024]},
}

@inproceedings{cohn2024human,
  title={Towards a human-in-the-loop LLM approach to collaborative discourse analysis},
  author={Cohn, Clayton and Snyder, Caitlin and Montenegro, Justin and Biswas, Gautam},
  booktitle={International Conference on Artificial Intelligence in Education},
  pages={11--19},
  year={2024},
  organization={Springer}
}

@unpublished{cohn2024chain,
  author = "Cohn, Clayton and T S, Ashwin and Mohammed, Naveeduddin and Biswas, Gautam",
  title  = "CoTAL: Human-in-the-Loop Prompt Engineering for Generalizable Formative Assessment Scoring",
  note   = "Submitted to the International Journal of Artificial Intelligence in Education (IJAIED). Currently under review",
  year   = "2025",
  url = "https://arxiv.org/abs/2504.02323"
}

@article{cohn2024multimodal,
  title={A multimodal approach to support teacher, researcher and AI collaboration in STEM+ C learning environments},
  author={Cohn, Clayton and Snyder, Caitlin and Fonteles, Joyce Horn and TS, Ashwin and Montenegro, Justin and Biswas, Gautam},
  journal={British Journal of Educational Technology},
  volume={56},
  number={2},
  pages={595--620},
  year={2025},
  publisher={Wiley Online Library}
}

@InProceedings{fonteles2024aied,
author="Fonteles, Joyce
and Davalos, Eduardo
and Ashwin, T. S.
and Zhang, Yike
and Zhou, Mengxi
and Ayalon, Efrat
and Lane, Alicia
and Steinberg, Selena
and Anton, Gabriella
and Danish, Joshua
and Enyedy, Noel
and Biswas, Gautam",
editor="Olney, Andrew M.
and Chounta, Irene-Angelica
and Liu, Zitao
and Santos, Olga C.
and Bittencourt, Ig Ibert",
title="A First Step in Using Machine Learning Methods to Enhance Interaction Analysis for Embodied Learning Environments",
booktitle="Artificial Intelligence in Education",
year="2024",
publisher="Springer Nature Switzerland",
address="Cham",
pages="3--16",
}

@inproceedings{lixiang2022scalability,
    author = {Yan, Lixiang and Zhao, Linxuan and Gasevic, Dragan and Martinez-Maldonado, Roberto},
    title = {Scalability, Sustainability, and Ethicality of Multimodal Learning Analytics},
    year = {2022},
    isbn = {9781450395731},
    publisher = {Association for Computing Machinery},
    address = {New York, NY, USA},
    url = {https://doi.org/10.1145/3506860.3506862},
    doi = {10.1145/3506860.3506862},
    booktitle = {LAK22: 12th International Learning Analytics and Knowledge Conference},
    pages = {13–23},
    numpages = {11},
    location = {Online, USA},
    series = {LAK22}
}

@article{boulton2019student,
    doi = {10.1371/journal.pone.0225770},
    author = {Boulton, Chris A. AND Hughes, Emily AND Kent, Carmel AND Smith, Joanne R. AND Williams, Hywel T. P.},
    journal = {PLOS ONE},
    publisher = {Public Library of Science},
    title = {Student engagement and wellbeing over time at a higher education institution},
    year = {2019},
    month = {11},
    volume = {14},
    url = {https://doi.org/10.1371/journal.pone.0225770},
    pages = {1-20},
    number = {11},

}

@article{cukurova2020promise,
  title={The promise and challenges of multimodal learning analytics},
  author={Cukurova, Mutlu and Giannakos, Michail and Martinez-Maldonado, Roberto},
  journal={British Journal of Educational Technology},
  volume={51},
  number={5},
  pages={1441--1449},
  year={2020},
  publisher={Wiley}
}

@article{cochran2023using,
  title={Using BERT to Identify Causal Structure in Students’ Scientific Explanations},
  author={Cochran, Keith and Cohn, Clayton and Hastings, Peter and Tomuro, Noriko and Hughes, Simon},
  journal={International Journal of Artificial Intelligence in Education},
  pages={1--39},
  year={2023},
  publisher={Springer},
  volume={N/A},
  number={N/A},
}

@inproceedings{sarmiento2022participatory,
author = {Sarmiento, Juan Pablo and Wise, Alyssa Friend},
title = {Participatory and Co-Design of Learning Analytics: An Initial Review of the Literature},
year = {2022},
isbn = {9781450395731},
publisher = {Association for Computing Machinery},
address = {New York, NY, USA},
url = {https://doi.org/10.1145/3506860.3506910},
doi = {10.1145/3506860.3506910},
booktitle = {LAK22: 12th International Learning Analytics and Knowledge Conference},
pages = {535–541},
numpages = {7},
location = {Online, USA},
series = {LAK22}
}

@inproceedings{Fonteles2024Underserved,
author = {Fonteles, Joyce Horn and Akpanoko, Celestine E and Wisniewski, Pamela J. and Biswas, Gautam},
title = {Promoting Equitable Learning Outcomes for Underserved Students in Open-Ended Learning Environments},
year = {2024},
isbn = {9798400704420},
publisher = {Association for Computing Machinery},
address = {New York, NY, USA},
url = {https://doi.org/10.1145/3628516.3655753},
doi = {10.1145/3628516.3655753},
booktitle = {Proceedings of the 23rd Annual ACM Interaction Design and Children Conference},
pages = {307–321},
numpages = {15},
location = {Delft, Netherlands},
series = {IDC '24}
}

@article{networkX,
    title = {Exploring network structure, dynamics, and function using NetworkX},
    author = {Hagberg, Aric and Swart, Pieter J. and Schult, Daniel A.},
    doi = {},
    url = {https://www.osti.gov/biblio/960616}, journal = {N/A},
    number = {N/A},
    volume = {N/A},
    pages = {N/A},
    place = {United States},
    year = {2008},
    month = {1}
}

@article{noble2015bias,
  title = {Issues of validity and reliability in qualitative research},
  volume = {18},
  ISSN = {1468-9618},
  DOI = {10.1136/eb-2015-102054},
  number = {2},
  journal = {Evidence Based Nursing},
  publisher = {BMJ},
  author = {Noble,  Helen and Smith,  Joanna},
  year = {2015},
  month = feb,
  pages = {34–35}
}

@article{mehra2015bias,
  title = {Bias in Qualitative Research: Voices from an Online Classroom},
  ISSN = {1052-0147},
  DOI = {10.46743/2160-3715/2002.1986},
  journal = {The Qualitative Report},
  publisher = {Nova Southeastern University},
  author = {Mehra,  Beloo},
  year = {2015},
  month = {Jan},
  numpages = {N/A},
  number = {N/A},
  volume = {N/A}
}

@article{giannakos2023role,
  title={The role of learning theory in multimodal learning analytics},
  author={Giannakos, Michail and Cukurova, Mutlu},
  journal={British Journal of Educational Technology},
  volume={54},
  number={5},
  pages={1246--1267},
  year={2023},
  publisher={Wiley Online Library}
}

@article{zhu2024review,
  title={A review of key technologies for emotion analysis using multimodal information},
  author={Zhu, Xianxun and Guo, Chaopeng and Feng, Heyang and Huang, Yao and Feng, Yichen and Wang, Xiangyang and Wang, Rui},
  journal={Cognitive Computation},
  volume={16},
  number={4},
  pages={1504--1530},
  year={2024},
  publisher={Springer}
}

@inproceedings{yan2024generative,
  title={Generative artificial intelligence in learning analytics: Contextualising opportunities and challenges through the learning analytics cycle},
  author={Yan, Lixiang and Martinez-Maldonado, Roberto and Gasevic, Dragan},
  booktitle={Proceedings of the 14th learning analytics and knowledge conference},
  pages={101--111},
  year={2024}
}

@inproceedings{liu2024llava,
  title={Llava-plus: Learning to use tools for creating multimodal agents},
  author={Liu, Shilong and Cheng, Hao and Liu, Haotian and Zhang, Hao and Li, Feng and Ren, Tianhe and Zou, Xueyan and Yang, Jianwei and Su, Hang and Zhu, Jun and others},
  booktitle={European conference on computer vision},
  pages={126--142},
  year={2024},
  organization={Springer}
}

@article{fonteles2026JLI,
  title={Analyzing Embodied Learning in Classroom Settings: A Human-in-the-Loop AI Approach for Multimodal Learning Analytics},
  author={Fonteles, Joyce-Horn and Cohn, Clayton and .... and Biswas, Gautam},
  journal={Journal of Learning and Instruction},
 note = {in press, special issue on Implementing Multimodal Learning Analytics (MMLA) in Ecological Settings for Generating Actionable Insights},
  year={2026},
  organization={Elsevier}
}

@article{fonteles2025exploring,
  title={Exploring Agentic Multimodal Late Fusion With LLMs for Embodied Learning},
  author={Fonteles, Joyce Horn and Cohn, Clayton and Mereddy, Divya and Ashwin, TS and Biswas, Gautam},
  journal={Planning},
  volume={7},
  number={15},
  pages={46--7},
  year={2025}
}

@article{li2024multimodal,
  title={Multimodal alignment and fusion: A survey},
  author={Li, Songtao and Tang, Hao},
  journal={arXiv preprint arXiv:2411.17040},
  year={2024}
}

@article{begum2024federated,
  title={Federated and multi-modal learning algorithms for healthcare and cross-domain analytics},
  author={Begum, Ummal Sariba},
  journal={PatternIQ Mining},
  volume={1},
  number={4},
  pages={38--51},
  year={2024}
}

@article{ma2024multilevel,
  title={A multilevel multimodal fusion transformer for remote sensing semantic segmentation},
  author={Ma, Xianping and Zhang, Xiaokang and Pun, Man-On and Liu, Ming},
  journal={IEEE Transactions on Geoscience and Remote Sensing},
  volume={62},
  pages={1--15},
  year={2024},
  publisher={IEEE}
}

@article{fang2022multi,
  title={Multi-modal cross-domain alignment network for video moment retrieval},
  author={Fang, Xiang and Liu, Daizong and Zhou, Pan and Hu, Yuchong},
  journal={IEEE Transactions on Multimedia},
  volume={25},
  pages={7517--7532},
  year={2022},
  publisher={IEEE}
}

@article{martinez2023lessons,
  title={Lessons learnt from a multimodal learning analytics deployment in-the-wild},
  author={Martinez-Maldonado, Roberto and Echeverria, Vanessa and Fernandez-Nieto, Gloria and Yan, Lixiang and Zhao, Linxuan and Alfredo, Riordan and Li, Xinyu and Dix, Samantha and Jaggard, Hollie and Wotherspoon, Rosie and others},
  journal={ACM Transactions on Computer-Human Interaction},
  volume={31},
  number={1},
  pages={1--41},
  year={2023},
  publisher={ACM New York, NY}
}

@inproceedings{schneider2018multimodal,
  title={Multimodal learning hub: A tool for capturing customizable multimodal learning experiences},
  author={Schneider, Jan and Di Mitri, Daniele and Limbu, Bibeg and Drachsler, Hendrik},
  booktitle={European conference on technology enhanced learning},
  pages={45--58},
  year={2018},
  organization={Springer}
}

@article{kinnebrew2013contextualized,
  title={A contextualized, differential sequence mining method to derive students' learning behavior patterns.},
  author={Kinnebrew, John S and Loretz, Kirk M and Biswas, Gautam},
  journal={Journal of Educational Data Mining},
  volume={5},
  number={1},
  pages={190--219},
  year={2013},
  publisher={ERIC}
}

@incollection{azevedo2013using,
  title={Using trace data to examine the complex roles of cognitive, metacognitive, and emotional self-regulatory processes during learning with multi-agent systems},
  author={Azevedo, Roger and Harley, Jason and Trevors, Gregory and Duffy, Melissa and Feyzi-Behnagh, Reza and Bouchet, Fran{\c{c}}ois and Landis, Ronald},
  booktitle={International handbook of metacognition and learning technologies},
  pages={427--449},
  year={2013},
  publisher={Springer}
}

@article{vanlehn2006behavior,
  title={The behavior of tutoring systems},
  author={VanLehn, Kurt},
  journal={International journal of artificial intelligence in education},
  volume={16},
  number={3},
  pages={227--265},
  year={2006},
  publisher={SAGE Publications Sage UK: London, England}
}

@article{liang2024foundations,
  title={Foundations \& trends in multimodal machine learning: Principles, challenges, and open questions},
  author={Liang, Paul Pu and Zadeh, Amir and Morency, Louis-Philippe},
  journal={ACM Computing Surveys},
  volume={56},
  number={10},
  pages={1--42},
  year={2024},
  publisher={ACM New York, NY}
}

@article{soenksen2022integrated,
  title={Integrated multimodal artificial intelligence framework for healthcare applications},
  author={Soenksen, Luis R and Ma, Yu and Zeng, Cynthia and Boussioux, Leonard and Villalobos Carballo, Kimberly and Na, Liangyuan and Wiberg, Holly M and Li, Michael L and Fuentes, Ignacio and Bertsimas, Dimitris},
  journal={NPJ digital medicine},
  volume={5},
  number={1},
  pages={149},
  year={2022},
  publisher={Nature Publishing Group UK London}
}

@article{zheng2023judging,
  title={Judging llm-as-a-judge with mt-bench and chatbot arena},
  author={Zheng, Lianmin and Chiang, Wei-Lin and Sheng, Ying and Zhuang, Siyuan and Wu, Zhanghao and Zhuang, Yonghao and Lin, Zi and Li, Zhuohan and Li, Dacheng and Xing, Eric and others},
  journal={Advances in neural information processing systems},
  volume={36},
  pages={46595--46623},
  year={2023}
}

@article{cohn2025theory,
  title={A theory of adaptive scaffolding for LLM-based pedagogical agents},
  author={Cohn, Clayton and Rayala, Surya and Srivastava, Namrata and Fonteles, Joyce Horn and Jain, Shruti and Luo, Xinying and Mereddy, Divya and Mohammed, Naveeduddin and Biswas, Gautam},
  journal={arXiv preprint arXiv:2508.01503},
  year={2025}
}

@article{cohn2025personalizing,
  title={Personalizing Student-Agent Interactions Using Log-Contextualized Retrieval Augmented Generation (RAG)},
  author={Cohn, Clayton and Rayala, Surya and Snyder, Caitlin and Fonteles, Joyce and Jain, Shruti and Mohammed, Naveeduddin and Timalsina, Umesh and Burriss, Sarah K and Srivastava, Namrata and Deweese, Menton and others},
  journal={arXiv preprint arXiv:2505.17238},
  year={2025}
}

@inproceedings{cohn2025exploring,
  title={Exploring the design of pedagogical agent roles in collaborative stem+ c learning},
  author={Cohn, Clayton and Fonteles, Joyce Horn and Snyder, Caitlin and Srivastava, Namrata and Campbell, Desmond and Montenegro, Justin and Biswas, Gautam and others},
  booktitle={Proceedings of the 18th International Conference on Computer-Supported Collaborative Learning-CSCL 2025, pp. 330-334},
  year={2025},
  organization={International Society of the Learning Sciences}
}

@inproceedings{di2020real,
  title={Real-time multimodal feedback with the CPR tutor},
  author={Di Mitri, Daniele and Schneider, Jan and Trebing, Kevin and Sopka, Sasa and Specht, Marcus and Drachsler, Hendrik},
  booktitle={International conference on artificial intelligence in education},
  pages={141--152},
  year={2020},
  organization={Springer}
}

@inproceedings{snyder2024investigating,
  title={Investigating collaborative problem solving behaviors during stem+ c learning in groups with different prior knowledge distributions},
  author={Snyder, Caitlin and Wen, Cai-Ting and Hutchins, Nicole M and Vatral, Caleb and Liu, Chen-Chung and Biswas, Gautam},
  booktitle={Proceedings of the 17th International Conference on Computer-Supported Collaborative Learning-CSCL 2024, pp. 107-114},
  year={2024},
  organization={International Society of the Learning Sciences}
}

@inproceedings{yan2024vizchat,
  title={VizChat: enhancing learning analytics dashboards with contextualised explanations using multimodal generative AI chatbots},
  author={Yan, Lixiang and Zhao, Linxuan and Echeverria, Vanessa and Jin, Yueqiao and Alfredo, Riordan and Li, Xinyu and Ga{\v{s}}evi’c, Dragan and Martinez-Maldonado, Roberto},
  booktitle={International conference on artificial intelligence in education},
  pages={180--193},
  year={2024},
  organization={Springer}
}

@article{acosta2024multimodal,
  title={Multimodal learning analytics for predicting student collaboration satisfaction in collaborative game-based learning},
  author={Acosta, Halim and Lee, Seung and Mott, Bradford and Bae, Haesol and Glazewski, Krista and Hmelo-Silver, Cindy and Lester, James},
  year={2024},
  publisher={International Educational Data Mining Society}
}

@inproceedings{baltruvsaitis2016openface,
  title={Openface: an open source facial behavior analysis toolkit},
  author={Baltru{\v{s}}aitis, Tadas and Robinson, Peter and Morency, Louis-Philippe},
  booktitle={2016 IEEE winter conference on applications of computer vision (WACV)},
  pages={1--10},
  year={2016},
  organization={IEEE}
}

@inproceedings{ashwin2025challenges,
  title={Challenges of Applying Computer Vision for Emotion Detection in Educational Settings: A Study on Bias},
  author={Ashwin, TS and Sanda, Nihar and Timalsina, Umesh and Biswas, Gautam},
  booktitle={International Conference on Artificial Intelligence in Education},
  pages={388--395},
  year={2025},
  organization={Springer}
}

@article{que2025using,
  title={Using eye movements, electrodermal activities, and heart rates to predict different types of cognitive load during reading with background music},
  author={Que, Ying and Zheng, Yueyuan and Hsiao, Janet H and Hu, Xiao},
  journal={Scientific Reports},
  volume={15},
  number={1},
  pages={32635},
  year={2025},
  publisher={Nature Publishing Group UK London}
}

@article{sung2024beyond,
  title={Beyond frequency: Using epistemic network analysis and multimodal traces to understand temporal dynamics of self-regulated learning},
  author={Sung, Hanall and Bernacki, Matthew L and Greene, Jeffrey A and Yu, Linyu and Plumley, Robert D},
  journal={Journal of Science Education and Technology},
  pages={1--18},
  year={2024},
  publisher={Springer}
}

@phdthesis{snyder2024understanding,
  title={Understanding Students' Collaborative Problem Solving during STEM+ C Learning using Multimodal Analysis},
  author={Snyder, Caitlin},
  year={2024},
  school={Vanderbilt University}
}

@article{whitehead2025utilizing,
  title={Utilizing multimodal large language models for video analysis of posture in studying collaborative learning: A case study},
  author={Whitehead, Ridwan and Nguyen, Andy and J{\"a}rvel{\"a}, Sanna},
  journal={Journal of Learning Analytics},
  volume={12},
  number={1},
  pages={186--200},
  year={2025}
}

@article{ke2016teaching,
  title={Teaching training in a mixed-reality integrated learning environment},
  author={Ke, Fengfeng and Lee, Sungwoong and Xu, Xinhao},
  journal={Computers in Human Behavior},
  volume={62},
  pages={212--220},
  year={2016},
  publisher={Elsevier}
}

@article{kaplan2021effects,
  title={The effects of virtual reality, augmented reality, and mixed reality as training enhancement methods: A meta-analysis},
  author={Kaplan, Alexandra D and Cruit, Jessica and Endsley, Mica and Beers, Suzanne M and Sawyer, Ben D and Hancock, Peter A},
  journal={Human factors},
  volume={63},
  number={4},
  pages={706--726},
  year={2021},
  publisher={Sage Publications Sage CA: Los Angeles, CA}
}

@article{algerafi2023unlocking,
  title={Unlocking the potential: A comprehensive evaluation of augmented reality and virtual reality in education},
  author={AlGerafi, Mohammed AM and Zhou, Yueliang and Oubibi, Mohamed and Wijaya, Tommy Tanu},
  journal={Electronics},
  volume={12},
  number={18},
  pages={3953},
  year={2023},
  publisher={MDPI}
}

@article{gomez2023confederacy,
  title={A confederacy of models: A comprehensive evaluation of LLMs on creative writing},
  author={G{\'o}mez-Rodr{\'\i}guez, Carlos and Williams, Paul},
  journal={arXiv preprint arXiv:2310.08433},
  year={2023}
}

@article{kim2025evaluating,
  title={Evaluating Creativity: Can LLMs Be Good Evaluators in Creative Writing Tasks?},
  author={Kim, Sungeun and Oh, Dongsuk},
  journal={Applied Sciences},
  volume={15},
  number={6},
  pages={2971},
  year={2025},
  publisher={MDPI}
}

@article{amershi2014power,
  title={Power to the people: The role of humans in interactive machine learning},
  author={Amershi, Saleema and Cakmak, Maya and Knox, William Bradley and Kulesza, Todd},
  journal={AI magazine},
  volume={35},
  number={4},
  pages={105--120},
  year={2014}
}

@article{zha2025data,
  title={Data-centric artificial intelligence: A survey},
  author={Zha, Daochen and Bhat, Zaid Pervaiz and Lai, Kwei-Herng and Yang, Fan and Jiang, Zhimeng and Zhong, Shaochen and Hu, Xia},
  journal={ACM Computing Surveys},
  volume={57},
  number={5},
  pages={1--42},
  year={2025},
  publisher={ACM New York, NY}
}

@article{song2022learning,
  title={Learning from noisy labels with deep neural networks: A survey},
  author={Song, Hwanjun and Kim, Minseok and Park, Dongmin and Shin, Yooju and Lee, Jae-Gil},
  journal={IEEE transactions on neural networks and learning systems},
  volume={34},
  number={11},
  pages={8135--8153},
  year={2022},
  publisher={IEEE}
}

@article{shamir2021facilitating,
  title={Facilitating emergency remote K-12 teaching in computing-enhanced virtual learning environments during COVID-19 pandemic-blessing or curse?},
  author={Shamir-Inbal, Tamar and Blau, Ina},
  journal={Journal of Educational Computing Research},
  volume={59},
  number={7},
  pages={1243--1271},
  year={2021},
  publisher={Sage Publications Sage CA: Los Angeles, CA}
}

@article{gokccearslan2024benefits,
  title={Benefits, challenges, and methods of artificial intelligence (AI) chatbots in education: A systematic literature review.},
  author={G{\"o}k{\c{c}}earslan, Sahin and Tosun, Cansel and Erdemir, Zeynep Gizem},
  journal={International Journal of Technology in Education},
  volume={7},
  number={1},
  pages={19--39},
  year={2024},
  publisher={ERIC}
}

@article{poria2016fusing,
  title={Fusing audio, visual and textual clues for sentiment analysis from multimodal content},
  author={Poria, Soujanya and Cambria, Erik and Howard, Newton and Huang, Guang-Bin and Hussain, Amir},
  journal={Neurocomputing},
  volume={174},
  pages={50--59},
  year={2016},
  publisher={Elsevier}
}

@article{noroozi2019multimodal,
  title={Multimodal data to design visual learning analytics for understanding regulation of learning},
  author={Noroozi, Omid and Alikhani, Iman and J{\"a}rvel{\"a}, Sanna and Kirschner, Paul A and Juuso, Ilkka and Sepp{\"a}nen, Tapio},
  journal={Computers in Human Behavior},
  volume={100},
  pages={298--304},
  year={2019},
  publisher={Elsevier}
}

@article{prinsloo2022answer,
  title={The answer is (not only) technological: Considering student data privacy in learning analytics},
  author={Prinsloo, Paul and Slade, Sharon and Khalil, Mohammad},
  journal={British Journal of Educational Technology},
  volume={53},
  number={4},
  pages={876--893},
  year={2022},
  publisher={Wiley Online Library}
}

@article{marin2023social,
  title={Social media and data privacy in education: An international comparative study of perceptions among pre-service teachers},
  author={Mar{\'\i}n, Victoria I and Carpenter, Jeffrey P and Tur, Gemma and Williamson-Leadley, Sandra},
  journal={Journal of Computers in Education},
  volume={10},
  number={4},
  pages={769--795},
  year={2023},
  publisher={Springer}
}

@inproceedings{harvey2025don,
  title={" Don't Forget the Teachers": Towards an Educator-Centered Understanding of Harms from Large Language Models in Education},
  author={Harvey, Emma and Koenecke, Allison and Kizilcec, Rene F},
  booktitle={Proceedings of the 2025 CHI Conference on Human Factors in Computing Systems},
  pages={1--19},
  year={2025}
}

@misc{mintz2023artificial,
  title={Artificial Intelligence and K-12 education: Possibilities, pedagogies and risks},
  author={Mintz, Joseph and Holmes, Wayne and Liu, Leping and Perez-Ortiz, Maria},
  journal={Computers in the Schools},
  volume={40},
  number={4},
  pages={325--333},
  year={2023},
  publisher={Taylor \& Francis}
}

@article{ober2021linking,
  title={Linking self-report and process data to performance as measured by different assessment types},
  author={Ober, Teresa M and Hong, Maxwell R and Rebou{\c{c}}as-Ju, Daniella A and Carter, Matthew F and Liu, Cheng and Cheng, Ying},
  journal={Computers \& Education},
  volume={167},
  pages={104188},
  year={2021},
  publisher={Elsevier}
}

@misc{anthropic2025agentSkills,
  author = {Anthropic},
  title = {Equipping Agents for the Real World with Agent Skills},
  howpublished = {Anthropic Engineering Blog},
  year = {2025},
  month = {Oct},
  day = {16},
  url = {https://www.anthropic.com/engineering/equipping-agents-for-the-real-world-with-agent-skills},
  note = {Accessed: 2025-12-14}
}

@misc{anthropic2024modelContext,
  author = {Anthropic},
  title = {Introducing the Model Context Protocol},
  howpublished = {Anthropic News},
  year = {2024},
  month = {Nov},
  day = {25},
  url = {https://www.anthropic.com/news/model-context-protocol},
  note = {Accessed: 2025-12-14}
}

@article{kosmyna2025your,
  title={Your brain on ChatGPT: Accumulation of cognitive debt when using an AI assistant for essay writing task},
  author={Kosmyna, Nataliya and Hauptmann, Eugene and Yuan, Ye Tong and Situ, Jessica and Liao, Xian-Hao and Beresnitzky, Ashly Vivian and Braunstein, Iris and Maes, Pattie},
  journal={arXiv preprint arXiv:2506.08872},
  volume={4},
  year={2025}
}

@inproceedings{zhou2025impact,
  author    = {Yiqiu Zhou and Maciej Pankiewicz and Luc Paquette and Ryan S. Baker},
  title     = {Impact of {LLM} Feedback on Learner Persistence in Programming},
  booktitle = {Proceedings of the 33rd International Conference on Computers in Education (ICCE 2025)},
  year      = {2025},
  publisher = {Asia-Pacific Society for Computers in Education},
  note      = {To appear}
}

@inproceedings{thomas2025llm,
  title={Llm-generated feedback supports learning if learners choose to use it},
  author={Thomas, Danielle R and Borchers, Conrad and Bhushan, Shambhavi and Gatz, Erin and Gupta, Shivang and Koedinger, Kenneth R},
  booktitle={European Conference on Technology Enhanced Learning},
  pages={489--503},
  year={2025},
  organization={Springer}
}

@incollection{janssen2013multilevel,
  title={Multilevel analysis for the analysis of collaborative learning},
  author={Janssen, Jeroen and Cress, Ulrike and Erkens, Gijsbert and Kirschner, Paul A},
  booktitle={The international handbook of collaborative learning},
  pages={112--125},
  year={2013},
  publisher={Routledge}
}

@inproceedings{blanchard2015study,
  title={A study of automatic speech recognition in noisy classroom environments for automated dialog analysis},
  author={Blanchard, Nathaniel and Brady, Michael and Olney, Andrew M and Glaus, Marci and Sun, Xiaoyi and Nystrand, Martin and Samei, Borhan and Kelly, Sean and D’Mello, Sidney},
  booktitle={International conference on artificial intelligence in education},
  pages={23--33},
  year={2015},
  organization={Springer}
}

@article{jonassen2006role,
  title={On the role of concepts in learning and instructional design},
  author={Jonassen, David H},
  journal={Educational Technology Research and Development},
  volume={54},
  number={2},
  pages={177--196},
  year={2006},
  publisher={Springer}
}

@article{doering2015fostering,
  title={Fostering creativity through inquiry and adventure in informal learning environment design},
  author={Doering, Aaron and Henrickson, Jeni},
  journal={Journal of Technology and Teacher Education},
  volume={23},
  number={3},
  pages={387--410},
  year={2015},
  publisher={Society for Information Technology \& Teacher Education}
}

@article{jiao2025llms,
  title={LLMs and Childhood Safety: Identifying Risks and Proposing a Protection Framework for Safe Child-LLM Interaction},
  author={Jiao, Junfeng and Afroogh, Saleh and Chen, Kevin and Murali, Abhejay and Atkinson, David and Dhurandhar, Amit},
  journal={arXiv preprint arXiv:2502.11242},
  year={2025}
}

@article{chang2024survey,
  title={A survey on evaluation of large language models},
  author={Chang, Yupeng and Wang, Xu and Wang, Jindong and Wu, Yuan and Yang, Linyi and Zhu, Kaijie and Chen, Hao and Yi, Xiaoyuan and Wang, Cunxiang and Wang, Yidong and others},
  journal={ACM transactions on intelligent systems and technology},
  volume={15},
  number={3},
  pages={1--45},
  year={2024},
  publisher={ACM New York, NY}
}

@book{bransford2000people,
  title={How people learn},
  author={Bransford, John D and Brown, Ann L and Cocking, Rodney R and others},
  volume={11},
  year={2000},
  publisher={Washington, DC: National academy press}
}

@article{english2013supporting,
  title={Supporting student self-regulated learning in problem-and project-based learning},
  author={English, Mary C and Kitsantas, Anastasia},
  journal={Interdisciplinary journal of problem-based learning},
  volume={7},
  number={2},
  pages={6},
  year={2013},
  publisher={Purdue University Press}
}

@article{si2022multimodal,
  title={Multimodal literacies classroom instruction for K-12 students: a review of research},
  author={Si, Qi and Hodges, Tracey S and Coleman, Julianne M},
  journal={Literacy Research and Instruction},
  volume={61},
  number={3},
  pages={276--297},
  year={2022},
  publisher={Taylor \& Francis}
}

@article{dindar2020matching,
  title={Matching self-reports with electrodermal activity data: Investigating temporal changes in self-regulated learning},
  author={Dindar, Muhterem and Malmberg, Jonna and J{\"a}rvel{\"a}, Sanna and Haataja, Eetu and Kirschner, Paul A},
  journal={Education and Information Technologies},
  volume={25},
  number={3},
  pages={1785--1802},
  year={2020},
  publisher={Springer}
}

@article{zhou2024detecting,
  title={Detecting non-verbal speech and gaze behaviours with multimodal data and computer vision to interpret effective collaborative learning interactions},
  author={Zhou, Qi and Suraworachet, Wannapon and Cukurova, Mutlu},
  journal={Education and information technologies},
  volume={29},
  number={1},
  pages={1071--1098},
  year={2024},
  publisher={Springer}
}

@article{achiam2023gpt,
  title={Gpt-4 technical report},
  author={Achiam, Josh and Adler, Steven and Agarwal, Sandhini and Ahmad, Lama and Akkaya, Ilge and Aleman, Florencia Leoni and Almeida, Diogo and Altenschmidt, Janko and Altman, Sam and Anadkat, Shyamal and others},
  journal={arXiv preprint arXiv:2303.08774},
  year={2023}
}

@article{huang2020pixel,
  title={Pixel-bert: Aligning image pixels with text by deep multi-modal transformers},
  author={Huang, Zhicheng and Zeng, Zhaoyang and Liu, Bei and Fu, Dongmei and Fu, Jianlong},
  journal={arXiv preprint arXiv:2004.00849},
  year={2020}
}

@article{colyer2018review,
  title={A review of the evolution of vision-based motion analysis and the integration of advanced computer vision methods towards developing a markerless system},
  author={Colyer, Steffi L and Evans, Murray and Cosker, Darren P and Salo, Aki IT},
  journal={Sports medicine-open},
  volume={4},
  number={1},
  pages={24},
  year={2018},
  publisher={Springer}
}

@inproceedings{worsley2012multimodal,
  title={Multimodal learning analytics: enabling the future of learning through multimodal data analysis and interfaces},
  author={Worsley, Marcelo},
  booktitle={Proceedings of the 14th ACM international conference on Multimodal interaction},
  pages={353--356},
  year={2012}
}

@book{shaffer2017quantitative,
  title={Quantitative ethnography},
  author={Shaffer, DW},
  year={2017},
  publisher={Cathcart Press}
}

@article{shaffer2016tutorial,
  title={A tutorial on epistemic network analysis: Analyzing the structure of connections in cognitive, social, and interaction data},
  author={Shaffer, David Williamson and Collier, Wesley and Ruis, Andrew R},
  journal={Journal of learning analytics},
  volume={3},
  number={3},
  pages={9--45},
  year={2016}
}

@article{nguyen2015don,
  title={Don’t forget about the body: Exploring the curricular possibilities of embodied pedagogy},
  author={Nguyen, David J and Larson, Jay B},
  journal={Innovative Higher Education},
  volume={40},
  number={4},
  pages={331--344},
  year={2015},
  publisher={Springer}
}

@article{hutchins2024co,
  title={Co-designing teacher support technology for problem-based learning in middle school science},
  author={Hutchins, Nicole M and Biswas, Gautam},
  journal={British Journal of Educational Technology},
  volume={55},
  number={3},
  pages={802--822},
  year={2024},
  publisher={Wiley Online Library}
}

@unpublished{timalsinasyncflow,
  title={SyncFlow: A Scalable Platform for Multimodal Learning Analytics},
  author={Timalsina, Umesh and Davalos, Eduardo and Sanda, Nihar Purshottam and Zhang, Yike and Fonteles, Joyce Horn and Ashwin, TS and Biswas, Gautam},
  date={2025},
  url={https://www.researchgate.net/publication/397356107_SyncFlow_A_Scalable_Platform_for_Multimodal_Learning_Analytics}
}

@inproceedings{2181637610,
	address = {London, UK},
	title = {Toward {Using} {Multi}-{Modal} {Learning} {Analytics} to {Support} and {Measure} {Collaboration} in {Co}-{Located} {Dyads}},
	language = {en},
	booktitle = {{ICLS} 2018},
	publisher = {International Society of the Learning Sciences},
	author = {Starr, Emma L and Reilly, Joseph M and Schneider, Bertrand},
	year = {2018},
	pages = {448--455},
}

@inproceedings{1118315889,
	address = {Philadelphia, PA USA},
	title = {Using {Multimodal} {Learning} {Analytics} to {Identify} {Aspects} of {Collaboration} in {Project}-{Based} {Learning}},
	volume = {1},
	language = {en},
	booktitle = {Making a {Difference}: {Prioritizing} {Equity} and {Access} in {CSCL}},
	publisher = {International Society of the Learning Sciences},
	author = {Spikol, Daniel and Ruffaldi, Emanuele and Cukurova, Mutlu},
	year = {2017},
	pages = {263--270},
}

@inproceedings{3308658121,
	address = {Buffalo, NY, USA},
	title = {Exploring {Collaboration} {Using} {Motion} {Sensors} and {Multi}- {Modal} {Learning} {Analytics}},
	language = {en},
	booktitle = {Proceedings of the 11th {International} {Conference} on {Educational} {Data} {Mining}},
	publisher = {International Educational Data Mining Society},
	author = {Reilly, Joseph M and Ravenell, Milan and Schneider, Bertrand},
	month = jul,
	year = {2018},
	pages = {333--339},
}

@inproceedings{85990093,
	address = {Stockholm, Sweden},
	title = {Multimodal {Markers} of {Persuasive} {Speech}: {Designing} a {Virtual} {Debate} {Coach}},
	shorttitle = {Multimodal {Markers} of {Persuasive} {Speech}},
	url = {https://www.isca-speech.org/archive/interspeech_2017/petukhova17_interspeech.html},
	doi = {10.21437/Interspeech.2017-98},
	language = {en},
	urldate = {2023-08-07},
	booktitle = {Interspeech 2017},
	publisher = {ISCA},
	author = {Petukhova, Volha and Raju, Manoj and Bunt, Harry},
	month = aug,
	year = {2017},
	pages = {142--146},
}

@inproceedings{2070224207,
	address = {Poznan, Poland},
	title = {Detecting {Medical} {Simulation} {Errors} with {Machine} learning and {Multimodal} {Data}},
	language = {en},
	booktitle = {17th {Conference} on {Artificial} {Intelligence} in {Medicine}},
	publisher = {Springer International Publishing},
	author = {Mitri, Daniele Di},
	year = {2019},
	pages = {1--6},
}

@inproceedings{518268671,
	address = {Brighton, UK},
	title = {Using {Multimodal} {Learning} {Analytics} to {Explore} {Collaboration} in a {Sustainability} {Co}-located {Tabletop} {Game}},
	language = {en},
	booktitle = {15th {European} {Conference} on {Game}-{Based} {Learning}},
	publisher = {Academic Conferences LTD},
	author = {López, María Ximena and Strada, Francesco and Bottino, Andrea and Fabricatore, Carlo},
	month = dec,
	year = {2021},
	pages = {482--489},
}

@article{3783339081,
	title = {A {Novel} {Method} for the {In}-{Depth} {Multimodal} {Analysis} of {Student} {Learning} {Trajectories} in {Intelligent} {Tutoring} {Systems}},
	volume = {5},
	issn = {1929-7750},
	url = {https://learning-analytics.info/index.php/JLA/article/view/5423},
	doi = {10.18608/jla.2018.51.4},
	language = {en},
	number = {1},
	urldate = {2023-08-07},
	journal = {Journal of Learning Analytics},
	author = {Liu, Ran and Stamper, John C and Davenport, Jodi},
	month = apr,
	year = {2018},
	pages = {41--54},
}

@inproceedings{853680639,
	address = {Montreal, CA},
	title = {Sensor-based {Data} {Fusion} for {Multimodal} {Affect} {Detection} in {Game}-based {Learning} {Environments}},
	volume = {2592},
	language = {en},
	booktitle = {Proceedings of the {EDM} and {Games} {Workshop} at the 12th {International} {Conference} on {Educational} {Data} {Mining}},
	publisher = {International Educational Data Mining Society},
	author = {Henderson, Nathan L and Rowe, Jonathan P and Mott, Bradford W and Lester, James C},
	month = jul,
	year = {2019},
	pages = {1--7},
}

@article{483140962,
	title = {Investigating multimodal affect sensing in an {Affective} {Tutoring} {System} using unobtrusive sensors.},
	volume = {29},
	journal = {Psychology of Programming Interest Group},
	author = {{Fwa, Hua Leong} and {Lindsay Marshall}},
	month = oct,
	year = {2018},
	pages = {78--85},
}

@article{1426267857,
	title = {Affect, {Support}, and {Personal} {Factors}: {Multimodal} {Causal} {Models} of {One}-on-one {Coaching}},
	volume = {13},
	language = {en},
	number = {3},
	journal = {Journal of Educational Data Mining},
	author = {Chen, Lujie Karen},
	year = {2021},
	pages = {36--68},
}

@article{2936220551,
	title = {Multi-source and multimodal data fusion for predicting academic performance in blended learning university courses},
	volume = {89},
	issn = {00457906},
	url = {https://linkinghub.elsevier.com/retrieve/pii/S0045790620307606},
	doi = {10.1016/j.compeleceng.2020.106908},
	language = {en},
	urldate = {2023-08-07},
	journal = {Computers \& Electrical Engineering},
	author = {Chango, Wilson and Cerezo, Rebeca and Romero, Cristóbal},
	month = jan,
	year = {2021},
	pages = {106908},
}

@article{3051560548,
	title = {Temporal analysis of multimodal data to predict collaborative learning outcomes},
	volume = {51},
	issn = {0007-1013, 1467-8535},
	url = {https://onlinelibrary.wiley.com/doi/10.1111/bjet.12982},
	doi = {10.1111/bjet.12982},
	language = {en},
	number = {5},
	urldate = {2023-08-07},
	journal = {British Journal of Educational Technology},
	author = {Olsen, Jennifer K. and Sharma, Kshitij and Rummel, Nikol and Aleven, Vincent},
	month = sep,
	year = {2020},
	pages = {1527--1547},
}

@article{3093310941,
	title = {Embodied conversational agents for multimodal automated social skills training in people with autism spectrum disorders},
	volume = {12},
	issn = {1932-6203},
	url = {https://dx.plos.org/10.1371/journal.pone.0182151},
	doi = {10.1371/journal.pone.0182151},
	language = {en},
	number = {8},
	urldate = {2023-08-07},
	journal = {PLOS ONE},
	author = {Tanaka, Hiroki and Negoro, Hideki and Iwasaka, Hidemi and Nakamura, Satoshi},
	editor = {Sakakibara, Manabu},
	month = aug,
	year = {2017},
	pages = {e0182151},
}

@article{3095923626,
	title = {A {Multimodal} {Analysis} of {Making}},
	volume = {28},
	issn = {1560-4292, 1560-4306},
	url = {http://link.springer.com/10.1007/s40593-017-0160-1},
	doi = {10.1007/s40593-017-0160-1},
	language = {en},
	number = {3},
	urldate = {2023-08-07},
	journal = {International Journal of Artificial Intelligence in Education},
	author = {Worsley, Marcelo and Blikstein, Paulo},
	month = sep,
	year = {2018},
	pages = {385--419},
}

@article{3135645357,
	title = {Multimodal teaching analytics: {Automated} extraction of orchestration graphs from wearable sensor data},
	volume = {34},
	issn = {02664909},
	shorttitle = {Multimodal teaching analytics},
	url = {https://onlinelibrary.wiley.com/doi/10.1111/jcal.12232},
	doi = {10.1111/jcal.12232},
	language = {en},
	number = {2},
	urldate = {2023-08-07},
	journal = {Journal of Computer Assisted Learning},
	author = {Prieto, L.P. and Sharma, K. and Kidzinski, Ł. and Rodríguez-Triana, M.J. and Dillenbourg, P.},
	month = apr,
	year = {2018},
	pages = {193--203},
}

@article{3146393211,
	title = {Mobile {Mixed} {Reality} for {Experiential} {Learning} and {Simulation} in {Medical} and {Health} {Sciences} {Education}},
	volume = {9},
	issn = {2078-2489},
	url = {http://www.mdpi.com/2078-2489/9/2/31},
	doi = {10.3390/info9020031},
	language = {en},
	number = {2},
	urldate = {2023-08-07},
	journal = {Information},
	author = {Birt, James and Stromberga, Zane and Cowling, Michael and Moro, Christian},
	month = jan,
	year = {2018},
	pages = {31},
}

@inproceedings{3309250332,
	address = {Sydney New South Wales Australia},
	title = {({Dis})engagement matters: identifying efficacious learning practices with multimodal learning analytics},
	isbn = {978-1-4503-6400-3},
	shorttitle = {({Dis})engagement matters},
	url = {https://dl.acm.org/doi/10.1145/3170358.3170420},
	doi = {10.1145/3170358.3170420},
	language = {en},
	urldate = {2023-08-07},
	booktitle = {Proceedings of the 8th {International} {Conference} on {Learning} {Analytics} and {Knowledge}},
	publisher = {ACM},
	author = {Worsley, Marcelo},
	month = mar,
	year = {2018},
	pages = {365--369},
}

@inproceedings{3339002981,
	address = {Timisoara, Romania},
	title = {Estimation of {Success} in {Collaborative} {Learning} {Based} on {Multimodal} {Learning} {Analytics} {Features}},
	isbn = {978-1-5386-3870-5},
	url = {http://ieeexplore.ieee.org/document/8001779/},
	doi = {10.1109/ICALT.2017.122},
	language = {en},
	urldate = {2023-08-07},
	booktitle = {2017 {IEEE} 17th {International} {Conference} on {Advanced} {Learning} {Technologies} ({ICALT})},
	publisher = {IEEE},
	author = {Spikol, Daniel and Ruffaldi, Emanuele and Landolfi, Lorenzo and Cukurova, Mutlu},
	month = jul,
	year = {2017},
	pages = {269--273},
}

@article{3398902089,
	title = {What multimodal data can tell us about the students’ regulation of their learning process?},
	volume = {72},
	issn = {09594752},
	url = {https://linkinghub.elsevier.com/retrieve/pii/S095947521830416X},
	doi = {10.1016/j.learninstruc.2019.04.004},
	language = {en},
	urldate = {2023-08-07},
	journal = {Learning and Instruction},
	author = {Järvelä, Sanna and Malmberg, Jonna and Haataja, Eetu and Sobocinski, Marta and Kirschner, Paul A.},
	month = apr,
	year = {2021},
	pages = {101203},
}

@article{3408664396,
	title = {Multimodal {Student} {Engagement} {Recognition} in {Prosocial} {Games}},
	volume = {10},
	issn = {2475-1502, 2475-1510},
	url = {https://ieeexplore.ieee.org/document/8015151/},
	doi = {10.1109/TCIAIG.2017.2743341},
	language = {en},
	number = {3},
	urldate = {2023-08-07},
	journal = {IEEE Transactions on Games},
	author = {Psaltis, Athanasios and Apostolakis, Konstantinos C. and Dimitropoulos, Kosmas and Daras, Petros},
	month = sep,
	year = {2018},
	pages = {292--303},
}

@inproceedings{3448122334,
	address = {Glasgow Scotland Uk},
	title = {Investigating the {Impact} of a {Real}-time, {Multimodal} {Student} {Engagement} {Analytics} {Technology} in {Authentic} {Classrooms}},
	isbn = {978-1-4503-5970-2},
	url = {https://dl.acm.org/doi/10.1145/3290605.3300534},
	doi = {10.1145/3290605.3300534},
	language = {en},
	urldate = {2023-08-07},
	booktitle = {Proceedings of the 2019 {CHI} {Conference} on {Human} {Factors} in {Computing} {Systems}},
	publisher = {ACM},
	author = {Aslan, Sinem and Alyuz, Nese and Tanriover, Cagri and Mete, Sinem E. and Okur, Eda and D'Mello, Sidney K. and Arslan Esme, Asli},
	month = may,
	year = {2019},
	pages = {1--12},
}

@article{3625722965,
	title = {Table {Tennis} {Tutor}: {Forehand} {Strokes} {Classification} {Based} on {Multimodal} {Data} and {Neural} {Networks}},
	volume = {21},
	issn = {1424-8220},
	shorttitle = {Table {Tennis} {Tutor}},
	url = {https://www.mdpi.com/1424-8220/21/9/3121},
	doi = {10.3390/s21093121},
	language = {en},
	number = {9},
	urldate = {2023-08-07},
	journal = {Sensors},
	author = {Mat Sanusi, Khaleel Asyraaf and Mitri, Daniele Di and Limbu, Bibeg and Klemke, Roland},
	month = apr,
	year = {2021},
	pages = {3121},
}

@article{3637456466,
	title = {Impact of inquiry interventions on students in e-learning and classroom environments using affective computing framework},
	volume = {30},
	issn = {0924-1868, 1573-1391},
	url = {http://link.springer.com/10.1007/s11257-019-09254-3},
	doi = {10.1007/s11257-019-09254-3},
	language = {en},
	number = {5},
	urldate = {2023-08-07},
	journal = {User Modeling and User-Adapted Interaction},
	author = {Ashwin, T. S. and Guddeti, Ram Mohana Reddy},
	month = nov,
	year = {2020},
	pages = {759--801},
}

@inproceedings{3660066725,
	address = {Athens Greece},
	title = {Children’s {Play} and {Problem} {Solving} in {Motion}-{Based} {Educational} {Games}: {Synergies} between {Human} {Annotations} and {Multi}-{Modal} {Data}},
	isbn = {978-1-4503-8452-0},
	shorttitle = {Children’s {Play} and {Problem} {Solving} in {Motion}-{Based} {Educational} {Games}},
	url = {https://dl.acm.org/doi/10.1145/3459990.3460702},
	doi = {10.1145/3459990.3460702},
	language = {en},
	urldate = {2023-08-07},
	booktitle = {Interaction {Design} and {Children}},
	publisher = {ACM},
	author = {Lee-Cultura, Serena and Sharma, Kshitij and Cosentino, Giulia and Papavlasopoulou, Sofia and Giannakos, Michail},
	month = jun,
	year = {2021},
	pages = {408--420},
}

@inproceedings{3754172825,
	address = {Online USA},
	title = {Detecting {Impasse} {During} {Collaborative} {Problem} {Solving} with {Multimodal} {Learning} {Analytics}},
	isbn = {978-1-4503-9573-1},
	url = {https://dl.acm.org/doi/10.1145/3506860.3506865},
	doi = {10.1145/3506860.3506865},
	language = {en},
	urldate = {2023-08-07},
	booktitle = {{LAK22}: 12th {International} {Learning} {Analytics} and {Knowledge} {Conference}},
	publisher = {ACM},
	author = {Ma, Yingbo and Celepkolu, Mehmet and Boyer, Kristy Elizabeth},
	month = mar,
	year = {2022},
	pages = {45--55},
}

@article{3796180663,
	title = {Learning linkages: {Integrating} data streams of multiple modalities and timescales},
	volume = {35},
	issn = {0266-4909, 1365-2729},
	shorttitle = {Learning linkages},
	url = {https://onlinelibrary.wiley.com/doi/10.1111/jcal.12315},
	doi = {10.1111/jcal.12315},
	language = {en},
	number = {1},
	urldate = {2023-08-07},
	journal = {Journal of Computer Assisted Learning},
	author = {Liu, Ran and Stamper, John and Davenport, Jodi and Crossley, Scott and McNamara, Danielle and Nzinga, Kalonji and Sherin, Bruce},
	month = feb,
	year = {2019},
	pages = {99--109},
}

@article{3796643912,
	title = {An evaluation of an adaptive learning system based on multimodal affect recognition for learners with intellectual disabilities},
	volume = {51},
	issn = {0007-1013, 1467-8535},
	url = {https://onlinelibrary.wiley.com/doi/10.1111/bjet.13010},
	doi = {10.1111/bjet.13010},
	language = {en},
	number = {5},
	urldate = {2023-08-07},
	journal = {British Journal of Educational Technology},
	author = {Standen, Penelope J. and Brown, David J. and Taheri, Mohammad and Galvez Trigo, Maria J. and Boulton, Helen and Burton, Andrew and Hallewell, Madeline J. and Lathe, James G. and Shopland, Nicholas and Blanco Gonzalez, Maria A. and Kwiatkowska, Gosia M. and Milli, Elena and Cobello, Stefano and Mazzucato, Annaleda and Traversi, Marco and Hortal, Enrique},
	month = sep,
	year = {2020},
	pages = {1748--1765},
}

@article{3856280479,
	title = {Children’s play and problem-solving in motion-based learning technologies using a multi-modal mixed methods approach},
	volume = {31},
	issn = {22128689},
	url = {https://linkinghub.elsevier.com/retrieve/pii/S2212868921000647},
	doi = {10.1016/j.ijcci.2021.100355},
	language = {en},
	urldate = {2023-08-07},
	journal = {International Journal of Child-Computer Interaction},
	author = {Lee-Cultura, Serena and Sharma, Kshitij and Giannakos, Michail},
	month = mar,
	year = {2022},
	pages = {100355},
}

@article{4019205162,
	title = {Introducing {Low}-{Cost} {Sensors} into the {Classroom} {Settings}: {Improving} the {Assessment} in {Agile} {Practices} with {Multimodal} {Learning} {Analytics}},
	volume = {19},
	issn = {1424-8220},
	shorttitle = {Introducing {Low}-{Cost} {Sensors} into the {Classroom} {Settings}},
	url = {https://www.mdpi.com/1424-8220/19/15/3291},
	doi = {10.3390/s19153291},
	language = {en},
	number = {15},
	urldate = {2023-08-07},
	journal = {Sensors},
	author = {Cornide-Reyes, Hector and Noël, René and Riquelme, Fabián and Gajardo, Matías and Cechinel, Cristian and Mac Lean, Roberto and Becerra, Carlos and Villarroel, Rodolfo and Munoz, Roberto},
	month = jul,
	year = {2019},
	pages = {3291},
}

@article{4035649049,
	title = {Storytelling {With} {Learner} {Data}: {Guiding} {Student} {Reflection} on {Multimodal} {Team} {Data}},
	volume = {14},
	issn = {1939-1382, 2372-0050},
	shorttitle = {Storytelling {With} {Learner} {Data}},
	url = {https://ieeexplore.ieee.org/document/9632388/},
	doi = {10.1109/TLT.2021.3131842},
	language = {en},
	number = {5},
	urldate = {2023-08-07},
	journal = {IEEE Transactions on Learning Technologies},
	author = {Fernandez-Nieto, Gloria Milena and Echeverria, Vanessa and Shum, Simon Buckingham and Mangaroska, Katerina and Kitto, Kirsty and Palominos, Evelyn and Axisa, Carmen and Martinez-Maldonado, Roberto},
	month = oct,
	year = {2021},
	pages = {695--708},
}

@article{4278392816,
	title = {Multimodal data as a means to understand the learning experience},
	volume = {48},
	issn = {02684012},
	url = {https://linkinghub.elsevier.com/retrieve/pii/S0268401218312751},
	doi = {10.1016/j.ijinfomgt.2019.02.003},
	language = {en},
	urldate = {2023-08-07},
	journal = {International Journal of Information Management},
	author = {Giannakos, Michail N. and Sharma, Kshitij and Pappas, Ilias O. and Kostakos, Vassilis and Velloso, Eduardo},
	month = oct,
	year = {2019},
	pages = {108--119},
}

@article{4277812050,
	title = {Improving prediction of students’ performance in intelligent tutoring systems using attribute selection and ensembles of different multimodal data sources},
	volume = {33},
	issn = {1042-1726, 1867-1233},
	url = {https://link.springer.com/10.1007/s12528-021-09298-8},
	doi = {10.1007/s12528-021-09298-8},
	language = {en},
	number = {3},
	urldate = {2023-08-07},
	journal = {Journal of Computing in Higher Education},
	author = {Chango, Wilson and Cerezo, Rebeca and Sanchez-Santillan, Miguel and Azevedo, Roger and Romero, Cristóbal},
	month = dec,
	year = {2021},
	pages = {614--634},
}

@article{3809293172,
	title = {Blending learning analytics and embodied design to model students’ comprehension of measurement using their actions, speech, and gestures},
	volume = {32},
	issn = {22128689},
	url = {https://linkinghub.elsevier.com/retrieve/pii/S2212868921000866},
	doi = {10.1016/j.ijcci.2021.100391},
	language = {en},
	urldate = {2023-08-07},
	journal = {International Journal of Child-Computer Interaction},
	author = {Closser, Avery H. and Erickson, John A. and Smith, Hannah and Varatharaj, Ashvini and Botelho, Anthony F.},
	month = jun,
	year = {2022},
	pages = {100391},
}

@incollection{2836996318,
	address = {Cham},
	title = {Predicting {Learners}’ {Emotions} in {Mobile} {MOOC} {Learning} via a {Multimodal} {Intelligent} {Tutor}},
	volume = {10858},
	isbn = {978-3-319-91463-3 978-3-319-91464-0},
	url = {http://link.springer.com/10.1007/978-3-319-91464-0_15},
	language = {en},
	urldate = {2023-08-07},
	booktitle = {Intelligent {Tutoring} {Systems}},
	publisher = {Springer International Publishing},
	author = {Pham, Phuong and Wang, Jingtao},
	editor = {Nkambou, Roger and Azevedo, Roger and Vassileva, Julita},
	year = {2018},
	pages = {150--159},
}

@article{2634033325,
	title = {Controlled evaluation of a multimodal system to improve oral presentation skills in a real learning setting},
	volume = {51},
	issn = {0007-1013, 1467-8535},
	url = {https://onlinelibrary.wiley.com/doi/10.1111/bjet.12987},
	doi = {10.1111/bjet.12987},
	language = {en},
	number = {5},
	urldate = {2023-08-07},
	journal = {British Journal of Educational Technology},
	author = {Ochoa, Xavier and Dominguez, Federico},
	month = sep,
	year = {2020},
	pages = {1615--1630},
}

@article{2609260641,
	title = {Visualizing {Collaboration} in {Teamwork}: {A} {Multimodal} {Learning} {Analytics} {Platform} for {Non}-{Verbal} {Communication}},
	volume = {12},
	issn = {2076-3417},
	shorttitle = {Visualizing {Collaboration} in {Teamwork}},
	url = {https://www.mdpi.com/2076-3417/12/15/7499},
	doi = {10.3390/app12157499},
	language = {en},
	number = {15},
	urldate = {2023-08-07},
	journal = {Applied Sciences},
	author = {Noël, René and Miranda, Diego and Cechinel, Cristian and Riquelme, Fabián and Primo, Tiago Thompsen and Munoz, Roberto},
	month = jul,
	year = {2022},
	pages = {7499},
}

@inproceedings{2497456347,
	address = {Sydney New South Wales Australia},
	title = {The {RAP} system: automatic feedback of oral presentation skills using multimodal analysis and low-cost sensors},
	isbn = {978-1-4503-6400-3},
	shorttitle = {The {RAP} system},
	url = {https://dl.acm.org/doi/10.1145/3170358.3170406},
	doi = {10.1145/3170358.3170406},
	language = {en},
	urldate = {2023-08-07},
	booktitle = {Proceedings of the 8th {International} {Conference} on {Learning} {Analytics} and {Knowledge}},
	publisher = {ACM},
	author = {Ochoa, Xavier and Domínguez, Federico and Guamán, Bruno and Maya, Ricardo and Falcones, Gabriel and Castells, Jaime},
	month = mar,
	year = {2018},
	pages = {360--364},
}

@inproceedings{2456887548,
	address = {Glasgow UK},
	title = {An unobtrusive and multimodal approach for behavioral engagement detection of students},
	isbn = {978-1-4503-5557-5},
	url = {https://dl.acm.org/doi/10.1145/3139513.3139521},
	doi = {10.1145/3139513.3139521},
	language = {en},
	urldate = {2023-08-07},
	booktitle = {Proceedings of the 1st {ACM} {SIGCHI} {International} {Workshop} on {Multimodal} {Interaction} for {Education}},
	publisher = {ACM},
	author = {Alyuz, Nese and Okur, Eda and Genc, Utku and Aslan, Sinem and Tanriover, Cagri and Esme, Asli Arslan},
	month = nov,
	year = {2017},
	pages = {26--32},
}

@article{2345021698,
	title = {Exploring {Collaborative} {Writing} of {User} {Stories} {With} {Multimodal} {Learning} {Analytics}: {A} {Case} {Study} on a {Software} {Engineering} {Course}},
	volume = {6},
	issn = {2169-3536},
	shorttitle = {Exploring {Collaborative} {Writing} of {User} {Stories} {With} {Multimodal} {Learning} {Analytics}},
	url = {https://ieeexplore.ieee.org/document/8496762/},
	doi = {10.1109/ACCESS.2018.2876801},
	language = {en},
	urldate = {2023-08-07},
	journal = {IEEE Access},
	author = {Noel, Rene and Riquelme, Fabian and Lean, Roberto Mac and Merino, Erick and Cechinel, Cristian and Barcelos, Thiago S. and Villarroel, Rodolfo and Munoz, Roberto},
	year = {2018},
	pages = {67783--67798},
}

@article{2273914836,
	title = {Many are the ways to learn identifying multi-modal behavioral profiles of collaborative learning in constructivist activities},
	volume = {16},
	issn = {1556-1607, 1556-1615},
	url = {https://link.springer.com/10.1007/s11412-021-09358-2},
	doi = {10.1007/s11412-021-09358-2},
	language = {en},
	number = {4},
	urldate = {2023-08-07},
	journal = {International Journal of Computer-Supported Collaborative Learning},
	author = {Nasir, Jauwairia and Kothiyal, Aditi and Bruno, Barbara and Dillenbourg, Pierre},
	month = dec,
	year = {2021},
	pages = {485--523},
}

@article{2155422499,
	title = {A multimodal analysis of pair work engagement episodes: {Implications} for {EMI} lecturer training},
	volume = {58},
	issn = {14751585},
	shorttitle = {A multimodal analysis of pair work engagement episodes},
	url = {https://linkinghub.elsevier.com/retrieve/pii/S1475158522000443},
	doi = {10.1016/j.jeap.2022.101124},
	language = {en},
	urldate = {2023-08-07},
	journal = {Journal of English for Academic Purposes},
	author = {Morell, Teresa and Beltrán-Palanques, Vicent and Norte, Natalia},
	month = jul,
	year = {2022},
	pages = {101124},
}

@article{2055153191,
	title = {Round or rectangular tables for collaborative problem solving? {A} multimodal learning analytics study},
	volume = {51},
	issn = {0007-1013, 1467-8535},
	shorttitle = {Round or rectangular tables for collaborative problem solving?},
	url = {https://onlinelibrary.wiley.com/doi/10.1111/bjet.12988},
	doi = {10.1111/bjet.12988},
	language = {en},
	number = {5},
	urldate = {2023-08-07},
	journal = {British Journal of Educational Technology},
	author = {Vujovic, Milica and Hernández‐Leo, Davinia and Tassani, Simone and Spikol, Daniel},
	month = sep,
	year = {2020},
	pages = {1597--1614},
}

@inproceedings{2000036002,
	address = {Frankfurt Germany},
	title = {Predicting learners' effortful behaviour in adaptive assessment using multimodal data},
	isbn = {978-1-4503-7712-6},
	url = {https://dl.acm.org/doi/10.1145/3375462.3375498},
	doi = {10.1145/3375462.3375498},
	language = {en},
	urldate = {2023-08-07},
	booktitle = {Proceedings of the {Tenth} {International} {Conference} on {Learning} {Analytics} \& {Knowledge}},
	publisher = {ACM},
	author = {Sharma, Kshitij and Papamitsiou, Zacharoula and Olsen, Jennifer K. and Giannakos, Michail},
	month = mar,
	year = {2020},
	pages = {480--489},
}

@inproceedings{1886134458,
	address = {San Jose, CA, USA},
	title = {Personalizing {Computer} {Science} {Education} by {Leveraging} {Multimodal} {Learning} {Analytics}},
	isbn = {978-1-5386-1174-6},
	url = {https://ieeexplore.ieee.org/document/8658596/},
	doi = {10.1109/FIE.2018.8658596},
	language = {en},
	urldate = {2023-08-07},
	booktitle = {2018 {IEEE} {Frontiers} in {Education} {Conference} ({FIE})},
	publisher = {IEEE},
	author = {Azcona, David and Hsiao, I-Han and Smeaton, Alan F.},
	month = oct,
	year = {2018},
	pages = {1--9},
}

@inproceedings{1877483551,
	address = {Osaka, Japan},
	title = {Motion-{Based} {Educational} {Games}: {Using} {Multi}-{Modal} {Data} to {Predict} {Player}’s {Performance}},
	isbn = {978-1-72814-533-4},
	shorttitle = {Motion-{Based} {Educational} {Games}},
	url = {https://ieeexplore.ieee.org/document/9231892/},
	doi = {10.1109/CoG47356.2020.9231892},
	language = {en},
	urldate = {2023-08-07},
	booktitle = {2020 {IEEE} {Conference} on {Games} ({CoG})},
	publisher = {IEEE},
	author = {Lee-Cultura, Serena and Sharma, Kshitij and Papavlasopoulou, Sofia and Giannakos, Michail},
	month = aug,
	year = {2020},
	pages = {17--24},
}

@incollection{1847468084,
	address = {Cham},
	title = {Computationally {Augmented} {Ethnography}: {Emotion} {Tracking} and {Learning} in {Museum} {Games}},
	volume = {1112},
	isbn = {978-3-030-33231-0 978-3-030-33232-7},
	shorttitle = {Computationally {Augmented} {Ethnography}},
	url = {http://link.springer.com/10.1007/978-3-030-33232-7_12},
	language = {en},
	urldate = {2023-08-07},
	booktitle = {Advances in {Quantitative} {Ethnography}},
	publisher = {Springer International Publishing},
	author = {Martin, Kit and Wang, Emily Q. and Bain, Connor and Worsley, Marcelo},
	editor = {Eagan, Brendan and Misfeldt, Morten and Siebert-Evenstone, Amanda},
	year = {2019},
	pages = {141--153},
}

@inproceedings{1770989706,
	address = {Frankfurt Germany},
	title = {Focused or stuck together: multimodal patterns reveal triads' performance in collaborative problem solving},
	isbn = {978-1-4503-7712-6},
	shorttitle = {Focused or stuck together},
	url = {https://dl.acm.org/doi/10.1145/3375462.3375467},
	doi = {10.1145/3375462.3375467},
	language = {en},
	urldate = {2023-08-07},
	booktitle = {Proceedings of the {Tenth} {International} {Conference} on {Learning} {Analytics} \& {Knowledge}},
	publisher = {ACM},
	author = {Vrzakova, Hana and Amon, Mary Jean and Stewart, Angela and Duran, Nicholas D. and D'Mello, Sidney K.},
	month = mar,
	year = {2020},
	pages = {295--304},
}

@article{1763513559,
	title = {Keep {Me} in the {Loop}: {Real}-{Time} {Feedback} with {Multimodal} {Data}},
	volume = {32},
	issn = {1560-4292, 1560-4306},
	shorttitle = {Keep {Me} in the {Loop}},
	url = {https://link.springer.com/10.1007/s40593-021-00281-z},
	doi = {10.1007/s40593-021-00281-z},
	language = {en},
	number = {4},
	urldate = {2023-08-07},
	journal = {International Journal of Artificial Intelligence in Education},
	author = {Di Mitri, Daniele and Schneider, Jan and Drachsler, Hendrik},
	month = dec,
	year = {2022},
	pages = {1093--1118},
}

@article{1637690235,
	title = {Supervised machine learning in multimodal learning analytics for estimating success in project-based learning},
	volume = {34},
	issn = {02664909},
	url = {https://onlinelibrary.wiley.com/doi/10.1111/jcal.12263},
	doi = {10.1111/jcal.12263},
	language = {en},
	number = {4},
	urldate = {2023-08-07},
	journal = {Journal of Computer Assisted Learning},
	author = {Spikol, Daniel and Ruffaldi, Emanuele and Dabisias, Giacomo and Cukurova, Mutlu},
	month = aug,
	year = {2018},
	pages = {366--377},
}

@inproceedings{1609706685,
	address = {Vancouver British Columbia Canada},
	title = {Learning pulse: a machine learning approach for predicting performance in self-regulated learning using multimodal data},
	isbn = {978-1-4503-4870-6},
	shorttitle = {Learning pulse},
	url = {https://dl.acm.org/doi/10.1145/3027385.3027447},
	doi = {10.1145/3027385.3027447},
	language = {en},
	urldate = {2023-08-07},
	booktitle = {Proceedings of the {Seventh} {International} {Learning} {Analytics} \& {Knowledge} {Conference}},
	publisher = {ACM},
	author = {Di Mitri, Daniele and Scheffel, Maren and Drachsler, Hendrik and Börner, Dirk and Ternier, Stefaan and Specht, Marcus},
	month = mar,
	year = {2017},
	pages = {188--197},
}

@article{1598166515,
	title = {Multimodal learning analytics for game‐based learning},
	volume = {51},
	issn = {0007-1013, 1467-8535},
	url = {https://onlinelibrary.wiley.com/doi/10.1111/bjet.12992},
	doi = {10.1111/bjet.12992},
	language = {en},
	number = {5},
	urldate = {2023-08-07},
	journal = {British Journal of Educational Technology},
	author = {Emerson, Andrew and Cloude, Elizabeth B. and Azevedo, Roger and Lester, James},
	month = sep,
	year = {2020},
	pages = {1505--1526},
}

@inproceedings{1581261659,
	address = {Virtual Event Netherlands},
	title = {Early {Prediction} of {Visitor} {Engagement} in {Science} {Museums} with {Multimodal} {Learning} {Analytics}},
	isbn = {978-1-4503-7581-8},
	url = {https://dl.acm.org/doi/10.1145/3382507.3418890},
	doi = {10.1145/3382507.3418890},
	language = {en},
	urldate = {2023-08-07},
	booktitle = {Proceedings of the 2020 {International} {Conference} on {Multimodal} {Interaction}},
	publisher = {ACM},
	author = {Emerson, Andrew and Henderson, Nathan and Rowe, Jonathan and Min, Wookhee and Lee, Seung and Minogue, James and Lester, James},
	month = oct,
	year = {2020},
	pages = {107--116},
}

@article{1576545447,
	title = {Artificial intelligence and multimodal data in the service of human decision‐making: {A} case study in debate tutoring},
	volume = {50},
	issn = {0007-1013, 1467-8535},
	shorttitle = {Artificial intelligence and multimodal data in the service of human decision‐making},
	url = {https://onlinelibrary.wiley.com/doi/10.1111/bjet.12829},
	doi = {10.1111/bjet.12829},
	language = {en},
	number = {6},
	urldate = {2023-08-07},
	journal = {British Journal of Educational Technology},
	author = {Cukurova, Mutlu and Kent, Carmel and Luckin, Rosemary},
	month = nov,
	year = {2019},
	pages = {3032--3046},
}

@article{1469065963,
	title = {Examining socially shared regulation and shared physiological arousal events with multimodal learning analytics},
	volume = {54},
	issn = {0007-1013, 1467-8535},
	url = {https://onlinelibrary.wiley.com/doi/10.1111/bjet.13280},
	doi = {10.1111/bjet.13280},
	language = {en},
	number = {1},
	urldate = {2023-08-07},
	journal = {British Journal of Educational Technology},
	author = {Nguyen, Andy and Järvelä, Sanna and Rosé, Carolyn and Järvenoja, Hanna and Malmberg, Jonna},
	month = jan,
	year = {2023},
	pages = {293--312},
}

@incollection{1374035721,
	address = {Cham},
	title = {{AttentiveLearner2}: {A} {Multimodal} {Approach} for {Improving} {MOOC} {Learning} on {Mobile} {Devices}},
	volume = {10331},
	isbn = {978-3-319-61424-3 978-3-319-61425-0},
	shorttitle = {{AttentiveLearner2}},
	url = {http://link.springer.com/10.1007/978-3-319-61425-0_64},
	language = {en},
	urldate = {2023-08-07},
	booktitle = {Artificial {Intelligence} in {Education}},
	publisher = {Springer International Publishing},
	author = {Pham, Phuong and Wang, Jingtao},
	editor = {André, Elisabeth and Baker, Ryan and Hu, Xiangen and Rodrigo, Ma. Mercedes T. and Du Boulay, Benedict},
	year = {2017},
	pages = {561--564},
}

@incollection{1345598079,
	address = {Cham},
	title = {Intermodality in {Multimodal} {Learning} {Analytics} for {Cognitive} {Theory} {Development}: {A} {Case} from {Embodied} {Design} for {Mathematics} {Learning}},
	isbn = {978-3-031-08075-3 978-3-031-08076-0},
	shorttitle = {Intermodality in {Multimodal} {Learning} {Analytics} for {Cognitive} {Theory} {Development}},
	url = {https://link.springer.com/10.1007/978-3-031-08076-0_6},
	language = {en},
	urldate = {2023-08-07},
	booktitle = {The {Multimodal} {Learning} {Analytics} {Handbook}},
	publisher = {Springer International Publishing},
	author = {Tancredi, Sofia and Abdu, Rotem and Balasubramaniam, Ramesh and Abrahamson, Dor},
	editor = {Giannakos, Michail and Spikol, Daniel and Di Mitri, Daniele and Sharma, Kshitij and Ochoa, Xavier and Hammad, Rawad},
	year = {2022},
	pages = {133--158},
}

@incollection{1326191931,
	address = {Cham},
	title = {Multimodal {Learning} {Analytics} in a {Laboratory} {Classroom}},
	volume = {158},
	isbn = {978-3-030-13742-7 978-3-030-13743-4},
	url = {http://link.springer.com/10.1007/978-3-030-13743-4_8},
	language = {en},
	urldate = {2023-08-07},
	booktitle = {Machine {Learning} {Paradigms}},
	publisher = {Springer International Publishing},
	author = {Chan, Man Ching Esther and Ochoa, Xavier and Clarke, David},
	editor = {Virvou, Maria and Alepis, Efthimios and Tsihrintzis, George A. and Jain, Lakhmi C.},
	year = {2020},
	pages = {131--156},
}

@article{1315379489,
	title = {Multimodal {Engagement} {Analysis} {From} {Facial} {Videos} in the {Classroom}},
	volume = {14},
	issn = {1949-3045, 2371-9850},
	url = {https://ieeexplore.ieee.org/document/9613750/},
	doi = {10.1109/TAFFC.2021.3127692},
	language = {en},
	number = {2},
	urldate = {2023-08-07},
	journal = {IEEE Transactions on Affective Computing},
	author = {Sümer, Ömer and Goldberg, Patricia and D’Mello, Sidney and Gerjets, Peter and Trautwein, Ulrich and Kasneci, Enkelejda},
	month = apr,
	year = {2023},
	pages = {1012--1027},
}

@inproceedings{1296637108,
	address = {Glasgow Scotland Uk},
	title = {Towards {Collaboration} {Translucence}: {Giving} {Meaning} to {Multimodal} {Group} {Data}},
	isbn = {978-1-4503-5970-2},
	shorttitle = {Towards {Collaboration} {Translucence}},
	url = {https://dl.acm.org/doi/10.1145/3290605.3300269},
	doi = {10.1145/3290605.3300269},
	language = {en},
	urldate = {2023-08-07},
	booktitle = {Proceedings of the 2019 {CHI} {Conference} on {Human} {Factors} in {Computing} {Systems}},
	publisher = {ACM},
	author = {Echeverria, Vanessa and Martinez-Maldonado, Roberto and Buckingham Shum, Simon},
	month = may,
	year = {2019},
	pages = {1--16},
}

@incollection{1019093033,
	address = {Cham},
	title = {{PRIME}: {Block}-{Wise} {Missingness} {Handling} for {Multi}-modalities in {Intelligent} {Tutoring} {Systems}},
	volume = {11962},
	isbn = {978-3-030-37733-5 978-3-030-37734-2},
	shorttitle = {{PRIME}},
	url = {http://link.springer.com/10.1007/978-3-030-37734-2_6},
	language = {en},
	urldate = {2023-08-07},
	booktitle = {{MultiMedia} {Modeling}},
	publisher = {Springer International Publishing},
	author = {Yang, Xi and Kim, Yeo-Jin and Taub, Michelle and Azevedo, Roger and Chi, Min},
	editor = {Ro, Yong Man and Cheng, Wen-Huang and Kim, Junmo and Chu, Wei-Ta and Cui, Peng and Choi, Jung-Woo and Hu, Min-Chun and De Neve, Wesley},
	year = {2020},
	pages = {63--75},
}

@inproceedings{957160695,
	address = {Glasgow UK},
	title = {Virtual debate coach design: assessing multimodal argumentation performance},
	isbn = {978-1-4503-5543-8},
	shorttitle = {Virtual debate coach design},
	url = {https://dl.acm.org/doi/10.1145/3136755.3136775},
	doi = {10.1145/3136755.3136775},
	language = {en},
	urldate = {2023-08-07},
	booktitle = {Proceedings of the 19th {ACM} {International} {Conference} on {Multimodal} {Interaction}},
	publisher = {ACM},
	author = {Petukhova, Volha and Mayer, Tobias and Malchanau, Andrei and Bunt, Harry},
	month = nov,
	year = {2017},
	pages = {41--50},
}

@inproceedings{818492192,
	address = {Vancouver British Columbia Canada},
	title = {Understanding student learning trajectories using multimodal learning analytics within an embodied-interaction learning environment},
	isbn = {978-1-4503-4870-6},
	url = {https://dl.acm.org/doi/10.1145/3027385.3027429},
	doi = {10.1145/3027385.3027429},
	language = {en},
	urldate = {2023-08-07},
	booktitle = {Proceedings of the {Seventh} {International} {Learning} {Analytics} \& {Knowledge} {Conference}},
	publisher = {ACM},
	author = {Andrade, Alejandro},
	month = mar,
	year = {2017},
	pages = {70--79},
}

@inproceedings{804659204,
	address = {Wollongong, NSW},
	title = {Towards {Smart} {Educational} {Recommendations} with {Reinforcement} {Learning} in {Classroom}},
	isbn = {978-1-5386-6522-0},
	url = {https://ieeexplore.ieee.org/document/8615217/},
	doi = {10.1109/TALE.2018.8615217},
	language = {en},
	urldate = {2023-08-07},
	booktitle = {2018 {IEEE} {International} {Conference} on {Teaching}, {Assessment}, and {Learning} for {Engineering} ({TALE})},
	publisher = {IEEE},
	author = {Liu, Su and Chen, Ye and Huang, Hui and Xiao, Liang and Hei, Xiaojun},
	month = dec,
	year = {2018},
	pages = {1079--1084},
}

@incollection{666050348,
	address = {Cham},
	title = {Multicraft: {A} {Multimodal} {Interface} for {Supporting} and {Studying} {Learning} in {Minecraft}},
	volume = {12790},
	isbn = {978-3-030-77413-4 978-3-030-77414-1},
	shorttitle = {Multicraft},
	url = {https://link.springer.com/10.1007/978-3-030-77414-1_10},
	language = {en},
	urldate = {2023-08-07},
	booktitle = {{HCI} in {Games}: {Serious} and {Immersive} {Games}},
	publisher = {Springer International Publishing},
	author = {Worsley, Marcelo and Mendoza Tudares, Kevin and Mwiti, Timothy and Zhen, Mitchell and Jiang, Marc},
	editor = {Fang, Xiaowen},
	year = {2021},
	pages = {113--131},
}

@article{566043228,
	title = {Automatic {Student} {Engagement} in {Online} {Learning} {Environment} {Based} on {Neural} {Turing} {Machine}},
	volume = {11},
	issn = {20103689},
	url = {http://www.ijiet.org/show-151-1748-1.html},
	doi = {10.18178/ijiet.2021.11.3.1497},
	language = {en},
	number = {3},
	urldate = {2023-08-07},
	journal = {International Journal of Information and Education Technology},
	author = {{Capital Normal University, Beijing, China} and Ma, Xiaoyang and Xu, Min and Dong, Yao and Sun, Zhong},
	year = {2021},
	pages = {107--111},
}

@inproceedings{433919853,
	address = {Braga Portugal},
	title = {Understanding {Fun} in {Learning} to {Code}: {A} {Multi}-{Modal} {Data} approach},
	isbn = {978-1-4503-9197-9},
	shorttitle = {Understanding {Fun} in {Learning} to {Code}},
	url = {https://dl.acm.org/doi/10.1145/3501712.3529716},
	doi = {10.1145/3501712.3529716},
	language = {en},
	urldate = {2023-08-07},
	booktitle = {Interaction {Design} and {Children}},
	publisher = {ACM},
	author = {Tisza, Gabriella and Sharma, Kshitij and Papavlasopoulou, Sofia and Markopoulos, Panos and Giannakos, Michail},
	month = jun,
	year = {2022},
	pages = {274--287},
}

@article{205660768,
	title = {Multimodal learning analytics to investigate cognitive load during online problem solving},
	volume = {51},
	issn = {0007-1013, 1467-8535},
	url = {https://onlinelibrary.wiley.com/doi/10.1111/bjet.12958},
	doi = {10.1111/bjet.12958},
	language = {en},
	number = {5},
	urldate = {2023-08-07},
	journal = {British Journal of Educational Technology},
	author = {Larmuseau, Charlotte and Cornelis, Jan and Lancieri, Luigi and Desmet, Piet and Depaepe, Fien},
	month = sep,
	year = {2020},
	pages = {1548--1562},
}

@article{147203129,
	title = {Multimodal {Learning} {Analytics} to {Inform} {Learning} {Design}: {Lessons} {Learned} from {Computing} {Education}},
	volume = {7},
	issn = {19297750},
	shorttitle = {Multimodal {Learning} {Analytics} to {Inform} {Learning} {Design}},
	url = {https://www.learning-analytics.info/index.php/JLA/article/view/6816},
	doi = {10.18608/jla.2020.73.7},
	language = {en},
	number = {3},
	urldate = {2023-08-07},
	journal = {Journal of Learning Analytics},
	author = {Mangaroska, Katerina and Sharma, Kshitij and Gašević, Dragan and Giannakos, Michalis},
	month = dec,
	year = {2020},
	pages = {79--97},
}

@article{123412197,
	title = {Utilizing {Multimodal} {Data} {Through} {fsQCA} to {Explain} {Engagement} in {Adaptive} {Learning}},
	volume = {13},
	issn = {1939-1382, 2372-0050},
	url = {https://ieeexplore.ieee.org/document/9181457/},
	doi = {10.1109/TLT.2020.3020499},
	language = {en},
	number = {4},
	urldate = {2023-08-07},
	journal = {IEEE Transactions on Learning Technologies},
	author = {Papamitsiou, Zacharoula and Pappas, Ilias O. and Sharma, Kshitij and Giannakos, Michail N.},
	month = oct,
	year = {2020},
	pages = {689--703},
}

@article{86191824,
	title = {Examining how different modes mediate adolescents’ interactions during their collaborative multimodal composing processes},
	volume = {29},
	issn = {1049-4820, 1744-5191},
	url = {https://www.tandfonline.com/doi/full/10.1080/10494820.2019.1612450},
	doi = {10.1080/10494820.2019.1612450},
	language = {en},
	number = {5},
	urldate = {2023-08-07},
	journal = {Interactive Learning Environments},
	author = {Jiang, Shiyan and Smith, Blaine E. and Shen, Ji},
	month = jul,
	year = {2021},
	pages = {807--820},
}

@incollection{32184286,
	address = {Cham},
	title = {Once {More} with {Feeling}: {Emotions} in {Multimodal} {Learning} {Analytics}},
	isbn = {978-3-031-08075-3 978-3-031-08076-0},
	shorttitle = {Once {More} with {Feeling}},
	url = {https://link.springer.com/10.1007/978-3-031-08076-0_11},
	language = {en},
	urldate = {2023-08-07},
	booktitle = {The {Multimodal} {Learning} {Analytics} {Handbook}},
	publisher = {Springer International Publishing},
	author = {Kubsch, Marcus and Caballero, Daniela and Uribe, Pablo},
	editor = {Giannakos, Michail and Spikol, Daniel and Di Mitri, Daniele and Sharma, Kshitij and Ochoa, Xavier and Hammad, Rawad},
	year = {2022},
	pages = {261--285},
}

@inbook{425012016,
    author = {Anton, Jacqueline and Cosentino, Giulia and Sharma, Kshitij and Gelsomini, Mirko and Mok, Micah and Giannakos, Michail and Abrahamson, Dor},
    title = {The Human Condition: Modal and Interactive Advantages of Teacher over AI Feedback on Children's Mathematical Performance},
    year = {2025},
    isbn = {9798400714733},
    publisher = {Association for Computing Machinery},
    address = {New York, NY, USA},
    url = {https://doi.org/10.1145/3713043.3728863},
    pages = {183–203},
    numpages = {21}
}

@article{2668965770,
  title={Exploring kid space in the wild: a preliminary study of multimodal and immersive collaborative play-based learning experiences},
  author={Aslan, Sinem and Agrawal, Ankur and Alyuz, Nese and Chierichetti, Rebecca and Durham, Lenitra M and Manuvinakurike, Ramesh and Okur, Eda and Sahay, Saurav and Sharma, Sangita and Sherry, John and others},
  journal={Educational technology research and development},
  volume={70},
  number={1},
  pages={205--230},
  year={2022},
  publisher={Springer}
}

@article{4089325423,
  title={Predicting behavior change in students with special education needs using multimodal learning analytics},
  author={Chan, Rosanna Yuen-Yan and Wong, Chun Man Victor and Yum, Yen Na},
  journal={IEEE Access},
  volume={11},
  pages={63238--63251},
  year={2023},
  publisher={IEEE}
}

@inproceedings{1196965665,
  title={How to build more generalizable models for collaboration quality? lessons learned from exploring multi-context audio-log datasets using multimodal learning analytics},
  author={Chejara, Pankaj and Prieto, Luis P and Rodriguez-Triana, Maria Jesus and Kasepalu, Reet and Ruiz-Calleja, Adolfo and Shankar, Shashi Kant},
  booktitle={LAK23: 13th International Learning Analytics and Knowledge Conference},
  pages={111--121},
  year={2023}
}

@inproceedings{1731146538,
  title={Impact of window size on the generalizability of collaboration quality estimation models developed using Multimodal Learning Analytics},
  author={Chejara, Pankaj and Prieto, Luis P and Rodriguez-Triana, Maria Jesus and Ruiz-Calleja, Adolfo and Khalil, Mohammad},
  booktitle={LAK23: 13th International Learning Analytics and Knowledge Conference},
  pages={559--565},
  year={2023}
}

@article{2764645776,
  title={MindScratch: A Visual Programming Support Tool for Classroom Learning Based on Multimodal Generative AI},
  author={Chen, Yunnong and Xiao, Shuhong and Song, Yaxuan and Li, Zejian and Sun, Lingyun and Chen, Liuqing},
  journal={International Journal of Human--Computer Interaction},
  pages={1--19},
  year={2025},
  publisher={Taylor \& Francis}
}

@article{328477558,
  title={Unpacking help-seeking process through multimodal learning analytics: A comparative study of ChatGPT vs Human expert},
  author={Chen, Angxuan and Xiang, Mengtong and Zhou, Junyi and Jia, Jiyou and Shang, Junjie and Li, Xinyu and Ga{\v{s}}evi{\'c}, Dragan and Fan, Yizhou},
  journal={Computers \& Education},
  volume={226},
  pages={105198},
  year={2025},
  publisher={Elsevier}
}

@article{1225141845,
  title={Exploring students’ multimodal representations of ideas about epistemic reading of scientific texts in generative AI tools},
  author={Cheung, Kason Ka Ching and Pun, Jack and Kenneth-Li, Wangyin and Mai, Jiayi},
  journal={Journal of Science Education and Technology},
  volume={34},
  number={2},
  pages={284--297},
  year={2025},
  publisher={Springer}
}

@article{3304069824,
  title={Class integration of ChatGPT and learning analytics for higher education},
  author={Civit, Miguel and Escalona, Mar{\'\i}a Jos{\'e} and Cuadrado, Francisco and Reyes-de-Cozar, Salvador},
  journal={Expert Systems},
  volume={41},
  number={12},
  pages={e13703},
  year={2024},
  publisher={Wiley Online Library}
}

@article{3537775194,
  title={Personalized and Timely Feedback in Online Education: Enhancing Learning with Deep Learning and Large Language Models},
  author={Cu{\'e}llar, {\'O}scar and Contero, Manuel and Hincapi{\'e}, Mauricio},
  journal={Multimodal Technologies and Interaction},
  volume={9},
  number={5},
  pages={45},
  year={2025},
  publisher={MDPI}
}

@article{2846172025,
  title={Generative AI and multimodal data for educational feedback: Insights from embodied math learning},
  author={Cosentino, Giulia and Anton, Jacqueline and Sharma, Kshitij and Gelsomini, Mirko and Giannakos, Michail and Abrahamson, Dor},
  journal={British Journal of Educational Technology},
  year={2025},
  publisher={Wiley Online Library}
}

@inproceedings{1040787959,
  title={TeamSlides: A multimodal teamwork analytics dashboard for teacher-guided reflection in a physical learning space},
  author={Echeverria, Vanessa and Yan, Lixiang and Zhao, Linxuan and Abel, Sophie and Alfredo, Riordan and Dix, Samantha and Jaggard, Hollie and Wotherspoon, Rosie and Osborne, Abra and Buckingham Shum, Simon and others},
  booktitle={Proceedings of the 14th learning analytics and knowledge conference},
  pages={112--122},
  year={2024}
}

@inproceedings{151988148,
  title={Data storytelling editor: A teacher-centred tool for customising learning analytics dashboard narratives},
  author={Fernandez-Nieto, Gloria Milena and Martinez-Maldonado, Roberto and Echeverria, Vanessa and Kitto, Kirsty and Ga{\v{s}}evi{\'c}, Dragan and Buckingham Shum, Simon},
  booktitle={Proceedings of the 14th Learning Analytics and Knowledge Conference},
  pages={678--689},
  year={2024}
}

@article{2243240858,
  title={Llm-based student plan generation for adaptive scaffolding in game-based learning environments},
  author={Goslen, Alex and Kim, Yeo Jin and Rowe, Jonathan and Lester, James},
  journal={International journal of artificial intelligence in education},
  volume={35},
  number={2},
  pages={533--558},
  year={2025},
  publisher={Springer}
}

@article{141378338,
  title={How Did the Generative Artificial Intelligence-Assisted Digital Multimodal Composing Process Facilitate the Production of Quality Digital Multimodal Compositions: Toward a Process-Genre Integrated Model},
  author={Jiang, Lianjiang and Lai, Chun},
  journal={TESOL Quarterly},
  year={2025},
  publisher={Wiley Online Library}
}

@inproceedings{2166765216,
  title={Chatting with a learning analytics dashboard: The role of generative AI literacy on learner interaction with conventional and scaffolding chatbots},
  author={Jin, Yueqiao and Yang, Kaixun and Yan, Lixiang and Echeverria, Vanessa and Zhao, Linxuan and Alfredo, Riordan and Milesi, Mikaela and Fan, Jie Xiang and Li, Xinyu and Gasevic, Dragan and others},
  booktitle={Proceedings of the 15th International Learning Analytics and Knowledge Conference},
  pages={579--590},
  year={2025}
}

@inproceedings{2280467946,
  title={Multimodal Writing Evaluation in Digital Storytelling using Video-Based Output: Comparing performance of AI and Human Raters.},
  author={Liu, Sichen and Kim, Eunyoung},
  booktitle={Proceedings of the 6th International Conference on Modern Educational Technology},
  pages={117--123},
  year={2024}
}

@article{962997360,
  title={Multimodal Communication and Peer Interaction during Equation-Solving Sessions with and without Tangible Technologies},
  author={Lehtonen, Daranee and Joutsenlahti, Jorma and Perkkil{\"a}, P{\"a}ivi},
  journal={Multimodal Technologies and Interaction},
  volume={7},
  number={1},
  pages={6},
  year={2023},
  publisher={MDPI}
}

@article{2429627610,
  title={Advancing self-directed learning in STEM education: integrating GPT-based learning aid with multimodal learning analytics},
  author={Lin, Chia-Ju and Wang, Wei-Sheng and Lee, Hsin-Yu and Li, Pin-Hui and Huang, Yueh-Min and Wu, Ting-Ting},
  journal={Journal of Research on Technology in Education},
  pages={1--19},
  year={2025},
  publisher={Taylor \& Francis}
}

@article{227355655,
  title={Recognitions of image and speech to improve learning diagnosis on STEM collaborative activity for precision education},
  author={Lin, Chia-Ju and Wang, Wei-Sheng and Lee, Hsin-Yu and Huang, Yueh-Min and Wu, Ting-Ting},
  journal={Education and Information Technologies},
  volume={29},
  number={11},
  pages={13859--13884},
  year={2024},
  publisher={Springer}
}

@article{1161441004,
  title={Investigating students’ cognitive processes in generative AI-assisted digital multimodal composing and traditional writing},
  author={Liu, Meilu and Zhang, Lawrence Jun and Biebricher, Christine},
  journal={Computers \& Education},
  volume={211},
  pages={104977},
  year={2024},
  publisher={Elsevier}
}

@article{603534886,
  title={Exploring students' cognitive and affective states during problem solving through multimodal data: Lessons learned from a programming activity},
  author={Mangaroska, Katerina and Sharma, Kshitij and Ga{\v{s}}evi{\'c}, Dragan and Giannakos, Michail},
  journal={Journal of Computer Assisted Learning},
  volume={38},
  number={1},
  pages={40--59},
  year={2022},
  publisher={Wiley Online Library}
}

@inproceedings{2737776963,
  title={``It's Really Enjoyable to See Me Solve the Problem like a Hero''': GenAI-enhanced Data Comics as a Learning Analytics Tool},
  author={Milesi, Mikaela E and Alfredo, Riordan and Echeverria, Vanessa and Yan, Lixiang and Zhao, Linxuan and Tsai, Yi-Shan and Martinez-Maldonado, Roberto},
  booktitle={Extended abstracts of the CHI conference on human factors in computing systems},
  pages={1--7},
  year={2024}
}

@article{1552158788,
  title={Smart glasses for 3D multimodal composition},
  author={Mills, Kathy A and Brown, Alinta},
  journal={Learning, Media and Technology},
  volume={50},
  number={2},
  pages={156--177},
  year={2025},
  publisher={Taylor \& Francis}
}

@article{1278817005,
  title={Using multimodal learning analytics as a formative assessment tool: Exploring collaborative dynamics in mathematics teacher education},
  author={Moon, Jewoong and Yeo, Sheunghyun and Banihashem, Seyyed Kazem and Noroozi, Omid},
  journal={Journal of Computer Assisted Learning},
  volume={40},
  number={6},
  pages={2753--2771},
  year={2024},
  publisher={Wiley Online Library}
}

@article{190066185,
  title={Bridging LMS and generative AI: dynamic course content integration (DCCI) for enhancing student satisfaction and engagement via the ask ME assistant},
  author={Mzwri, Kovan and Turcs{\'a}nyi-Szabo, M{\'a}rta},
  journal={Journal of Computers in Education},
  pages={1--38},
  year={2025},
  publisher={Springer}
}

@inproceedings{3224774131,
  title={Providing Automated Feedback on Formative Science Assessments: Uses of Multimodal Large Language Models},
  author={Nguyen, Ha and Park, Saerok},
  booktitle={Proceedings of the 15th International Learning Analytics and Knowledge Conference},
  pages={803--809},
  year={2025}
}

@inproceedings{3888330750,
  title={Beyond the learning analytics dashboard: Alternative ways to communicate student data insights combining visualisation, narrative and storytelling},
  author={Fernandez Nieto, Gloria Milena and Kitto, Kirsty and Buckingham Shum, Simon and Martinez-Maldonado, Roberto},
  booktitle={LAK22: 12th international learning analytics and knowledge conference},
  pages={219--229},
  year={2022}
}

@article{116733479,
  title={Integration of artificial intelligence performance prediction and learning analytics to improve student learning in online engineering course},
  author={Ouyang, Fan and Wu, Mian and Zheng, Luyi and Zhang, Liyin and Jiao, Pengcheng},
  journal={International Journal of Educational Technology in Higher Education},
  volume={20},
  number={1},
  pages={4},
  year={2023},
  publisher={Springer}
}

@article{2005607968,
  title={Multimodal learning analytics of collaborative patterns during pair programming in higher education},
  author={Xu, Weiqi and Wu, Yajuan and Ouyang, Fan},
  journal={International Journal of Educational Technology in Higher Education},
  volume={20},
  number={1},
  pages={8},
  year={2023},
  publisher={Springer}
}

@article{2995141815,
  title={An artificial intelligence-driven learning analytics method to examine the collaborative problem-solving process from the complex adaptive systems perspective},
  author={Ouyang, Fan and Xu, Weiqi and Cukurova, Mutlu},
  journal={International Journal of Computer-Supported Collaborative Learning},
  volume={18},
  number={1},
  pages={39--66},
  year={2023},
  publisher={Springer}
}

@article{1500258376,
  title={Developing a multimodal classroom engagement analysis dashboard for higher-education},
  author={Sabuncuoglu, Alpay and Sezgin, T Metin},
  journal={Proceedings of the ACM on Human-Computer Interaction},
  volume={7},
  number={EICS},
  pages={1--23},
  year={2023},
  publisher={ACM New York, NY, USA}
}

@article{1844320601,
  title={Gaze-Driven Adaptive Learning System with ChatGPT-Generated Summaries},
  author={Santhosh, Jayasankar and Dengel, Andreas and Ishimaru, Shoya},
  journal={IEEE Access},
  volume={12},
  pages={173714--173733},
  year={2024},
  publisher={IEEE}
}

@article{261302708,
  title={Multimodal teacher dashboards: Challenges and opportunities of enhancing teacher insights through a case study},
  author={Lee-Cultura, Serena and Sharma, Kshitij and Giannakos, Michail N},
  journal={IEEE Transactions on Learning Technologies},
  volume={17},
  pages={181--201},
  year={2023},
  publisher={IEEE}
}

@article{780281159,
  title={Multimodal composing with generative AI: Examining preservice teachers’ processes and perspectives},
  author={Smith, Blaine E and Shimizu, Amanda Yoshiko and Burriss, Sarah K and Hundley, Melanie and Pendergrass, Emily},
  journal={Computers and Composition},
  volume={75},
  pages={102896},
  year={2025},
  publisher={Elsevier}
}

@article{1687167932,
  title={Using multimodal analytics to systemically investigate online collaborative problem-solving},
  author={Tang, Hengtao and Dai, Miao and Yang, Shuoqiu and Du, Xu and Hung, Jui-Long and Li, Hao},
  journal={Distance Education},
  volume={43},
  number={2},
  pages={290--317},
  year={2022},
  publisher={Taylor \& Francis}
}

@article{1285699194,
  title={A multimodal analysis of college students’ collaborative problem solving in virtual experimentation activities: A perspective of cognitive load},
  author={Du, Xu and Dai, Miao and Tang, Hengtao and Hung, Jui-Long and Li, Hao and Zheng, Jinqiu},
  journal={Journal of Computing in Higher Education},
  volume={35},
  number={2},
  pages={272--295},
  year={2023},
  publisher={Springer}
}

@article{1441411748,
  title={Enhancing self-directed learning and Python mastery through integration of a large language model and learning analytics dashboard},
  author={Liu, Ming and Wu, Zhongming and Dai, Haimin and Su, Yifei and Malik, Laiba and Liao, Jian and Zhang, Wei and Guo, Shuo and Liu, Li and Zhao, Junqiang},
  journal={British Journal of Educational Technology},
  year={2025},
  publisher={Wiley Online Library}
}

@inproceedings{3313249608,
  title={Classroom Simulacra: Building Contextual Student Generative Agents in Online Education for Learning Behavioral Simulation},
  author={Xu, Songlin and Wen, Hao-Ning and Pan, Hongyi and Dominguez, Dallas and Hu, Dongyin and Zhang, Xinyu},
  booktitle={Proceedings of the 2025 CHI Conference on Human Factors in Computing Systems},
  pages={1--26},
  year={2025}
}

@article{3522635517,
  title={Evidence-based multimodal learning analytics for feedback and reflection in collaborative learning},
  author={Yan, Lixiang and Echeverria, Vanessa and Jin, Yueqiao and Fernandez-Nieto, Gloria and Zhao, Linxuan and Li, Xinyu and Alfredo, Riordan and Swiecki, Zachari and Ga{\v{s}}evi{\'c}, Dragan and Martinez-Maldonado, Roberto},
  journal={British Journal of Educational Technology},
  volume={55},
  number={5},
  pages={1900--1925},
  year={2024},
  publisher={Wiley Online Library}
}

@article{1436887306,
  title={Enhancing EFL vocabulary learning with multimodal cues supported by an educational robot and an IoT-Based 3D book},
  author={Lin, Vivien and Yeh, Hui-Chin and Huang, Huai-Hsuan and Chen, Nian-Shing},
  journal={System},
  volume={104},
  pages={102691},
  year={2022},
  publisher={Elsevier}
}

@inproceedings{177743022,
  title={AI-Driven Intelligent Learning Companions: A Multimodal Fusion Framework for Personalized Education},
  author={You, Cunqian and Lu, Huijuan and Li, Ping and Zhao, Xiaoyu and Yao, Yudong},
  booktitle={2025 IEEE 34th Wireless and Optical Communications Conference (WOCC)},
  pages={424--428},
  year={2025},
  organization={IEEE}
}

@article{1935812764,
  title={Using multimodal learning analytics to model students’ learning behavior in animated programming classroom},
  author={Yusuf, Abdullahi and Noor, Norah Md and Bello, Shamsudeen},
  journal={Education and Information Technologies},
  volume={29},
  number={6},
  pages={6947--6990},
  year={2024},
  publisher={Springer}
}

@article{1675503665,
  title={AI and peer reviews in higher education: students’ multimodal views on benefits, differences and limitations},
  author={Zapata, Gabriela C and Cope, Bill and Kalantzis, Mary and Tzirides, Anastasia Olga and Saini, Akash K and Searsmith, Duane and Whiting, Jennifer and Kastania, Nikoleta Polyxeni and Castro, Vania and Kourkoulou, Theodora and others},
  journal={Technology, Pedagogy and Education},
  pages={1--19},
  year={2025},
  publisher={Taylor \& Francis}
}

@article{2737977054,
  title={Can AI-generated pedagogical agents (AIPA) replace human teacher in picture book videos? The effects of appearance and voice of AIPA on children’s learning},
  author={Bai, Jie and Cheng, Xiulan and Zhang, Hui and Qin, Yihang and Xu, Tao and Zhou, Yun},
  journal={Education and Information Technologies},
  pages={1--21},
  year={2025},
  publisher={Springer}
}

@inproceedings{209328204,
  title={METS: Multimodal learning analytics of embodied teamwork learning},
  author={Zhao, Linxuan and Swiecki, Zachari and Gasevic, Dragan and Yan, Lixiang and Dix, Samantha and Jaggard, Hollie and Wotherspoon, Rosie and Osborne, Abra and Li, Xinyu and Alfredo, Riordan and others},
  booktitle={LAK23: 13th International learning analytics and knowledge conference},
  pages={186--196},
  year={2023}
}

@article{3602263061,
  title={Towards automated transcribing and coding of embodied teamwork communication through multimodal learning analytics},
  author={Zhao, Linxuan and Ga{\v{s}}evi{\'c}, Dragan and Swiecki, Zachari and Li, Yuheng and Lin, Jionghao and Sha, Lele and Yan, Lixiang and Alfredo, Riordan and Li, Xinyu and Martinez-Maldonado, Roberto},
  journal={British Journal of Educational Technology},
  volume={55},
  number={4},
  pages={1673--1702},
  year={2024},
  publisher={Wiley Online Library}
}

@inproceedings{martinez-maldonado_data_2020,
	address = {Honolulu HI USA},
	title = {From {Data} to {Insights}: {A} {Layered} {Storytelling} {Approach} for {Multimodal} {Learning} {Analytics}},
	isbn = {978-1-4503-6708-0},
	shorttitle = {From {Data} to {Insights}},
	url = {https://dl.acm.org/doi/10.1145/3313831.3376148},
	doi = {10.1145/3313831.3376148},
	language = {en},
	urldate = {2023-04-27},
	booktitle = {Proceedings of the 2020 {CHI} {Conference} on {Human} {Factors} in {Computing} {Systems}},
	publisher = {ACM},
	author = {Martinez-Maldonado, Roberto and Echeverria, Vanessa and Fernandez Nieto, Gloria and Buckingham Shum, Simon},
	month = apr,
	year = {2020},
	pages = {1--15},
}

@article{jarvela_what_2021,
	title = {What multimodal data can tell us about the students’ regulation of their learning process?},
	volume = {72},
	issn = {09594752},
	url = {https://linkinghub.elsevier.com/retrieve/pii/S095947521830416X},
	doi = {10.1016/j.learninstruc.2019.04.004},
	language = {en},
	urldate = {2023-04-05},
	journal = {Learning and Instruction},
	author = {Järvelä, Sanna and Malmberg, Jonna and Haataja, Eetu and Sobocinski, Marta and Kirschner, Paul A.},
	month = apr,
	year = {2021},
	pages = {101203},
}

@inproceedings{vrzakova_focused_2020,
	address = {Frankfurt Germany},
	title = {Focused or stuck together: multimodal patterns reveal triads' performance in collaborative problem solving},
	isbn = {978-1-4503-7712-6},
	shorttitle = {Focused or stuck together},
	url = {https://dl.acm.org/doi/10.1145/3375462.3375467},
	doi = {10.1145/3375462.3375467},
	language = {en},
	urldate = {2023-02-10},
	booktitle = {Proceedings of the {Tenth} {International} {Conference} on {Learning} {Analytics} \& {Knowledge}},
	publisher = {ACM},
	author = {Vrzakova, Hana and Amon, Mary Jean and Stewart, Angela and Duran, Nicholas D. and D'Mello, Sidney K.},
	month = mar,
	year = {2020},
	pages = {295--304},
}

\newpage
\appendix

\begin{landscape}

\section{Corpus Table} \label{sec:corpus_table}
Table \ref{tab:corpus_table} enumerates the 122 papers in this literature review's corpus.

\begin{longtable}{llp{3.4in}rll}
\toprule
UUID & First Author & Title & Year & Publication & Corpus \\
\midrule
2456887548 \cite{2456887548} & Alyuz & An Unobtrusive And Multimodal Approach For Behavioral Engagement Detection Of Students & 2017 & MIE & A\\
818492192 \cite{818492192} & Andrade & Understanding Student Learning Trajectories Using Multimodal Learning Analytics Within An Embodied-Interaction Learning Environment & 2017 & LAK & A \\
425012016 \cite{425012016} & Anton & The Human Condition: Modal and Interactive Advantages of Teacher over AI Feedback on Children's Mathematical Performance & 2025 & IDC & B \\
3637456466 \cite{3637456466} & Ashwin & Impact Of Inquiry Interventions On Students In E-Learning And Classroom Environments Using Affective Computing Framework & 2020 & UMUAI & A\\
3448122334 \cite{3448122334} & Aslan & Investigating The Impact Of A Real-Time, Multimodal Student Engagement Analytics Technology In Authentic Classrooms & 2019 & CHI & A \\
2668965770 \cite{2668965770} & Aslan & Exploring kid space in the wild: a preliminary study of multimodal and immersive collaborative play-based learning experiences & 2022 & ETRD & B \\
1886134458 \cite{1886134458} & Azcona & Personalizing Computer Science Education By Leveraging Multimodal Learning Analytics & 2018 & FIE & A \\
3146393211 \cite{3146393211} & Birt & Mobile Mixed Reality For Experiential Learning And Simulation In Medical And Health Sciences Education & 2018 & Information & A \\
1326191931 \cite{1326191931} & Chan & Multimodal Learning Analytics In A Laboratory Classroom & 2019 & MLPALA & A \\
4089325423 \cite{4089325423} & Chan & Predicting behavior change in students with special education needs using multimodal learning analytics & 2023 & Access & B \\
2936220551 \cite{2936220551} & Chango & Multi-Source And Multimodal Data Fusion For Predicting Academic Performance In Blended Learning University Courses & 2020 & CEE & A \\
4277812050 \cite{4277812050} & Chango & Improving Prediction Of Students' Performance In Intelligent Tutoring Systems Using Attribute Selection And Ensembles Of Different Multimodal Data Sources & 2021 & JCHE & A \\
1196965665 \cite{1196965665} & Chejara & How to build more generalizable models for collaboration quality? lessons learned from exploring multi-context audio-log datasets using multimodal learning analytics & 2023 & LAK & B \\
1731146538 \cite{1731146538} & Chejara & Impact of window size on the generalizability of collaboration quality estimation models developed using Multimodal Learning Analytics & 2023 & LAK & B \\
1426267857 \cite{1426267857} & Chen & Affect, Support, And Personal Factors: Multimodal Causal Models Of One-On-One Coaching & 2021 & JEDM & A \\
2764645776 \cite{2764645776} & Chen & MindScratch: A Visual Programming Support Tool for Classroom Learning Based on Multimodal Generative AI & 2025 & IJHCI & B \\
328477558 \cite{328477558} & Chen & Unpacking help-seeking process through multimodal learning analytics: A comparative study of ChatGPT vs Human expert & 2025 & CompEdu & B \\
1225141845 \cite{1225141845} & Cheung & Exploring students' multimodal representations of ideas about epistemic reading of scientific texts in generative AI tools & 2025 & JSET & B \\
3304069824 \cite{3304069824} & Civit & Class integration of ChatGPT and learning analytics for higher education & 2024 & Expert Sys & B \\
3809293172 \cite{3809293172} & Closser & Blending Learning Analytics And Embodied Design To Model Students' Comprehension Of Measurement Using Their Actions, Speech, And Gestures & 2021 & IJCCI & A \\
570697424 \cite{cohn2024multimodal} & Cohn & A multimodal approach to support teacher, researcher and AI collaboration in STEM+ C learning environments & 2025 & BJET & B \\
3537775194 \cite{3537775194} & Contero & Personalized and Timely Feedback in Online Education: Enhancing Learning with Deep Learning and Large Language Models & 2025 & MTI & B \\
4019205162 \cite{4019205162} & Cornide-Reyes & Introducing Low-Cost Sensors Into The Classroom Settings: Improving The Assessment In Agile Practices With Multimodal Learning Analytics & 2019 & Sensors & A \\
2846172025 \cite{2846172025} & Cosentino & Generative AI and multimodal data for educational feedback: Insights from embodied math learning & 2025 & BJET & B \\
1576545447 \cite{1576545447} & Cukurova & Artificial Intelligence And Multimodal Data In The Service Of Human Decision-Making: A Case Study In Debate Tutoring & 2019 & BJET & A \\
1609706685 \cite{1609706685} & Di Mitri & Learning Pulse: A Machine Learning Approach For Predicting Performance In Self-Regulated Learning Using Multimodal Data & 2017 & LAK & A \\
2070224207 \cite{2070224207} & Di Mitri & Detecting Medical Simulation Errors With Machine Learning And Multimodal Data & 2019 & CAIM & A \\
3009548670 \cite{di2020real} & Di Mitri & Real-Time Multimodal Feedback With The Cpr Tutor & 2020 & AIED & A\\
1763513559 \cite{1763513559} & Di Mitri & Keep Me In The Loop: Real-Time Feedback With Multimodal Data & 2021 & IJAIED & A \\
1296637108 \cite{1296637108} & Echeverria & Towards Collaboration Translucence: Giving Meaning To Multimodal Group Data & 2019 & CHI & A \\
1040787959 \cite{1040787959} & Echeverria & TeamSlides: A multimodal teamwork analytics dashboard for teacher-guided reflection in a physical learning space & 2024 & LAK & B \\
1581261659 \cite{1581261659} & Emerson & Early Prediction Of Visitor Engagement In Science Museums With Multimodal Learning Analytics & 2020 & ICMI & A \\
1598166515 \cite{1598166515} & Emerson & Multimodal Learning Analytics For Game-Based Learning & 2020 & BJET & A \\
4035649049 \cite{4035649049} & Fernández-Nieto & Storytelling With Learner Data: Guiding Student Reflection On Multimodal Team Data & 2021 & TLT & A \\
151988148 \cite{151988148} & Fernández-Nieto & Data storytelling editor: A teacher-centred tool for customising learning analytics dashboard narratives & 2024 & LAK & B \\
483140962 \cite{483140962} & Fwa & Investigating Multimodal Affect Sensing In An Affective Tutoring System Using Unobtrusive Sensors & 2018 & PPIG & A \\
4278392816 \cite{4278392816} & Giannakos & Multimodal Data As A Means To Understand The Learning Experience & 2019 & IJIM & A \\
2243240858 \cite{2243240858} & Goslen & Llm-based student plan generation for adaptive scaffolding in game-based learning environments & 2025 & IJAIED & B \\
853680639 \cite{853680639} & Henderson & Sensor-Based Data Fusion For Multimodal Affect Detection In Game-Based Learning Environments & 2019 & EDM & A \\
3398902089 \cite{3398902089} & Järvelä & What Multimodal Data Can Tell Us About The Students’ Regulation Of Their Learning Process? & 2019 & LAI & A\\
86191824 \cite{86191824} & Jiang & Examining How Different Modes Mediate Adolescents’ Interactions During Their Collaborative Multimodal Composing Processes & 2019 & ILE \\
141378338 \cite{141378338} & Jiang & How Did the Generative Artificial Intelligence-Assisted Digital Multimodal Composing Process Facilitate the Production of Quality Digital Multimodal Compositions: Toward a Process-Genre Integrated Model & 2025 & TESQ & B \\
2166765216 \cite{2166765216} & Jin & Chatting with a learning analytics dashboard: The role of generative AI literacy on learner interaction with conventional and scaffolding chatbots & 2025 & LAK & B \\
2280467946 \cite{2280467946} & Kim & Multimodal Writing Evaluation in Digital Storytelling using Video-Based Output: Comparing performance of AI and Human Raters. & 2024 & ICMET & B \\
32184286 \cite{32184286} & Kubsch & Once More With Feeling: Emotions In Multimodal Learning Analytics & 2022 & MMLA Handbook & A \\
205660768 \cite{205660768} & Larmuseau & Multimodal Learning Analytics To Investigate Cognitive Load During Online Problem Solving & 2020 & BJET & A \\
1877483551 \cite{1877483551} & Lee-Cultura & Motion-Based Educational Games: Using Multi-Modal Data To Predict Player’S Performance & 2020 & COG & A\\
3660066725 \cite{3660066725} & Lee-Cultura & Children'S Play And Problem Solving In Motion-Based Educational Games: Synergies Between Human Annotations And Multi-Modal Data & 2021 & IDC & A\\
3856280479 \cite{3856280479} & Lee-Cultura & Children'S Play And Problem-Solving In Motion-Based Learning Technologies Using A Multi-Modal Mixed Methods Approach & 2021 & IJCCI & A\\
962997360 \cite{962997360} & Lehtonen & Multimodal Communication and Peer Interaction during Equation-Solving Sessions with and without Tangible Technologies & 2023 & MTI & B \\
2429627610 \cite{2429627610} & Lin & Advancing self-directed learning in STEM education: integrating GPT-based learning aid with multimodal learning analytics & 2025 & JRTE & B \\
227355655 \cite{227355655} & Lin & Recognitions of image and speech to improve learning diagnosis on STEM collaborative activity for precision education & 2024 & EIT & B \\
804659204 \cite{804659204} & Liu & Towards Smart Educational Recommendations With Reinforcement Learning In Classroom & 2018 & TALE & A\\
3783339081 \cite{3783339081} & Liu & A Novel Method For The In-Depth Multimodal Analysis Of Student Learning Trajectories In Intelligent Tutoring Systems & 2018 & JLA & A\\
3796180663 \cite{3796180663} & Liu & Learning Linkages: Integrating Data Streams Of Multiple Modalities And Timescales & 2018 & JCAL & A\\
1161441004 \cite{1161441004} & Liu & Investigating students' cognitive processes in generative AI-assisted digital multimodal composing and traditional writing & 2024 & CompEdu & B \\
518268671 \cite{518268671} & López & Using Multimodal Learning Analytics To Explore Collaboration In A Sustainability Co-Located Tabletop Game & 2021 & ECGBL & A\\
566043228 \cite{566043228} & Ma & Automatic Student Engagement In Online Learning Environment Based On Neural Turing Machine & 2021 & IJIET & A\\
3754172825 \cite{3754172825} & Ma & Detecting Impasse During Collaborative Problem Solving With Multimodal Learning Analytics & 2022 & LAK & A\\
147203129 \cite{147203129} & Mangaroska & Multimodal Learning Analytics To Inform Learning Design: Lessons Learned From Computing Education & 2020 & JLA & A\\
603534886 \cite{603534886} & Mangaroska & Exploring students' cognitive and affective states during problem solving through multimodal data: Lessons learned from a programming activity & 2022 & JCAL & B \\
1847468084 \cite{1847468084} & Martin & Computationally Augmented Ethnography: Emotion Tracking And Learning In Museum Games & 2019 & ICQE & A\\
2879332689 \cite{martinez-maldonado_data_2020} & Martinez-Maldonado & From Data To Insights: A Layered Storytelling Approach For Multimodal Learning Analytics & 2020 & CHI & A\\
549526582 \cite{martinez2023lessons} & Martinez-Maldonado & Lessons learnt from a multimodal learning analytics deployment in-the-wild & 2023 & TOCHI & B \\
2737776963 \cite{2737776963} & Milesi & ``It's Really Enjoyable to See Me Solve the Problem like a Hero'': GenAI-enhanced Data Comics as a Learning Analytics Tool & 2024 & CHI EA & B \\
1552158788 \cite{1552158788} & Mills & Smart glasses for 3D multimodal composition & 2025 & LMT & B \\
1278817005 \cite{1278817005} & Moon & Using multimodal learning analytics as a formative assessment tool: Exploring collaborative dynamics in mathematics teacher education & 2024 & JCAL & B \\
2155422499 \cite{2155422499} & Morell & A Multimodal Analysis Of Pair Work Engagement Episodes: Implications For Emi Lecturer Training & 2022 & JEAP & A\\
190066185 \cite{190066185} & Mzwri & Bridging LMS and Generative AI: Dynamic Course Content Integration (DCCI) for Connecting LLMs to Course Content--The Ask ME Assistant & 2025 & JCE & B \\
2273914836 \cite{2273914836} & Nasir & Many Are The Ways To Learn Identifying Multi-Modal Behavioral Profiles Of Collaborative Learning In Constructivist Activities & 2022 & IJCSCL & A\\
1469065963 \cite{1469065963} & Nguyen & Examining Socially Shared Regulation And Shared Physiological Arousal Events With Multimodal Learning Analytics & 2022 & BJET & A\\
3224774131 \cite{3224774131} & Nguyen & Providing Automated Feedback on Formative Science Assessments: Uses of Multimodal Large Language Models & 2025 & LAK & B \\
3888330750 \cite{3888330750} & Nieto & Beyond the learning analytics dashboard: Alternative ways to communicate student data insights combining visualisation, narrative and storytelling & 2022 & LAK & B \\
2345021698 \cite{2345021698} & Noël & Exploring Collaborative Writing Of User Stories With Multimodal Learning Analytics: A Case Study On A Software Engineering Course & 2018 & Access & A\\
2609260641 \cite{2609260641} & Noël & Visualizing Collaboration In Teamwork: A Multimodal Learning Analytics Platform For Non-Verbal Communication & 2022 & DAMLE & A\\
2497456347 \cite{2497456347} & Ochoa & The Rap System: Automatic Feedback Of Oral Presentation Skills Using Multimodal Analysis And Low-Cost Sensors & 2018 & LAK & A\\
2634033325 \cite{2634033325} & Ochoa & Controlled Evaluation Of A Multimodal System To Improve Oral Presentation Skills In A Real Learning Setting & 2020 & BJET & A\\
3051560548 \cite{3051560548} & Olsen & Temporal Analysis Of Multimodal Data To Predict Collaborative Learning Outcomes & 2020 & BJET & A\\
116733479 \cite{116733479} & Ouyang & Integration of artificial intelligence performance prediction and learning analytics to improve student learning in online engineering course & 2023 & ETHE & B \\
2005607968 \cite{2005607968} & Ouyang & Multimodal learning analytics of collaborative patterns during pair programming in higher education & 2023 & ETHE & B \\
2995141815 \cite{2995141815} & Ouyang & An artificial intelligence-driven learning analytics method to examine the collaborative problem-solving process from the complex adaptive systems perspective & 2023 & IJCSCL & B \\
123412197 \cite{123412197} & Papamitsiou & Utilizing Multimodal Data Through Fsqca To Explain Engagement In Adaptive Learning & 2020 & TLT & A\\
85990093 \cite{85990093} & Petukhova & Multimodal Markers Of Persuasive Speech : Designing A Virtual Debate Coach & 2017 & INTERSPEECH & A\\
957160695 \cite{957160695} & Petukhova & Virtual Debate Coach Design: Assessing Multimodal Argumentation Performance & 2017 & ICMI & A\\
1374035721 \cite{1374035721} & Pham & Attentivelearner2: A Multimodal Approach For Improving Mooc Learning On Mobile Devices & 2017 & AIED & A\\
2836996318 \cite{2836996318} & Pham & Predicting Learners' Emotions In Mobile Mooc Learning Via A Multimodal Intelligent Tutor & 2018 & ITS & A\\
3135645357 \cite{3135645357} & Prieto & Multimodal Teaching Analytics: Automated Extraction Of Orchestration Graphs From Wearable Sensor Data & 2018 & JCAL & A\\
3408664396 \cite{3408664396} & Psaltis & Multimodal Student Engagement Recognition In Prosocial Games & 2017 & T-CIAIG & A\\
3308658121 \cite{3308658121} & Reilly & Exploring Collaboration Using Motion Sensors And Multi-Modal Learning Analytics & 2018 & EDM & A\\
1500258376 \cite{1500258376} & Sabuncuoglu & Developing a multimodal classroom engagement analysis dashboard for higher-education & 2023 & PACM HCI & B \\
1844320601 \cite{1844320601} & Santhosh & Gaze-Driven Adaptive Learning System with ChatGPT-Generated Summaries & 2024 & Access & B \\
3625722965 \cite{3625722965} & Sanusi & Table Tennis Tutor: Forehand Strokes Classification Based On Multimodal Data And Neural Networks & 2021 & Sensors & A\\
2000036002 \cite{2000036002} & Sharma & Predicting Learners’ Effortful Behaviour In Adaptive Assessment Using Multimodal Data & 2020 & LAK & A\\
261302708 \cite{261302708} & Sharma & Multimodal teacher dashboards: Challenges and opportunities of enhancing teacher insights through a case study & 2023 & TLT & B \\
780281159 \cite{780281159} & Smith & Multimodal composing with generative AI: Examining preservice teachers' processes and perspectives & 2025 & CompComp & B \\
1118315889 \cite{1118315889} & Spikol & Using Multimodal Learning Analytics To Identify Aspects Of Collaboration In Project-Based Learning & 2017 & CSCL & A\\
3339002981 \cite{3339002981} & Spikol & Estimation Of Success In Collaborative Learning Based On Multimodal Learning Analytics Features & 2017 & ICALT & A\\
1637690235 \cite{1637690235} & Spikol & Supervised Machine Learning In Multimodal Learning Analytics For Estimating Success In Project-Based Learning & 2018 & JCAL & A\\
3796643912 \cite{3796643912} & Standen & An Evaluation Of An Adaptive Learning System Based On Multimodal Affect Recognition For Learners With Intellectual Disabilities & 2020 & BJET & A\\
2181637610 \cite{2181637610} & Starr & Toward Using Multi-Modal Learning Analytics To Support And Measure Collaboration In Co-Located Dyads & 2018 & ICLS & A\\
1315379489 \cite{1315379489} & Sümer & Multimodal Engagement Analysis From Facial Videos In The Classroom & 2021 & TAC & A\\
3093310941 \cite{3093310941} & Tanaka & Embodied Conversational Agents For Multimodal Automated Social Skills Training In People With Autism Spectrum Disorders & 2017 & PLOS & A\\
1345598079 \cite{1345598079} & Tancredi & Intermodality In Multimodal Learning Analytics For Cognitive Theory Development: A Case From Embodied Design For Mathematics Learning & 2022 & MMLA Handbook & A\\
1687167932 \cite{1687167932} & Tang & Using multimodal analytics to systemically investigate online collaborative problem-solving & 2022 & DistEdu & B \\
1285699194 \cite{1285699194} & Tang & A multimodal analysis of college students' collaborative problem solving in virtual experimentation activities: A perspective of cognitive load & 2023 & JCHE & B \\
433919853 \cite{433919853} & Tisza & Understanding Fun In Learning To Code: A Multi-Modal Data Approach & 2022 & IDC & A\\
1770989706 \cite{1770989706} & Vrzakova & Focused Or Stuck Together: Multimodal Patterns Reveal Triads' Performance In Collaborative Problem Solving & 2020 & LAK & A\\
2055153191 \cite{2055153191} & Vujovic & Round Or Rectangular Tables For Collaborative Problem Solving? A Multimodal Learning Analytics Study & 2020 & BJET & A\\
3095923626 \cite{3095923626} & Worsley & A Multimodal Analysis Of Making & 2017 & IJAIED & A\\
3309250332 \cite{3309250332} & Worsley & (Dis)Engagement Matters: Identifying Efficacious Learning Practices With Multimodal Learning Analytics & 2018 & LAK & A\\
666050348 \cite{666050348} & Worsley & Multicraft: A Multimodal Interface For Supporting And Studying Learning In Minecraft & 2021 & HCII & A\\
1441411748 \cite{1441411748} & Wu & Enhancing self‐directed learning and Python mastery through integration of a large language model and learning analytics dashboard & 2025 & BJET & B \\
3313249608 \cite{3313249608} & Xu & Classroom Simulacra: Building Contextual Student Generative Agents in Online Education for Learning Behavioral Simulation & 2025 & CHI & B \\
3522635517 \cite{3522635517} & Yan & Evidence‐based multimodal learning analytics for feedback and reflection in collaborative learning & 2024 & BJET & B \\
1019093033 \cite{1019093033} & Yang & Prime: Block-Wise Missingness Handling For Multi-Modalities In Intelligent Tutoring Systems & 2019 & MMM & A\\
1436887306 \cite{1436887306} & Yeh & Enhancing EFL vocabulary learning with multimodal cues supported by an educational robot and an IoT-Based 3D book & 2022 & System & B \\
177743022 \cite{177743022} & You & AI-Driven Intelligent Learning Companions: A Multimodal Fusion Framework for Personalized Education & 2025 & WOCC & B \\
1935812764 \cite{1935812764} & Yusuf & Using multimodal learning analytics to model students' learning behavior in animated programming classroom & 2024 & EIT & B \\
1675503665 \cite{1675503665} & Zapata & AI and peer reviews in higher education: students' multimodal views on benefits, differences and limitations & 2025 & TPE & B \\
2737977054 \cite{2737977054} & Zhang & Can AI-generated pedagogical agents (AIPA) replace human teacher in picture book videos? The effects of appearance and voice of AIPA on children's learning & 2025 & EIT & B \\
209328204 \cite{209328204} & Zhao & METS: Multimodal learning analytics of embodied teamwork learning & 2023 & LAK & B \\
3602263061 \cite{3602263061} & Zhao & Towards automated transcribing and coding of embodied teamwork communication through multimodal learning analytics & 2024 & BJET & B \\
\bottomrule
\caption{Each of the 122 works in our corpus.}
\label{tab:corpus_table}
\end{longtable}
\end{landscape}

\section{Corpus Distillation Procedure} \label{app:corpus_distillation_procedure}

This appendix contains a detailed account of the steps we took to gather relevant works for our literature review and how we distilled the initial search results to the 73 and 49 papers for Corpora A and B, respectively.

\subsection{Literature Search} \label{subsec:literature_search_appendix}

The literature search for both corpora was based on search strings collaboratively defined and agreed upon by the authors as representative of the target research space. Rather than conducting queries manually, we used SerpAPI~\cite{serpapi}, a third-party Google Scholar scraping API selected for its ability to return organic search results---unlike alternatives such as \texttt{scholarly}~\cite{cholewiak2021scholarly} and \texttt{gscholar}~\cite{gscholar}, whose outputs differ from browser-based queries.

For Corpus~A, we queried Google Scholar via API for papers published between January 2017 and October 2022. The 2017 cutoff was chosen to capture developments from the past five years while excluding earlier foundational work, which is discussed in Section~\ref{sec:intro} but not included in the corpus. The Corpus~B search was conducted in August 2025 and backdated to begin in November 2022, covering the period following the release of ChatGPT. We timed the search to follow major conference publication cycles (LAK, AIED, EDM, and L@S) to ensure comprehensive coverage.

The Corpus~A search included 14 distinct phrases, each queried three times using variations of the word \textit{multimodal} (\textit{multimodal}, \textit{multi-modal}, and \textit{multi modal}) as prefixes.\footnote{The term ``xai'' was included to identify works on explainable AI in learning and training contexts; however, no relevant results were returned during the initial search.} For Corpus~B, we used 12 updated queries reflecting recent developments in GenAI and LLMs, employing only the standard spelling of \textit{multimodal} after confirming that alternative spellings had no impact on results. We also omitted broad terms such as ``multimodal survey'' and ``multimodal literature review,'' which surfaced naturally in other targeted searches. The complete list of search phrases is shown in Table~\ref{tab:app_search_terms}.

\begin{table}[htbp]
  \centering
  \caption{Full Corpus Search Terms}
  \label{tab:app_search_terms}

  \begin{subtable}[t]{0.48\linewidth}
    \centering
    \begin{tabular}{l}
      \toprule
      education technology \\
      explainable artificial intelligence \\
      learning analytics \\
      learning environments \\
      learning environments literature review \\
      learning environments survey \\
      literature review \\
      simulation environments \\
      survey \\
      training environments \\
      training environments literature review \\
      training environments survey \\
      tutoring systems \\
      xai \\
      \bottomrule
    \end{tabular}
    \caption{Corpus A Search Terms}
    \label{tab:app_search_terms_a}
  \end{subtable}
  \hfill
  \begin{subtable}[t]{0.48\linewidth}
    \centering
    \begin{tabular}{l}
      \toprule
      education technology \\
      learning analytics \\
      learning environments \\
      training environments \\
      simulation environments \\
      llm learning environments \\
      llm training environments \\
      llm learning analytics \\
      pedagogical agents \\
      llm pedagogical agents \\
      ChatGPT in education \\
      generative AI in education \\
      \bottomrule
    \end{tabular}
    \caption{Corpus B Search Terms}
    \label{tab:app_search_terms_b}
  \end{subtable}

\end{table}

For each search string, we collected the top five pages (100 publications) returned by Google Scholar. This top-5 cutoff was imposed for practical and financial reasons related to the subsequent construction of a citation graph (see Appendix~\ref{subsubsec:cgp_appendix}). SerpAPI limits citation queries to 20 citations per API call, requiring multiple calls for highly cited papers (e.g., five calls for a paper with 100 citations). Without a cutoff, the number of API calls would become intractable. 

The initial search yielded 4,200 papers for Corpus~A (14 search terms \texttimes 3 multimodal spelling variants \texttimes 100 results) and 1,200 papers for Corpus~B (12 search terms \texttimes 1 multimodal spelling variant). The full corpus reduction procedure is detailed in Table~\ref{tab:procedure} and discussed in the following subappendices. Each step is referenced using the corresponding Step ID in Table~\ref{tab:procedure}.

\begin{table}[htbp]
\renewcommand{\arraystretch}{1.2}
\centering
\begin{tabularx}{\linewidth}{@{}lX rr rr@{}}
\toprule
Step & Procedure & Removed A & Remain A & Removed B & Remain B \\
\midrule

0   & Literature search                        & 0    & 4200 & 0    & 1200 \\
1   & Remove duplicates                        & 2079 & 2121 & 355  & 845  \\
2   & Remove non-English                       & 1    & 2120 & 0    & 845  \\
3   & Remove degree-0 nodes/disconnected components & 589  & 1531 &  33  &  812 \\
4   & Iteratively remove degree-1 nodes        & 468     &   1063   & 253    & 559  \\
5   & Title reads                              & 675  & 388  & 305    & 254  \\
6   & Abstract reads                           &  261    &  127    & 110   & 144  \\
7   & Full paper reads                         &  54   &  \underline{\textbf{73}}   & 95   & \underline{\textbf{49}}   \\

\bottomrule
\end{tabularx}
\caption{Corpus reduction procedure.}
\label{tab:procedure}
\end{table}

We removed 2,079 duplicates from Corpus~A and 355 from Corpus~B by hashing paper titles (Table~\ref{tab:procedure}, Step~1), retaining the official published version when multiple copies existed. We then excluded one non-English paper from Corpus~A (Step~2), identified using spaCy FastLang~\cite{spacy_fastlang} and verified through manual inspection. After these steps, the combined search yielded 2,120 unique English-language papers for Corpus~A and 855 for Corpus~B.

\subsection{Study Selection} \label{subsec:study_selection_appendix}

To reduce the corpora to a reviewable set, we applied both quantitative and qualitative methods. First, we performed citation graph pruning (CGP) to distill the corpus algorithmically (Appendix~\ref{subsubsec:cgp_appendix}). This was followed by qualitative screening, detailed in Appendix~\ref{subsubsec:quality_control_appendix}.

\subsubsection{Citation Graph Pruning (Quantitative Corpus Reduction).} \label{subsubsec:cgp_appendix}

For visualization, analysis, and corpus distillation, we used NetworkX~\cite{networkX} to construct a directed citation graph for all remaining papers. Each node corresponds to a paper identified by its Google Scholar UUID, and each directed edge denotes a citation from one corpus paper to another. Following SerpAPI's ``cited by'' results, only inbound citation queries were required; citations from papers outside the list of remaining papers were ignored.

We first removed all 0-degree nodes and disconnected components (Step~3)---papers that neither cited nor were cited by any other paper in the corpus and components with no edges to or from the primary (i.e., largest by number of nodes) component. Because incoming and outgoing citations jointly determine degree, this approach balances early papers (with few outgoing edges) and recent papers (with few incoming edges). Step~3 removed 589 papers from Corpus~A and 33 from Corpus~B, resulting in connected citation graphs of 1,531 and 812 papers, respectively.

We then applied iterative degree‑1 pruning (Step~4), removing nodes with only one citation edge and repeating the process until none remained. Corpus~A required four iterations, removing 468 papers and yielding 1,063; Corpus~B required two, removing 253 and yielding 559. This approach allowed us to eliminate loosely connected papers unlikely to be central to the field. Given that multimodal learning and training research spans multiple disciplines (e.g., computer science, education, psychology), the authors agreed that papers with minimal citation connectivity were unlikely to meet the scope of this review. The CGP algorithm is detailed in Algorithm~\ref{alg:cgp}.

\begin{algorithm}
\caption{Citation Graph Pruning Algorithm}
\begin{algorithmic}[1]
\Require Acyclic directed graph $G=(V, E)$
\Procedure{Degree Trimming}{$G, n$}
\State $S, D \gets \{\}, \{\}$
\ForAll{$v \in V$}
    \If{$\deg(v) <= n$}
        $S = S \cup \{v\}$
    \EndIf
\EndFor
\ForAll{$v \in S$}
    \ForAll{$e \in E$}
        \If{$v \in e \land e \notin D$}
            $D = D \cup \{e\}$
        \EndIf
    \EndFor
\EndFor
\State \Return $\left(V \setminus S, E \setminus D\right)$
\EndProcedure

\Procedure{Subconnected Graph Trimming}{$G$}
    \State $\left[S_1, S_2, S_3,..., S_n \right] = \text{ConnectedComponent}(G), \text{where each } S_i = (V_i, E_i)$
    \State $j = \arg \max\{ |V_1|, |V_2|, |V_3|, ..., |V_n| \}$
    \State \Return $\left(V_j, E_j\right)$
\EndProcedure

\Procedure{Iterative Trimming}{$G$}
    \While{True}
        \State $G' = \text{DegreeTrimming}(G, 1), \text{where } G'=(V', E')$
        \If{$|V| == |V'|$}
            \State \textbf{break}
        \EndIf
    \EndWhile
    \State \Return $\left(V', E'\right)$
\EndProcedure
\State $G' = \text{DegreeTrimming}(G, 0)$ \Comment{Remove 0-deg vertices}
\State $G' = \text{SubconnectedGraphTrimming}(G')$ \Comment{Keep largest connected subgraph}
\State $G' = \text{IterativeTrimming}(G')$ \Comment{Iteratively remove 1-deg vertices until equilibrium}
\State \Return $G'$
\end{algorithmic}
\label{alg:cgp}
\end{algorithm}

At this point, we concluded the quantitative pruning procedure. The resulting citation graphs served as the basis for subsequent qualitative screening.

\subsubsection{Quality Control (Qualitative Corpus Reduction)} 
\label{subsubsec:quality_control_appendix}

Following quantitative pruning, qualitative screening further reduced each corpus according to the procedures summarized in Table~\ref{tab:procedure}. For Corpus~A, the remaining 1,063 papers proceeded through title, abstract, and full‑paper review. For Corpus~B, due to time constraints, we used an LLM‑as‑a‑Judge workflow~\cite{zheng2023judging} for title, abstract, and full‑paper decisions, with human verification on the final distilled set.

\paragraph{Title Screening.}
For Corpus~A, four reviewers independently evaluated all 1,063 titles for relevance to multimodal learning or training. Inclusion and exclusion were determined by majority vote, with ties resolved by a fifth reviewer. This resulted in 388 retained titles and 675 exclusions (Table~\ref{tab:procedure}, Step~5 for A). For Corpus~B, title decisions were made jointly by GPT‑4o and Gemini~2.5; agreement between both models determined inclusion or exclusion, and disagreements were adjudicated by a human reviewer. Title screening retained 254 papers, excluding 305 (Step~5 for B).

\paragraph{Abstract Screening.}
For abstract screening (Step~6), each Corpus~A abstract was reviewed by two reviewers using the exclusion criteria in Table~\ref{tab:abstract_exclusion_criteria}. Papers without unanimous reviewer agreement underwent a second round of review using majority voting. This yielded 127 retained abstracts and 261 exclusions. For Corpus~B, both LLM judges independently evaluated all abstracts under the same criteria. Agreement resulted in automatic inclusion or exclusion; disagreements were resolved by a human reviewer. A total of 144 abstracts were retained.

\begin{wraptable}{R}{9cm}
    \centering
    \begin{tabular}{l}
        \midrule
        1. Paper does not involve a learning or training environment \\
        2. Environment is VR‑only \\
        3. No multimodal data are analyzed \\
        4. No multimodal analysis methods are applied \\
        5. Paper is not original applied research \\
        \bottomrule
    \end{tabular}
    \caption{Exclusion criteria for abstract screening.}
    \label{tab:abstract_exclusion_criteria}
\end{wraptable}

\paragraph{Full‑Paper Screening.}
Full‑paper review followed the same exclusion framework with two additional criteria introduced during reading (Table~\ref{tab:full_paper_exclusion_criteria}). For Corpus~A, 127 papers were divided among five reviewers. Papers were labeled ``immediate accept,'' ``immediate exclude,'' or ``borderline.'' Exclusion required unanimous agreement across all reviewers. After this stage, 73 papers remained (Table~\ref{tab:procedure}, Step~7). For Corpus~B, both LLM judges evaluated all 144 papers end‑to‑end using the cumulative exclusion criteria, selecting 79 papers (including a human tie-breaker) for inclusion. Two human reviewers then manually reviewed and discussed each of these papers to assess their alignment with the scope of this review. Based on consensus coding~\cite{chinh2019ways}, 30 papers were excluded, resulting in a final set of 49 papers. This human-in-the-loop validation ensured that all retained papers met the inclusion criteria and were within the scope of this literature review.

\begin{wraptable}{R}{11cm}
    \centering
    \begin{tabular}{l}
        \midrule
        1. Results are not informative about learning or training \\
        2. Analysis methods cannot be determined from the manuscript \\
        \bottomrule
    \end{tabular}
    \caption{Additional exclusion criteria for full‑paper screening.}
    \label{tab:full_paper_exclusion_criteria}
\end{wraptable}

Across both corpora, qualitative screening ensured that only papers presenting original multimodal methods applied to learning or training environments advanced to the final analysis set: 73 papers for Corpus~A and 49 for Corpus~B. 

\subsection{Feature Extraction} 
\label{subsec:feature_extraction_appendix}

Feature extraction was performed after the full paper review stages (Table~\ref{tab:procedure}, Steps~7) and was conducted manually by two human reviewers for all 73 papers in Corpus~A and all 49 papers in Corpus~B. Extracted features included identifying information (e.g., title, author, year) and methodological descriptors (e.g., data collection media, modalities, fusion strategies, and analysis methods). Table~\ref{tab:feature_extraction} lists the initial feature set.

\begin{table}[htbp]
    \renewcommand{\arraystretch}{1.3}%
    \centering
    \begin{tabular}{p{0.25\linewidth}@{\hskip .1in} | @{\hskip .1in}p{0.65\linewidth}@{\hskip .1in}}
        \toprule
        Feature & Description\\
        \toprule
        UUID & Universally unique identifier on Google Scholar \\
        Title & Publication title \\
        First Author & Publication's first author \\
        Year & Year first publicly available \\
        Environment Type & Type of environment analyzed \\
        Data Collection Media & Types of data collected \\
        Modalities & Modalities used during analysis \\
        Analysis Methods & Methods applied in the analysis \\
        Fusion Type & Data fusion strategies used \\
        Publication Source & Journal, conference, workshop, etc. \\
        \bottomrule
    \end{tabular}
    \caption{Initial features extracted from each paper.}
    \label{tab:feature_extraction}
\end{table}

To ensure consistency, feature categories were initially discretized through inductive coding~\cite{thomas2003general}. Four reviewers each coded a portion of the papers in Corpus~A to define discrete feature sets. For example, ``video camera,'' ``webcam,'' and ``Kinect'' were consolidated under the medium ``video.'' Reviewers then re-extracted features into these discrete sets. The resulting circumscribing features are shown in Table~\ref{tab:circumscribing_features_1} (Cohen's $\kappa=0.87$).

\begin{table}[htbp]
    \renewcommand{\arraystretch}{1.3}%
    \centering
    \begin{tabular}{p{0.16\linewidth}@{\hskip .1in} | @{\hskip .1in}p{0.75\linewidth}@{\hskip .1in}}
        \toprule
        Feature & Feature Set\\
        \toprule
        Environment Type & learning, training\\
        Data Collection Media & video, audio, screen recording, eye tracking, logs, physiological sensor, interview, survey, participant produced artifacts, researcher produced artifacts, motion, text\\
        Modalities & affect, pose, gesture, activity, prosodic speech, transcribed speech, qualitative observation, logs, gaze, interview notes, survey, pulse, EDA, body temperature, blood pressure, EEG, fatigue, EMG, participant artifacts, researcher artifacts, audio spectrogram, text, pixel\\
        Analysis Methods & classification, regression, clustering, qualitative, statistical methods, network analysis, pattern extraction \\
        Fusion Type & early, mid, late, hybrid, other\\
        \bottomrule
    \end{tabular}
    \caption{First set of circumscribing features and their feature sets.}
    \label{tab:circumscribing_features_1}
\end{table}

A second Corpus~A feature extraction round gathered additional features supporting later analysis. These circumscribing features---environment setting, domain, participant interaction structure, didactic nature, level of instruction or training, analysis approach, and analysis results---are listed in Table~\ref{tab:circumscribing_features_2}. All were discretized except analysis results, which were recorded in free form for thematic analysis~\cite{braun2006using}. As with the first round, feature extraction for the second feature set of Corpus~A involved independent coding by two reviewers followed by consensus. For this round, Cohen's $\kappa$ prior to consensus was $0.71$.  

\begin{table}[htbp]
    \renewcommand{\arraystretch}{1.3}%
    \centering
    \begin{tabular}{p{0.27\linewidth}@{\hskip .1in} | @{\hskip .1in}p{0.48\linewidth}@{\hskip .1in}}
        \toprule
        Feature & Feature Set\\
        \toprule
        Environment Setting & physical, virtual, blended, unspecified\\
        Domain of Study & STEM, humanities, psychomotor skills, other, unspecified\\
        Participant Interaction Structure & individual, multi-person\\
        Didactic Nature & instructional, training, informal, unspecified\\
        Level of Instruction or Training & K-12, university, professional development, unspecified\\
        Analysis Approach & model-free, model-based\\
        Feedback & direct, indirect \\
        \bottomrule
    \end{tabular}
    \caption{Second set of circumscribing features and their feature sets.}
    \label{tab:circumscribing_features_2}
\end{table}

Once the feature sets were finalized, this process was applied to Corpus~B using two human reviewers for consensus coding (Cohen's $\kappa=0.68$ prior to consensus). For each corpus, final feature sets represent agreement between the reviewers who coded each paper. 

\section{Literature Review Limitations} \label{sec:literature_review_limitations}

This review has three primary limitations: the use of Google Scholar for the literature search, the application of citation graph pruning for corpus reduction, and inconsistencies in versioning across published papers. Each is discussed below.

\subsection{Google Scholar} 

Google Scholar, while widely used in academia and industry, presents challenges for reproducibility. Its proprietary ranking algorithm is opaque and presumably variable, with results influenced by factors such as user location, search history, the time of query, and possible A/B testing. Because the algorithm is constantly evolving and individualized, exact reproduction of our search results cannot be guaranteed.

\begin{wrapfigure}{l}{0.6\textwidth}
    \begin{center}
        \includegraphics[width=.5\textwidth]{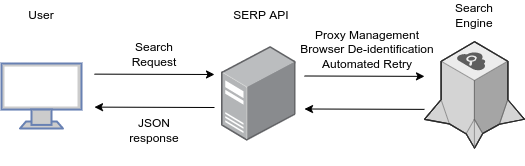}
    \end{center}
    \caption{Searching Google Scholar via SerpAPI.}
    \label{fig:serpAPI}
    \Description[Searching Google Scholar via SerpAPI]{Searching Google Scholar via SerpAPI.}
\end{wrapfigure}

To mitigate this, we used SerpAPI---a web scraping service that queries Google Scholar via randomized headers and proxies without user account data (Figure~\ref{fig:serpAPI}). SerpAPI's documentation confirms that no personal information is attached to API requests, and the API supports reproducibility through standardized, non-personalized queries. Additionally, SerpAPI recommends validating results using the included search URLs in browser incognito mode.

We also contacted SerpAPI directly, who confirmed: ``No, we don't add any personal information,'' and noted that others can reproduce results by using the same search parameters. While we agree this may be optimistic given Google's opacity, we are reasonably confident that our initial search was free from personal bias due to the nature of the SerpAPI interface. For reference, all searches were conducted in Nashville, TN, USA.

\subsection{Citation Graph Pruning} 

As detailed in Appendix Section~\ref{subsubsec:cgp_appendix}, we used citation graph pruning to programmatically reduce our corpus. This approach may have excluded some relevant works that had few citation connections within the corpus. However, our goal was to identify core contributions in the field---papers that either built upon, or were built upon by, other relevant works. We reasoned that isolated papers with few citation ties were less likely to represent foundational or widely used methods.

Importantly, even after CGP, over half the remaining papers were ultimately excluded during qualitative screening. This suggests that CGP still retained many irrelevant works, reinforcing our confidence that the method did not omit significant in-scope works.

\subsection{Versioning}

Many papers in our corpus appeared in multiple forms across preprint servers and publication venues, often with inconsistent metadata. We used the earliest available public release date when possible. However, discrepancies may remain, particularly for papers from 2022-2023, where preprint and publication dates may differ by months.

As a result, it is possible that some Corpus~B papers were written before the public release of ChatGPT (November 2022). Nonetheless, the sharp rise in publications in 2025 (see Figure~\ref{fig:corpus_dist_by_year}) strongly suggests that generative AI was the driving force behind these works, which our own analysis reinforces. Any misclassification in dating is expected to be minor and unlikely to affect our overall findings or conclusions.

\end{document}